\documentclass[10pt,twocolumn,letterpaper]{article}

\usepackage{cvpr}              

\usepackage{xspace}
\usepackage{colortbl}
\definecolor{resultYellow}{HTML}{FFF2CC}
\definecolor{resultOrange}{HTML}{FAD7AC}
\definecolor{resultRed}{HTML}{F4B4AE}
\definecolor{resultNA}{HTML}{F1F3F4}
\newcommand{\our}{FoamSurf\xspace}  

\usepackage{iftex}
\ifXeTeX
  \usepackage{fontspec}
\fi

\usepackage{graphicx}

\usepackage{booktabs}

\newcommand{\std}[1]{{\,\scriptsize$\pm$#1}}

\usepackage{afterpage}

\usepackage{capt-of}

\definecolor{cvprblue}{rgb}{0.21,0.49,0.74}
\usepackage[pagebackref,breaklinks,colorlinks,allcolors=cvprblue]{hyperref}

\def\confName{CVPR}
\def\confYear{2027}

\title{FoamSurf: A Voronoi-Surfel Representation for Differentiable Rendering}

\author{
Grzegorz Wilczy\'nski\textsuperscript{1,2} \quad
Julia Farganus\textsuperscript{3} \quad
Rafa{\l} Tobiasz\textsuperscript{1,2} \quad
Krzysztof Byrski\textsuperscript{1} \\
Miko{\l}aj Zieli\'nski\textsuperscript{1,4,5} \quad
Dominik Belter\textsuperscript{4} \quad
Przemysław~Spurek\textsuperscript{1,2} \\
\vspace{0.2cm}\\
\textsuperscript{1}Jagiellonian University  \quad
\textsuperscript{2} IDEAS Research Institute \quad
\textsuperscript{3}Wrocław University of  Science and Technology \\
\textsuperscript{4}Poznan University of Technology \quad
\textsuperscript{5}Jagiellonian Center for Artificial Intelligence \\
}

\begin{document}

\twocolumn[{%
\renewcommand\twocolumn[1][]{#1}%
\maketitle
\centering
\vspace{-0.5cm}
\includegraphics[width=0.99\textwidth]{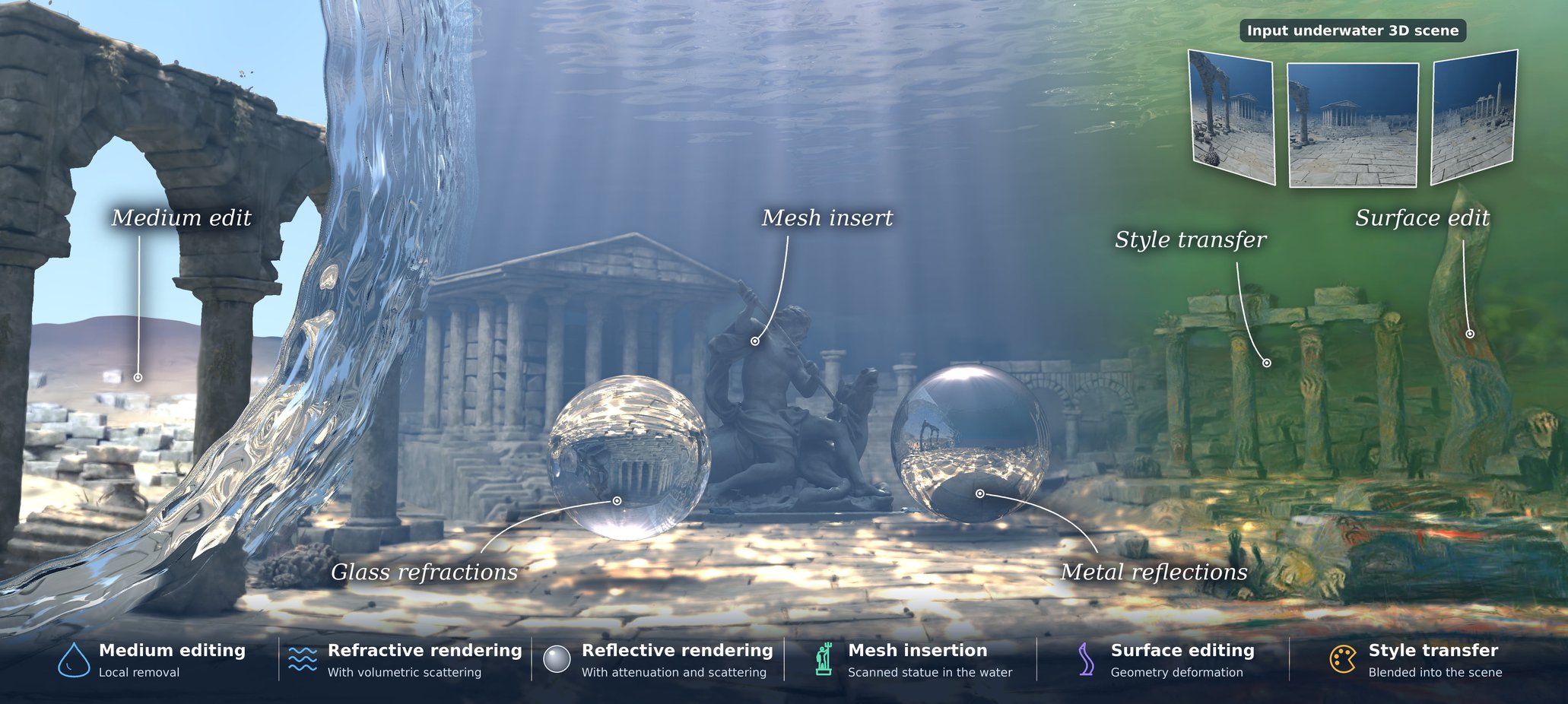}
\captionof{figure}{\our{} gives the surfaces and the water their own primitives, so an edit reaches one and leaves the other in place. The scene was rendered in Blender and reconstructed from the resulting photographs, three of which appear in the inset. Every panel is re-rendered by the same ray tracer after the change, without retraining. Removing the medium inside a boundary uncovers seabed that the cameras saw only through the water. The spheres and the statue are traced along rays that cross the water as camera rays do. Statue: ``Neptune -- Louvre Museum'' by Benjamin Bardou, CC BY 4.0.}
\label{fig:teaser}
\vspace{1em}
}]

\begin{figure*}[t]
  \centering
  \includegraphics[width=\textwidth]{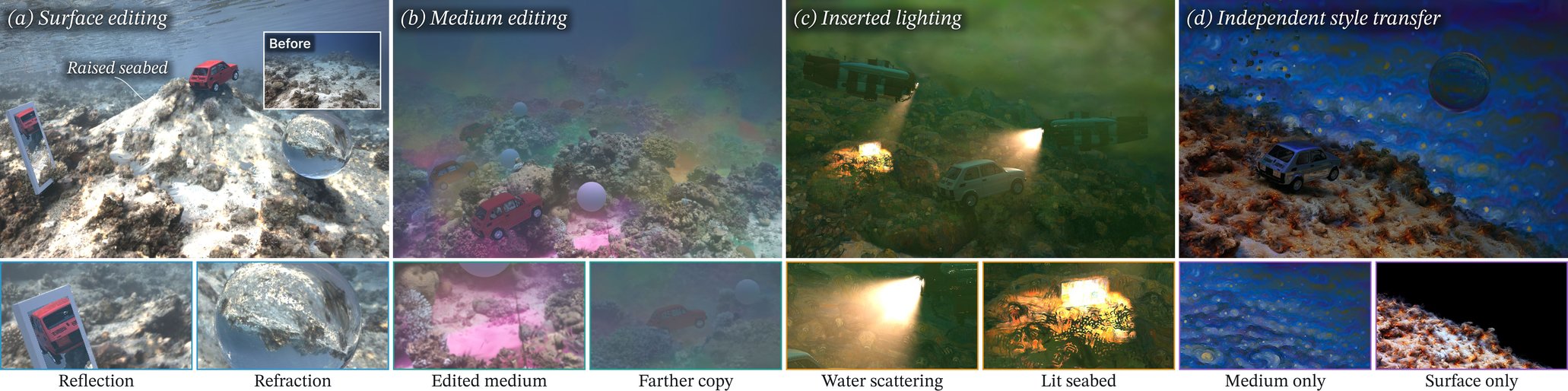}
  \caption{Applications on reconstructed real scenes, re-rendered by our tracer after each change and without retraining. (a) The seabed is deformed while the water stays as reconstructed, and the inset shows the scene before. The close-ups follow rays that no camera carried. (b) The medium is edited inside a chosen region, and copies of one mesh stand at several distances. Each copy is attenuated and veiled in proportion to the water in front of it. (c) Inserted lights scatter in the water and reach the seabed. (d) Surfaces and water are stylized separately, and the close-ups show each component alone, because the reconstruction keeps them apart.}
  \label{fig:applications}

\end{figure*}
\begin{figure*}[t]
  \centering
  \includegraphics[width=\textwidth]{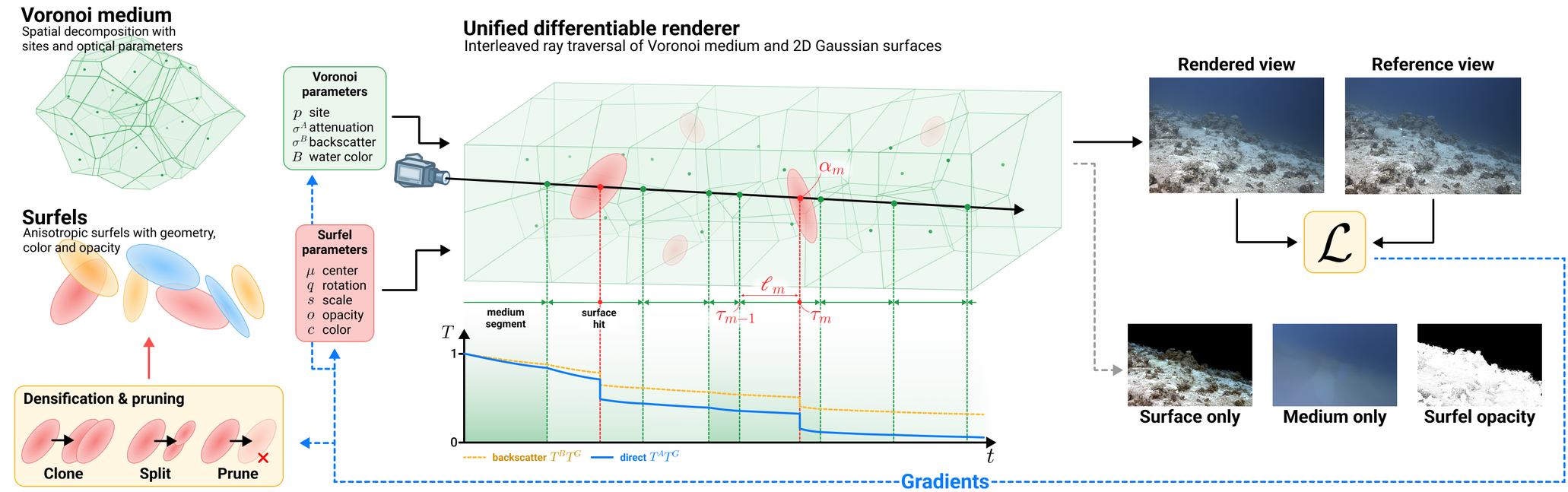}
  \caption{Overview of \our{}. Surfels (left, bottom) hold the surfaces and Voronoi cells (left, top) hold the water. One differentiable ray tracer visits surfel hits and cell boundaries in their order along the ray. It tracks one transmittance for the direct signal and one for the backscatter (bottom plot). Each medium segment lowers them at its own rates, and each surfel hit lowers both. The tracer also renders the surface alone, the medium alone, and the surfel opacity. The loss compares the rendered view with the photo. Its gradients update both representations, and densification and pruning adapt the surfels.}
  \label{fig:overview}
\end{figure*}
\begin{figure*}[t]
  \centering
  \captionof{table}{Novel view synthesis and separation on SeaThru-NeRF. Values are means over four scenes, with standard deviations over three seeds at the final iteration. LPIPS uses VGG. Timings use an RTX 4090, excluding RadFoam training. Size counts the parameters of the trained model at four bytes each, not the checkpoint file. Separation evaluates surface-only opacity against seabed and water masks, and requires a separate medium. Throughout the result tables, colors scale each column from yellow (worst) through orange to red (best), and the best value is bold. Gray marks unavailable values or $^\dagger$author-reported results, excluded from the scale. Per-scene results are in \cref{sec:suppl_per_scene}.}
  \label{tab:seathru}
  \small
\begin{tabular}{l ccc ccc c}
\toprule
 & \multicolumn{3}{c}{Image quality} & \multicolumn{3}{c}{Efficiency} & \multicolumn{1}{c}{Disentanglement} \\
\cmidrule(lr){2-4} \cmidrule(lr){5-7} \cmidrule(lr){8-8}
Method & PSNR$\uparrow$ & SSIM$\uparrow$ & LPIPS$\downarrow$ & FPS$\uparrow$ & Train (min)$\downarrow$ & Size (MB)$\downarrow$ & Separation$\uparrow$ \\
\midrule
SeaThru-NeRF (JAX) & \cellcolor{resultOrange!50.00!resultYellow}27.11\std{0.05} & \cellcolor{resultOrange!22.22!resultYellow}.845\std{.003} & \cellcolor{resultOrange!25.00!resultYellow}.341\std{.003} & \cellcolor{resultOrange!22.22!resultYellow}0.10\std{0.00} & \cellcolor{resultOrange!25.00!resultYellow}614\std{4} & \cellcolor{resultRed!100.00!resultOrange}\textbf{3}\std{0} & \cellcolor{resultRed!33.33!resultOrange}.762\std{.058} \\
SeaThru-NeRF (Nerfstudio) & \cellcolor{resultOrange!100.00!resultYellow}28.31\std{0.20} & \cellcolor{resultOrange!88.89!resultYellow}.914\std{.000} & \cellcolor{resultRed!50.00!resultOrange}.188\std{.002} & \cellcolor{resultOrange!0.00!resultYellow}0.06\std{0.00} & \cellcolor{resultOrange!0.00!resultYellow}722\std{3} & \cellcolor{resultOrange!44.44!resultYellow}282\std{0} & \cellcolor{resultRed!66.67!resultOrange}.851\std{.031} \\
Zip-NeRF & \cellcolor{resultRed!25.00!resultOrange}29.62\std{0.09} & \cellcolor{resultRed!11.11!resultOrange}.923\std{.000} & \cellcolor{resultRed!100.00!resultOrange}\textbf{.173}\std{.002} & \cellcolor{resultOrange!44.44!resultYellow}0.13\std{0.01} & \cellcolor{resultOrange!50.00!resultYellow}160\std{2} & \cellcolor{resultOrange!22.22!resultYellow}310\std{0} & \cellcolor{resultNA}n/a \\
3DGS & \cellcolor{resultOrange!25.00!resultYellow}25.91\std{0.09} & \cellcolor{resultOrange!44.44!resultYellow}.872\std{.002} & \cellcolor{resultOrange!50.00!resultYellow}.217\std{.000} & \cellcolor{resultRed!100.00!resultOrange}\textbf{346}\std{6} & \cellcolor{resultRed!25.00!resultOrange}13.2\std{0.4} & \cellcolor{resultRed!77.78!resultOrange}131\std{1} & \cellcolor{resultNA}n/a \\
RadFoam & \cellcolor{resultOrange!0.00!resultYellow}23.52\std{0.78} & \cellcolor{resultOrange!0.00!resultYellow}.810\std{.022} & \cellcolor{resultOrange!0.00!resultYellow}.359\std{.022} & \cellcolor{resultOrange!66.67!resultYellow}33\std{1} & \cellcolor{resultNA}n/a & \cellcolor{resultOrange!0.00!resultYellow}831\std{0} & \cellcolor{resultNA}n/a \\
SeaSplat & \cellcolor{resultOrange!75.00!resultYellow}27.22\std{0.16} & \cellcolor{resultOrange!66.67!resultYellow}.883\std{.001} & \cellcolor{resultOrange!100.00!resultYellow}.191\std{.003} & \cellcolor{resultOrange!88.89!resultYellow}115\std{10} & \cellcolor{resultOrange!75.00!resultYellow}49.4\std{1.3} & \cellcolor{resultRed!33.33!resultOrange}164\std{21} & \cellcolor{resultOrange!33.33!resultYellow}.545\std{.001} \\
WaterSplatting & \cellcolor{resultRed!25.00!resultOrange}29.62\std{0.11} & \cellcolor{resultRed!33.33!resultOrange}.928\std{.000} & \cellcolor{resultRed!25.00!resultOrange}.190\std{.001} & \cellcolor{resultRed!33.33!resultOrange}154\std{5} & \cellcolor{resultRed!50.00!resultOrange}8.9\std{3.6} & \cellcolor{resultOrange!88.89!resultYellow}239\std{1} & \cellcolor{resultOrange!100.00!resultYellow}.624\std{.012} \\
WaterSplatting + PPISP & \cellcolor{resultRed!75.00!resultOrange}30.71\std{0.02} & \cellcolor{resultRed!77.78!resultOrange}.930\std{.000} & \cellcolor{resultRed!75.00!resultOrange}.185\std{.000} & \cellcolor{resultRed!55.56!resultOrange}164\std{2} & \cellcolor{resultRed!100.00!resultOrange}\textbf{7.4}\std{0.1} & \cellcolor{resultOrange!66.67!resultYellow}243\std{0} & \cellcolor{resultOrange!66.67!resultYellow}.616\std{.016} \\
Plenodium & \cellcolor{resultRed!50.00!resultOrange}30.68\std{0.10} & \cellcolor{resultRed!100.00!resultOrange}\textbf{.931}\std{.000} & \cellcolor{resultRed!75.00!resultOrange}.185\std{.000} & \cellcolor{resultRed!77.78!resultOrange}211\std{4} & \cellcolor{resultRed!75.00!resultOrange}8.7\std{2.7} & \cellcolor{resultRed!11.11!resultOrange}234\std{0} & \cellcolor{resultOrange!0.00!resultYellow}.505\std{.002} \\
\rowcolor{resultNA}
OceanSplat$^\dagger$ & 30.71 & .935 & n/a & n/a & n/a & n/a & n/a \\
\midrule
\our{} & \cellcolor{resultRed!100.00!resultOrange}\textbf{30.75}\std{0.01} & \cellcolor{resultRed!55.56!resultOrange}.929\std{.001} & \cellcolor{resultOrange!75.00!resultYellow}.202\std{.001} & \cellcolor{resultRed!11.11!resultOrange}120\std{0} & \cellcolor{resultOrange!100.00!resultYellow}28.7\std{0.3} & \cellcolor{resultRed!55.56!resultOrange}159\std{0} & \cellcolor{resultRed!100.00!resultOrange}\textbf{.901}\std{.001} \\
\bottomrule
\end{tabular}

\end{figure*}
\begin{figure}[t]
  \centering
  \includegraphics[width=\columnwidth]{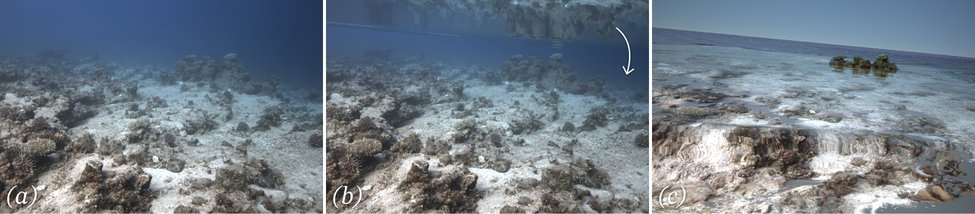}
  \caption{Lowering the water level of a reconstructed scene. The medium of every ray is clipped below a wavy surface that reflects and refracts. (a) The reconstruction. (b) The surface enters the view, with the seabed reflected on its underside. (c) The seabed is exposed. The waves and the sky are visual effects.}
  \label{fig:draining}
\end{figure}

\begin{figure*}[t]
  \centering
  \includegraphics[width=\textwidth]{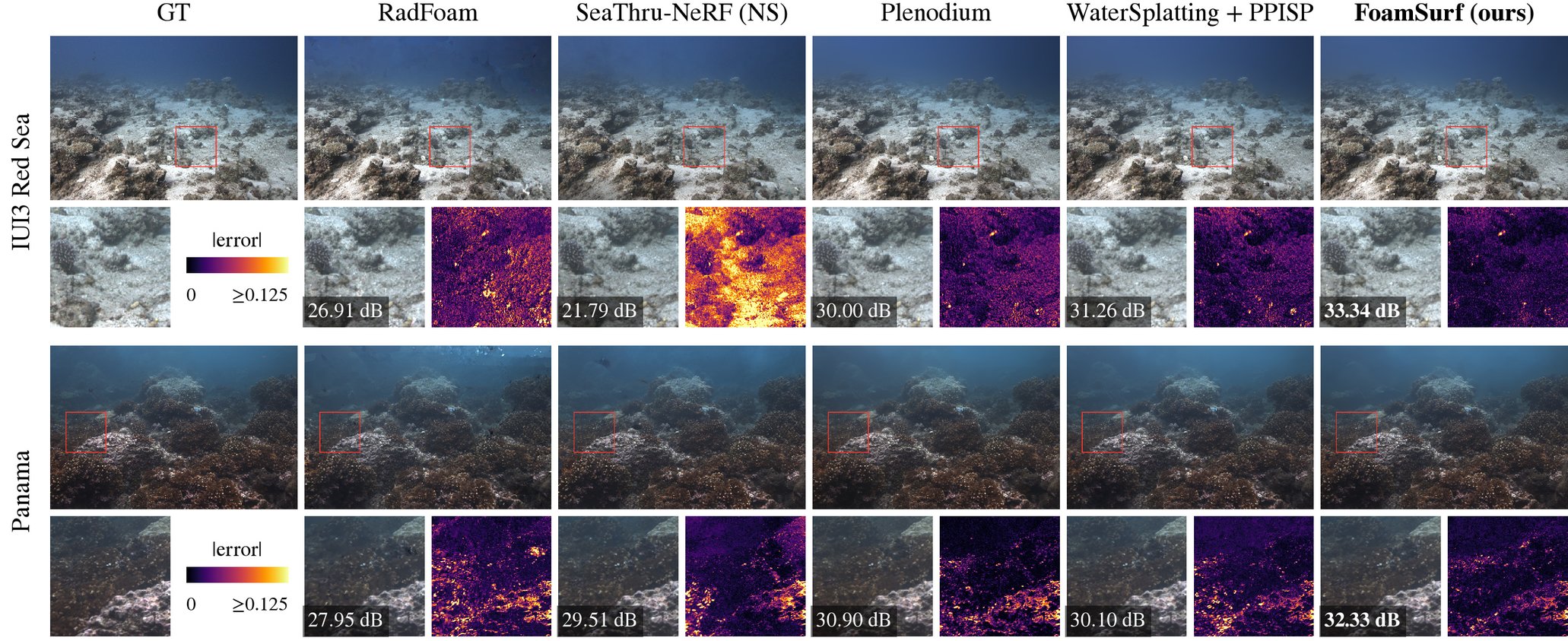}
  \caption{Novel views on two scenes of the SeaThru-NeRF dataset. Under each full view, the crop in the red box is shown next to its absolute error against the photo. All error maps share one color scale. The value in each crop is its PSNR. \our{} has the lowest error in both crops.}
  \label{fig:nvs}
\end{figure*}
\begin{figure*}[t]
\centering
\centering
  \includegraphics[width=\textwidth]{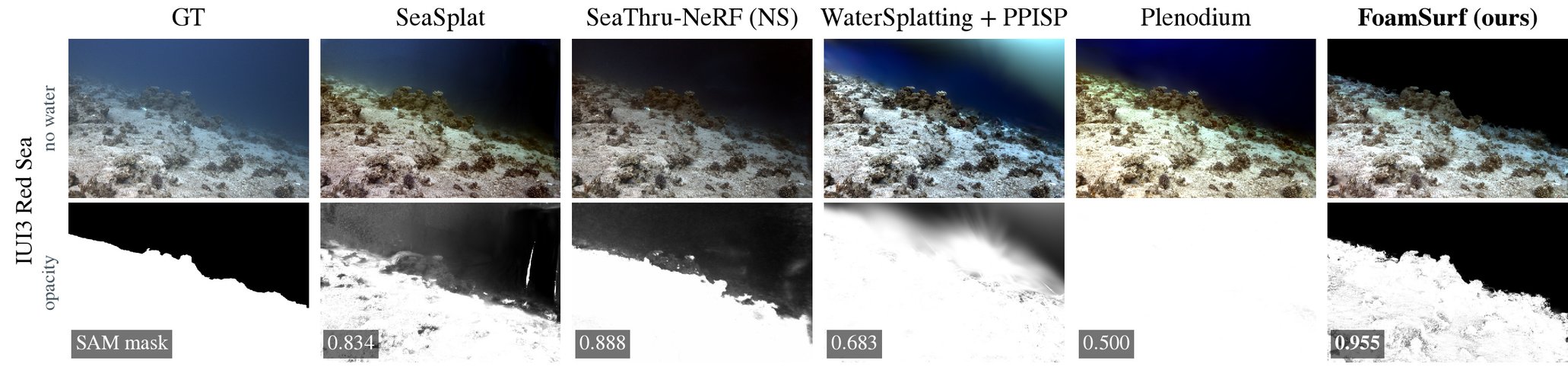}
  \caption{Surface–medium separation on IUI3 Red Sea. Water-free renders appear above surface-only opacity maps, with per-view separation scores. White denotes seabed in the evaluation mask and opaque surfaces in predictions. \our{} retains seabed structure while leaving open water nearly free of surfels. An additional scene is shown in \cref{sec:suppl_qualitative}.}
  \label{fig:decomposition}

\end{figure*}
\newcommand{\foamsurftablesix}{%
\begin{figure}[t]
\centering
\captionof{table}{Synthetic reconstruction PSNR on a black background, evaluated against full-scene, object-only, and cloud-only ground truth. Means and standard deviations over three seeds. Baselines use our initialization. Additional metrics and initialization comparisons are in \cref{sec:suppl_synthetic}.}
  \label{tab:synthetic}
  \resizebox{\linewidth}{!}{
\begin{tabular}{l ccc ccc}
\toprule
 & \multicolumn{3}{c}{Lego} & \multicolumn{3}{c}{Ficus} \\
\cmidrule(lr){2-4} \cmidrule(lr){5-7}
 & Full & Object & Cloud & Full & Object & Cloud \\
Method & PSNR$\uparrow$ & PSNR$\uparrow$ & PSNR$\uparrow$ & PSNR$\uparrow$ & PSNR$\uparrow$ & PSNR$\uparrow$ \\
\midrule
WaterSplatting & \cellcolor{resultOrange!0.00!resultYellow}32.90\std{0.06} & \cellcolor{resultOrange!0.00!resultYellow}15.44\std{0.21} & \cellcolor{resultOrange!0.00!resultYellow}19.47\std{0.00} & \cellcolor{resultOrange!0.00!resultYellow}24.10\std{4.18} & \cellcolor{resultOrange!0.00!resultYellow}18.07\std{3.56} & \cellcolor{resultOrange!0.00!resultYellow}21.71\std{0.00} \\
SeaSplat & \cellcolor{resultRed!100.00!resultOrange}\textbf{35.44}\std{0.12} & \cellcolor{resultOrange!100.00!resultYellow}19.02\std{0.05} & \cellcolor{resultOrange!0.00!resultYellow}19.47\std{0.00} & \cellcolor{resultOrange!100.00!resultYellow}31.69\std{0.10} & \cellcolor{resultOrange!100.00!resultYellow}21.31\std{0.30} & \cellcolor{resultOrange!0.00!resultYellow}21.71\std{0.00} \\
\midrule
\our{} & \cellcolor{resultOrange!100.00!resultYellow}34.13\std{0.08} & \cellcolor{resultRed!100.00!resultOrange}\textbf{34.77}\std{0.09} & \cellcolor{resultRed!100.00!resultOrange}\textbf{38.04}\std{0.03} & \cellcolor{resultRed!100.00!resultOrange}\textbf{32.88}\std{0.02} & \cellcolor{resultRed!100.00!resultOrange}\textbf{32.89}\std{0.08} & \cellcolor{resultRed!100.00!resultOrange}\textbf{45.12}\std{4.02} \\
\bottomrule
\end{tabular}
}
\end{figure}
}
\newcommand{\foamsurfseven}{%
\begin{figure}[t]
\centering
\includegraphics[width=\linewidth]{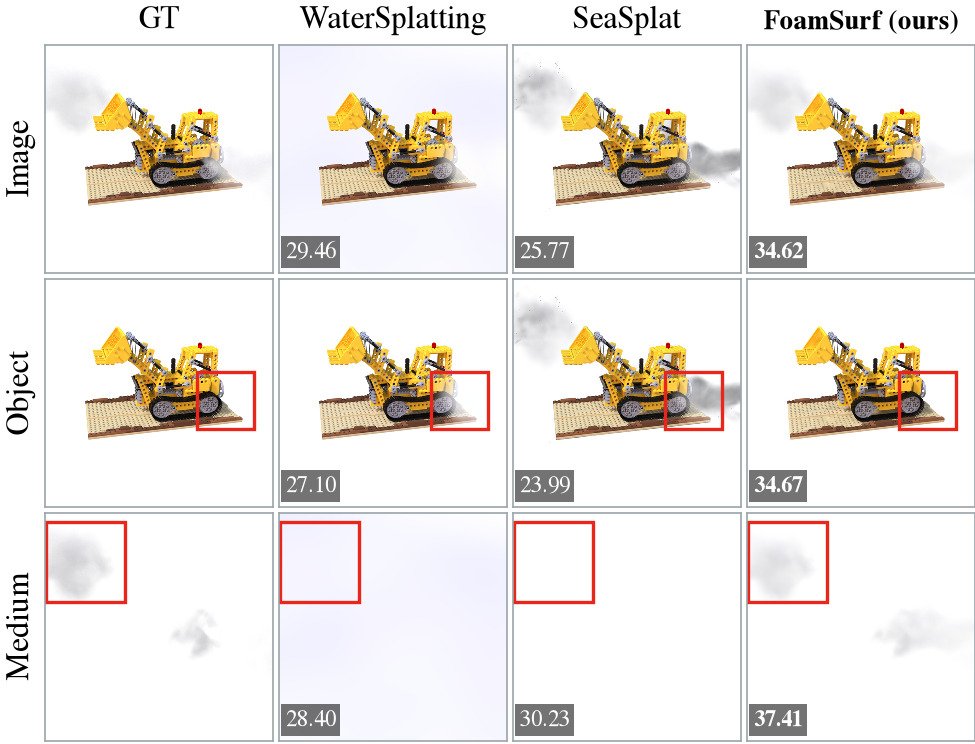}
  \caption{Lego with three surrounding clouds. Rows show the full scene, object, and medium, with PSNR against separate ground truth. All methods share initial surface points. Red boxes highlight cloud structure retained in baseline surfaces but assigned to the medium by \our{}. The white background differs from \cref{tab:synthetic}.}
  \label{fig:synthetic}
\end{figure}
\begin{figure}[t]
\centering
\captionof{table}{Foggy Garden restoration against clear reference images. Means and standard deviations over two seeds for \our{} and WaterSplatting, three otherwise. Hard-fog parameters are fitted to the published figure (\cref{sec:supp_setup}).}
  \label{tab:garden}
  \resizebox{\linewidth}{!}{
\begin{tabular}{l ccc ccc}
\toprule
 & \multicolumn{3}{c}{Easy fog} & \multicolumn{3}{c}{Hard fog} \\
\cmidrule(lr){2-4} \cmidrule(lr){5-7}
Method & PSNR$\uparrow$ & SSIM$\uparrow$ & LPIPS$\downarrow$ & PSNR$\uparrow$ & SSIM$\uparrow$ & LPIPS$\downarrow$ \\
\midrule
SeaThru-NeRF (JAX) & \cellcolor{resultOrange!80.00!resultYellow}15.20\std{0.01} & \cellcolor{resultOrange!40.00!resultYellow}.341\std{.003} & \cellcolor{resultOrange!0.00!resultYellow}.622\std{.006} & \cellcolor{resultOrange!80.00!resultYellow}13.97\std{0.55} & \cellcolor{resultOrange!40.00!resultYellow}.289\std{.025} & \cellcolor{resultOrange!0.00!resultYellow}.740\std{.033} \\
SeaThru-NeRF (Nerfstudio) & \cellcolor{resultOrange!0.00!resultYellow}12.89\std{2.47} & \cellcolor{resultOrange!0.00!resultYellow}.288\std{.040} & \cellcolor{resultOrange!40.00!resultYellow}.594\std{.023} & \cellcolor{resultOrange!0.00!resultYellow}10.02\std{0.14} & \cellcolor{resultOrange!0.00!resultYellow}.211\std{.007} & \cellcolor{resultOrange!40.00!resultYellow}.650\std{.013} \\
SeaSplat & \cellcolor{resultOrange!40.00!resultYellow}13.96\std{0.03} & \cellcolor{resultRed!20.00!resultOrange}.634\std{.002} & \cellcolor{resultRed!100.00!resultOrange}\textbf{.315}\std{.002} & \cellcolor{resultOrange!40.00!resultYellow}10.44\std{0.13} & \cellcolor{resultOrange!80.00!resultYellow}.380\std{.010} & \cellcolor{resultRed!60.00!resultOrange}.487\std{.009} \\
WaterSplatting & \cellcolor{resultRed!60.00!resultOrange}19.69\std{0.14} & \cellcolor{resultRed!100.00!resultOrange}\textbf{.696}\std{.001} & \cellcolor{resultRed!60.00!resultOrange}.318\std{.001} & \cellcolor{resultRed!60.00!resultOrange}15.46\std{0.00} & \cellcolor{resultRed!60.00!resultOrange}.495\std{.010} & \cellcolor{resultRed!20.00!resultOrange}.489\std{.001} \\
Plenodium & \cellcolor{resultRed!20.00!resultOrange}17.83\std{0.13} & \cellcolor{resultOrange!80.00!resultYellow}.601\std{.002} & \cellcolor{resultOrange!80.00!resultYellow}.351\std{.003} & \cellcolor{resultRed!20.00!resultOrange}15.29\std{0.05} & \cellcolor{resultRed!20.00!resultOrange}.423\std{.002} & \cellcolor{resultOrange!80.00!resultYellow}.522\std{.001} \\
\midrule
\our{} & \cellcolor{resultRed!100.00!resultOrange}\textbf{19.89}\std{0.29} & \cellcolor{resultRed!60.00!resultOrange}.657\std{.006} & \cellcolor{resultRed!20.00!resultOrange}.349\std{.003} & \cellcolor{resultRed!100.00!resultOrange}\textbf{16.81}\std{0.60} & \cellcolor{resultRed!100.00!resultOrange}\textbf{.584}\std{.015} & \cellcolor{resultRed!100.00!resultOrange}\textbf{.391}\std{.005} \\
\bottomrule
\end{tabular}
}
\end{figure}
}
\newcommand{\foamsurfinsertion}{%
\begin{figure}[t]
\centering
\captionof{table}{Object insertion against ground truth. An unlit sphere is added at a fixed position, and every method renders the same cameras. Scene: the reconstruction without the sphere. Object: its full silhouette. Seed 42, the baselines from our initial points. Protocol and further metrics in \cref{sec:suppl_insertion}.}
  \label{tab:insertion}
  \resizebox{\linewidth}{!}{
\begin{tabular}{l cc cc}
\toprule
 & \multicolumn{2}{c}{Lego} & \multicolumn{2}{c}{Ficus} \\
\cmidrule(lr){2-3} \cmidrule(lr){4-5}
 & Scene & Object & Scene & Object \\
Method & PSNR$\uparrow$ & PSNR$\uparrow$ & PSNR$\uparrow$ & PSNR$\uparrow$ \\
\midrule
WaterSplatting & \cellcolor{resultOrange!0.00!resultYellow}32.85 & \cellcolor{resultOrange!0.00!resultYellow}17.25 & \cellcolor{resultOrange!0.00!resultYellow}19.27 & \cellcolor{resultOrange!0.00!resultYellow}20.82 \\
SeaSplat & \cellcolor{resultRed!100.00!resultOrange}\textbf{35.30} & \cellcolor{resultOrange!100.00!resultYellow}33.09 & \cellcolor{resultOrange!100.00!resultYellow}31.58 & \cellcolor{resultOrange!100.00!resultYellow}26.78 \\
\midrule
\our{} & \cellcolor{resultOrange!100.00!resultYellow}34.22 & \cellcolor{resultRed!100.00!resultOrange}\textbf{33.39} & \cellcolor{resultRed!100.00!resultOrange}\textbf{32.86} & \cellcolor{resultRed!100.00!resultOrange}\textbf{30.22} \\
\bottomrule
\end{tabular}
}
\end{figure}
}
\newcommand{\foamsurfeight}{%
\begin{figure}[t]
\centering
\includegraphics[width=\linewidth]{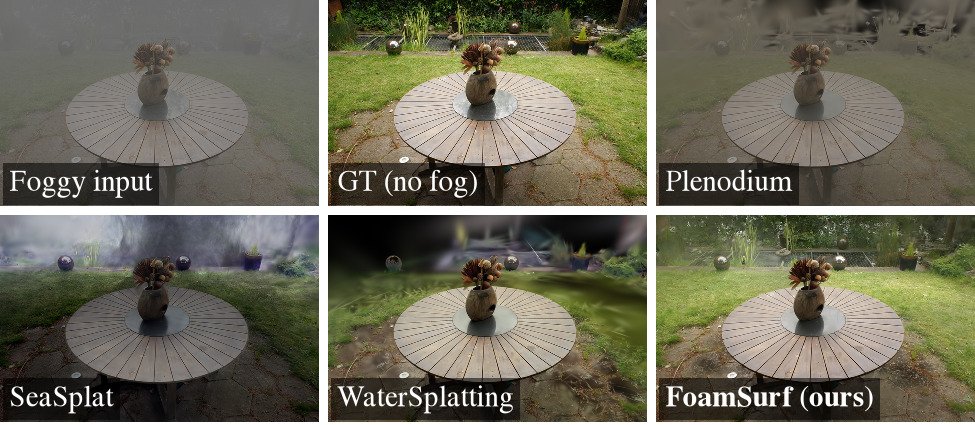}
  \caption{Hard-fog restoration with the medium removed. \our{} recovers background vegetation where baselines show residual haze, darkening or blur. See \cref{tab:garden} for quantitative results.}
  \label{fig:garden}
\end{figure}
\begin{figure}[t]
\centering
\captionof{table}{Ablation on SeaThru-NeRF, removing or replacing one component per variant. Values are means over four scenes and two seeds at the final iteration, with standard deviations in small numbers. $\Delta$PSNR is relative to the full method. Water is the surface-only opacity on water pixels, one component of the separation score of \cref{tab:seathru}.}
  \label{tab:ablation}
  \resizebox{\linewidth}{!}{
\begin{tabular}{l ccc cc}
\toprule
 & \multicolumn{3}{c}{Image quality} & \multicolumn{2}{c}{Disentanglement} \\
\cmidrule(lr){2-4} \cmidrule(lr){5-6}
Variant & PSNR$\uparrow$ & $\Delta$PSNR & SSIM$\uparrow$ & Water$\downarrow$ & Separation$\uparrow$ \\
\midrule
\our{} (full) & \cellcolor{resultRed!100.00!resultOrange}30.76\std{0.02} & \cellcolor{resultRed!100.00!resultOrange}0.00 & \cellcolor{resultRed!100.00!resultOrange}.929\std{.000} & \cellcolor{resultRed!20.00!resultOrange}.005\std{.001} & \cellcolor{resultRed!14.29!resultOrange}.901\std{.001} \\
\midrule
w/o medium (surfels only) & \cellcolor{resultOrange!0.00!resultYellow}29.03\std{0.45} & \cellcolor{resultOrange!0.00!resultYellow}$-$1.73 & \cellcolor{resultOrange!0.00!resultYellow}.915\std{.002} & \cellcolor{resultOrange!0.00!resultYellow}.977\std{.000} & \cellcolor{resultOrange!0.00!resultYellow}.506\std{.001} \\
homogeneous medium instead of cells & \cellcolor{resultRed!42.86!resultOrange}30.53\std{0.02} & \cellcolor{resultRed!33.33!resultOrange}$-$0.23 & \cellcolor{resultRed!100.00!resultOrange}.929\std{.000} & \cellcolor{resultOrange!40.00!resultYellow}.214\std{.011} & \cellcolor{resultOrange!28.57!resultYellow}.857\std{.005} \\
w/o separation priors & \cellcolor{resultRed!14.29!resultOrange}30.52\std{0.17} & \cellcolor{resultRed!33.33!resultOrange}$-$0.23 & \cellcolor{resultRed!50.00!resultOrange}.928\std{.000} & \cellcolor{resultOrange!80.00!resultYellow}.030\std{.002} & \cellcolor{resultOrange!57.14!resultYellow}.883\std{.001} \\
\midrule
w/o directional scattering & \cellcolor{resultRed!71.43!resultOrange}30.64\std{0.04} & \cellcolor{resultRed!66.67!resultOrange}$-$0.12 & \cellcolor{resultRed!100.00!resultOrange}.929\std{.000} & \cellcolor{resultRed!20.00!resultOrange}.005\std{.002} & \cellcolor{resultRed!42.86!resultOrange}.902\std{.000} \\
w/o camera model & \cellcolor{resultOrange!28.57!resultYellow}29.28\std{0.23} & \cellcolor{resultOrange!33.33!resultYellow}$-$1.48 & \cellcolor{resultOrange!100.00!resultYellow}.926\std{.001} & \cellcolor{resultRed!60.00!resultOrange}.004\std{.000} & \cellcolor{resultRed!71.43!resultOrange}.910\std{.001} \\
camera model w/o novel-view interpolation & \cellcolor{resultOrange!57.14!resultYellow}30.22\std{0.01} & \cellcolor{resultOrange!66.67!resultYellow}$-$0.53 & \cellcolor{resultRed!50.00!resultOrange}.928\std{.000} & \cellcolor{resultRed!20.00!resultOrange}.005\std{.000} & \cellcolor{resultOrange!85.71!resultYellow}.900\std{.001} \\
w/o multi-view depth terms & \cellcolor{resultOrange!85.71!resultYellow}30.30\std{0.22} & \cellcolor{resultOrange!100.00!resultYellow}$-$0.45 & \cellcolor{resultOrange!50.00!resultYellow}.924\std{.001} & \cellcolor{resultRed!100.00!resultOrange}.002\std{.001} & \cellcolor{resultRed!100.00!resultOrange}.921\std{.004} \\
\bottomrule
\end{tabular}
}
\end{figure}
}
\afterpage{\afterpage{\afterpage{\afterpage{\afterpage{%
\begin{figure}[t]
\centering
\captionof{table}{Synthetic reconstruction PSNR on a black background, evaluated against full-scene, object-only, and cloud-only ground truth. Means and standard deviations over three seeds. Baselines use our initialization. Additional metrics and initialization comparisons are in \cref{sec:suppl_synthetic}.}
  \label{tab:synthetic}
  \resizebox{\linewidth}{!}{
\begin{tabular}{l ccc ccc}
\toprule
 & \multicolumn{3}{c}{Lego} & \multicolumn{3}{c}{Ficus} \\
\cmidrule(lr){2-4} \cmidrule(lr){5-7}
 & Full & Object & Cloud & Full & Object & Cloud \\
Method & PSNR$\uparrow$ & PSNR$\uparrow$ & PSNR$\uparrow$ & PSNR$\uparrow$ & PSNR$\uparrow$ & PSNR$\uparrow$ \\
\midrule
WaterSplatting & \cellcolor{resultOrange!0.00!resultYellow}32.90\std{0.06} & \cellcolor{resultOrange!0.00!resultYellow}15.44\std{0.21} & \cellcolor{resultOrange!0.00!resultYellow}19.47\std{0.00} & \cellcolor{resultOrange!0.00!resultYellow}24.10\std{4.18} & \cellcolor{resultOrange!0.00!resultYellow}18.07\std{3.56} & \cellcolor{resultOrange!0.00!resultYellow}21.71\std{0.00} \\
SeaSplat & \cellcolor{resultRed!100.00!resultOrange}\textbf{35.44}\std{0.12} & \cellcolor{resultOrange!100.00!resultYellow}19.02\std{0.05} & \cellcolor{resultOrange!0.00!resultYellow}19.47\std{0.00} & \cellcolor{resultOrange!100.00!resultYellow}31.69\std{0.10} & \cellcolor{resultOrange!100.00!resultYellow}21.31\std{0.30} & \cellcolor{resultOrange!0.00!resultYellow}21.71\std{0.00} \\
\midrule
\our{} & \cellcolor{resultOrange!100.00!resultYellow}34.13\std{0.08} & \cellcolor{resultRed!100.00!resultOrange}\textbf{34.77}\std{0.09} & \cellcolor{resultRed!100.00!resultOrange}\textbf{38.04}\std{0.03} & \cellcolor{resultRed!100.00!resultOrange}\textbf{32.88}\std{0.02} & \cellcolor{resultRed!100.00!resultOrange}\textbf{32.89}\std{0.08} & \cellcolor{resultRed!100.00!resultOrange}\textbf{45.12}\std{4.02} \\
\bottomrule
\end{tabular}
}
\end{figure}

}}}}}
\afterpage{\afterpage{\afterpage{\afterpage{\afterpage{\afterpage{\afterpage{%
\begin{figure}[t]
\centering
\captionof{table}{Object insertion against ground truth. An unlit sphere is added at a fixed position, and every method renders the same cameras. Scene: the reconstruction without the sphere. Object: its full silhouette. Seed 42, the baselines from our initial points. Protocol and further metrics in \cref{sec:suppl_insertion}.}
  \label{tab:insertion}
  \resizebox{\linewidth}{!}{
\begin{tabular}{l cc cc}
\toprule
 & \multicolumn{2}{c}{Lego} & \multicolumn{2}{c}{Ficus} \\
\cmidrule(lr){2-3} \cmidrule(lr){4-5}
 & Scene & Object & Scene & Object \\
Method & PSNR$\uparrow$ & PSNR$\uparrow$ & PSNR$\uparrow$ & PSNR$\uparrow$ \\
\midrule
WaterSplatting & \cellcolor{resultOrange!0.00!resultYellow}32.85 & \cellcolor{resultOrange!0.00!resultYellow}17.25 & \cellcolor{resultOrange!0.00!resultYellow}19.27 & \cellcolor{resultOrange!0.00!resultYellow}20.82 \\
SeaSplat & \cellcolor{resultRed!100.00!resultOrange}\textbf{35.30} & \cellcolor{resultOrange!100.00!resultYellow}33.09 & \cellcolor{resultOrange!100.00!resultYellow}31.58 & \cellcolor{resultOrange!100.00!resultYellow}26.78 \\
\midrule
\our{} & \cellcolor{resultOrange!100.00!resultYellow}34.22 & \cellcolor{resultRed!100.00!resultOrange}\textbf{33.39} & \cellcolor{resultRed!100.00!resultOrange}\textbf{32.86} & \cellcolor{resultRed!100.00!resultOrange}\textbf{30.22} \\
\bottomrule
\end{tabular}
}
\end{figure}

}}}}}}}

\begin{abstract}
Scenes inside a scattering medium contain solid surfaces, such as the seabed or rocks, and a semi-transparent medium, such as water, fog, or smoke. Recent Gaussian-based methods produce realistic images of scenes in scattering media, but traces of the medium can remain visible in surface-only renders. Neural fields separate surfaces and medium more accurately, but they are slow to train and difficult to edit. Both approaches evaluate the medium only along camera rays, so its effects along reflected and refracted rays are not modeled. These limitations prevent reliable surface recovery, independent editing of surfaces and the medium, and consistent rendering of inserted objects. We present \our{}, a hybrid representation that models surfaces with planar Gaussian surfels and the medium with space-filling Voronoi cells. A single differentiable ray tracer composites both along arbitrary rays, while training priors keep them separate without segmentation masks. Across real underwater scenes, synthetic scenes, and dense fog, \our{} achieves image quality on par with the state of the art and the most accurate surface–medium separation among the compared methods. The recovered components can be removed, edited, or stylized independently, and the same reconstruction supports inserted mesh objects with attenuation and scattering along both direct and reflected or refracted rays. Together, these capabilities enable scene restoration, editing, and composition within a single reconstructed model.\\
Project page: \url{https://grzegorzwilczynski.com/foamsurf/}
\end{abstract}
\vspace{-0.9cm}
\section{Introduction}
\label{sec:intro}

Photographs of a scene inside a scattering medium carry surface light and scattered light in the same pixels.
Recovering the surfaces and the medium as separate parts is therefore non-trivial.
A reconstruction that fails at this renders novel views, yet neither component can be changed without disturbing the other.
Editing, restyling and inserting objects stay out of reach (\cref{fig:teaser}).

Neural underwater methods support novel views and medium removal~\cite{levy2023seathru,sethuraman2023waternerf}, while Gaussian-based approaches enable real-time rendering~\cite{li2025watersplatting,kweon2026oceansplat,yang2025seasplat,liu2024aquatic}.
Many methods also describe the medium through ray direction, image-space parameters or attributes of surface primitives~\cite{levy2023seathru,li2025watersplatting,yang2025seasplat}.
These formulations do not directly provide a spatial medium for tracing paths that change direction inside the scene, such as reflected and refracted rays.
Useful scene control requires both editable components and a renderer that follows their interaction along such paths.

We introduce \our{}, built around a unified differentiable ray tracer that learns surfaces and a spatial medium from posed photographs.
Planar Gaussian surfels describe surfaces, while Voronoi cells give the medium its own spatial representation.
At each surface hit, the tracer accounts for the intervening water, combining occlusion with attenuation and backscatter.
Differentiating this joint rendering process lets the photographs constrain both components within one explicit scene.
The same mechanism connects reconstruction to scene composition, because a ray can start inside the medium and cross it after reflection or refraction.
Inserted meshes are rendered alongside the recovered scene, with their appearance affected by the water crossed along each path (\cref{fig:applications}).

Learning useful components also requires preventing incorrect assignments of appearance.
Surface color assigned to the medium causes \emph{medium leakage}, while surface primitives that absorb backscatter become \emph{haze surfels}.
Either error can remain hidden in a matching image but becomes visible when the components are edited separately.
Our separation priors discourage both errors without segmentation masks (\cref{sec:priors}).
Directional scattering and a camera model account for view-dependent water brightness and photographic variation.
Together, these constraints support independent component removal, editing, and style transfer.

On real underwater scenes, \our{} combines competitive image quality with the highest mean separation score and real-time rendering.
The object and the medium each achieve the highest reconstruction PSNR on our synthetic scenes.
The strongest restoration PSNR under dense fog also demonstrates applicability beyond underwater captures.

Our contributions are:
\begin{itemize}
  \item A hybrid representation and differentiable ray tracer for surface–medium reconstruction, supporting scene composition along primary and secondary rays.
  \item Separation priors that recover editable surfaces and medium without supervision of the split, supporting independent editing, style transfer and meshes seen along reflected and refracted rays (\cref{fig:applications}).
  \item The most accurate surface--medium separation among the compared methods, at image quality on par with the strongest baselines.
\end{itemize}

\section{Related work}
\label{sec:related}

\textbf{Neural fields in scattering media.} Extensions of neural radiance fields~\cite{mildenhall2021nerf} recover underwater views and are able to remove the medium~\cite{levy2023seathru,sethuraman2023waternerf}.
They incorporate attenuation and backscatter into image formation rather than treating the water's appearance as part of the surface.
SeaThru-NeRF uses direction-dependent medium parameters constant along each ray, while fog methods learn spatial fields~\cite{ramazzina2023scatternerf,chen2023dehazenerf}, including editable fog density~\cite{ramazzina2023scatternerf}.
Neural Relightable Participating Media~\cite{zheng2021neural} reconstructs under known lighting conditions and supports material edits and mesh composition.
The field is first resampled onto a voxel grid and rendered offline.
These approaches establish separation, but editing them requires queries into the field or the same resampling.
\our{} instead reconstructs directly editable surface and medium primitives, rendered together in real time.

\textbf{Gaussian primitives in scattering media.} Extensions of Gaussian splatting~\cite{kerbl20233d} integrate a ray-conditioned medium between primitives~\cite{li2025watersplatting,kweon2026oceansplat}, modify rendered color and depth~\cite{yang2025seasplat}, or attach optical proxies to Gaussians~\cite{zhang2026waterclear}.
Spatial alternatives place the medium coefficients in space, as voxels~\cite{xing2025uw}, a tri-plane~\cite{jiang2026tri}, or the spherical harmonics of Plenodium~\cite{wu2025plenodium}.
The first two compose the medium with the rasterized image, while the third queries it between depth-sorted primitives.
Rasterization still ties these queries to camera rays, so inserted geometry and secondary rays fall outside them.
Color and depth priors, multi-view consistency, and pruning help separate the learned components~\cite{yang2025seasplat,zhang2026waterclear,kweon2026oceansplat,xing2025uw}.
The remaining challenge is to combine that separation with rendering along paths introduced by scene edits.
\our{} combines separation priors with joint surfel and cell traversal, allowing both components to be rendered together after scene changes.

\textbf{Ray-traced primitives and volumetric partitions.} 2D Gaussian splatting~\cite{huang20242d} computes ray--surfel intersections within rasterization, while ray-traced Gaussians~\cite{moenneloccoz20243d} support mesh insertion, reflection, and refraction.
Volumetric representations include Voronoi cells with exact integration of piecewise-constant radiance and density~\cite{govindarajan2025radiant}, and primitives supporting volumetric path tracing with multiple scattering~\cite{condor2025dont}.
These methods provide rendering capabilities, but do not jointly recover distinct surfaces and water from underwater photographs.
\our{} connects this reconstruction task to scene composition by tracing both components under the underwater image formation model~\cite{akkaynak2018revised}.
Separate direct attenuation and backscatter govern their contributions, while a shared tracer evaluates visibility through the hybrid scene.
This preserves distinct surface and medium parameters during joint reconstruction.
\section{Method}
\label{sec:method}

Given photos with known camera poses, \our{} reconstructs a hybrid scene of surfels and medium cells (\cref{fig:overview}).
A differentiable ray tracer renders both components together, while separation priors guide their joint reconstruction.

\subsection{Scene representation}
\label{sec:prelim}
\label{sec:representation}

\textbf{Image formation in water.} Water attenuates surface light and adds backscatter, with distinct coefficients~\cite{akkaynak2018revised,akkaynak2019sea}.
A surface of color $\mathbf{c}$ at distance $z$ in homogeneous water appears as
\begin{equation}
  \mathbf{C} = \mathbf{c}\odot e^{-\boldsymbol{\sigma}^A z} + \mathbf{B}\odot\bigl(\mathbf{1}-e^{-\boldsymbol{\sigma}^B z}\bigr),
  \label{eq:seathru}
\end{equation}
where $\boldsymbol{\sigma}^A$ and $\boldsymbol{\sigma}^B$ control attenuation and backscatter, and $\mathbf{B}$ is the water color at infinite distance.
These quantities are RGB, with products and exponentials applied per channel.

\textbf{Surfaces.} Each planar surfel~\cite{huang20242d} has a center $\boldsymbol{\mu}_i$, a rotation quaternion $\mathbf{q}_i$ and tangent scales $(s_i^u,s_i^v)$.
Its appearance parameters are opacity $o_i$, shape exponent $\kappa_i$ and RGB spherical-harmonic coefficients.
The quaternion defines the tangent axes and normal, while the coefficients encode view-dependent color.
We use a generalized Gaussian kernel~\cite{byrski2026floating}, with $\kappa_i$ controlling its profile.
All these parameters are learned.

\textbf{Medium.} Each cell has a learned site $\mathbf{p}_j$, the point that generates it.
It carries attenuation $\boldsymbol{\sigma}^A_j$, backscatter $\boldsymbol{\sigma}^B_j$ and water color $\mathbf{B}_j$, each with one coefficient per color channel.
A point belongs to its nearest site's Voronoi cell, where the medium parameters are constant.
The outer cells are unbounded, so the medium fills space independently of the surfels.
Joint training learns the sites and optical parameters while keeping the cell count fixed.
This piecewise-constant representation permits exact medium integration within each traversed segment.

\subsection{Exact hybrid ray tracing}
\label{sec:tracing}

A surfel is seen through foreground water and occludes the water behind it.
We therefore composite surface hits and medium segments together, front to back~\cite{kerbl20233d,mildenhall2021nerf}.
Unlike ray-traced primitives in clear air~\cite{moenneloccoz20243d,govindarajan2025radiant}, the tracer interleaves two kinds of events and carries separate transmittances for the direct signal and the backscatter.

\textbf{Events.} We intersect rays with planar polygon proxies~\cite{byrski2025redisplats} using OptiX~\cite{parker2010optix}.
The retained hits along the ray give exact plane intersections within their support, where we evaluate generalized Gaussian opacity.
Voronoi traversal~\cite{govindarajan2025radiant} starts in the ray origin's cell and crosses bisector planes between Delaunay neighbors.

\textbf{Compositing recurrence.} We merge both event types with a finite tracing bound into distances $0=\tau_0<\tau_1<\dots<\tau_M=t_{\mathrm{far}}$.
Segment $m$ lies in one cell $j_m$ and has length $\ell_m=\tau_m-\tau_{m-1}$.
At $\tau_m$, a surfel contributes opacity $\alpha_m$ and color $\mathbf{c}_m$.
Cell boundaries and the tracing bound have $\alpha_m=0$.
Let $\mathbf{a}_m=e^{-\boldsymbol{\sigma}^A_{j_m}\ell_m}$ and $\mathbf{b}_m=e^{-\boldsymbol{\sigma}^B_{j_m}\ell_m}$ be the transmittances of the segment.
The tracer accumulates direct radiance $\mathbf{D}$ and backscatter $\mathbf{S}$, tracking their medium transmittances $\mathbf{T}^A$ and $\mathbf{T}^B$.
Both share surfel transmittance $T^G$.
From unit transmittances and zero radiance, the tracer updates for $m=1,\dots,M$
\begin{equation}
  \begin{aligned}
    \makebox[2em][l]{$\mathbf{S}$} &\leftarrow \mathbf{S} + T^G_{m-1}\,\mathbf{T}^B_{m-1}\odot\mathbf{B}_{j_m}\odot(\mathbf{1}-\mathbf{b}_m),\\
    \makebox[2em][l]{$\mathbf{T}^A_m$} &\leftarrow \mathbf{T}^A_{m-1}\odot\mathbf{a}_m,\\
    \makebox[2em][l]{$\mathbf{T}^B_m$} &\leftarrow \mathbf{T}^B_{m-1}\odot\mathbf{b}_m,\\
    \makebox[2em][l]{$\mathbf{D}$} &\leftarrow \mathbf{D} + T^G_{m-1}\,\alpha_m\,\mathbf{T}^A_m\odot\mathbf{c}_m,\\
    \makebox[2em][l]{$T^G_m$} &\leftarrow T^G_{m-1}\,(1-\alpha_m).
  \end{aligned}
  \label{eq:recurrence}
\end{equation}
The first line integrates the backscatter of the segment, and the next two advance the medium transmittances.
The fourth adds the surfel contribution attenuated by the water before it, and the last removes the light it blocks.
Backscatter accumulates under $\mathbf{T}^B$, which keeps the recurrence invariant to how homogeneous water is split into cells.
\Cref{fig:overview} draws one such ray, with the same symbols.
The background contributes no radiance.

\textbf{Gradients and outputs.} A custom backward pass analytically differentiates \cref{eq:recurrence} with respect to surfel parameters, cell coefficients, and sites.
Site gradients pass through bisector intersections with fixed Delaunay adjacency during differentiation.
For ray $\mathbf{r}$ with unit direction $\mathbf{d}$, it also returns surfel-only opacity $A(\mathbf{r})=1-T^G_M$, excluding the medium.

\subsection{Directional scattering and camera model}
\label{sec:appearance}

\textbf{Directional scattering.} Water brightness varies with viewing direction relative to the light.
A Henyey and Greenstein phase gain~\cite{henyey1941diffuse} scales only the backscatter, so the rendered radiance $\mathbf{L}$ of a ray becomes
\begin{equation}
  \mathbf{L} = \mathbf{D} + \boldsymbol{\gamma}(\mathbf{d})\odot\mathbf{S},
  \label{eq:light}
\end{equation}
with a learned light direction and per-channel anisotropy shared across cells.
Medium parameters remain position-dependent, not direction-dependent.
At evaluation, a training-photo fit extends the gain beyond observed directions without changing it inside their range.

\textbf{Camera model.} Photometric differences between photographs are often absorbed by appearance embeddings~\cite{martin2021nerf} or bilateral grids~\cite{wang2024bilateral}, which do not distinguish surface color from scattering.
PPISP~\cite{deutsch2026ppisp} models exposure, color, vignetting, and response after medium rendering.
Novel-view exposure and color are interpolated from nearby training cameras, without fitting to test photographs.

\subsection{Separating surface and medium}
\label{sec:priors}

Matching photographs does not determine their surface–medium decomposition.
Our priors discourage medium leakage and haze surfels with no supervision of the split, using surfel-only opacity $A$ and backscatter $\mathbf{S}$ before the phase gain.
Unlike losses on Gaussians rendered with a ray-conditioned or image-space medium, they act on two sets of primitives that the tracer can render alone.

\textbf{Keeping haze out of the surfels.} A binarization term pushes surfel opacities $o_i$ toward zero or one, so a surfel is either surface or absent.
Coverage and clearing act on rays selected by their current surfel opacity,
\begin{equation}
  \mathcal{L}_{\mathrm{cov}} = \bigl\langle 1-A(\mathbf{r})\bigr\rangle_{A>\eta_{\mathrm{cov}}},
  \label{eq:coverage}
\end{equation}
\begin{equation}
  \mathcal{L}_{\mathrm{clr}} = \bigl\langle A(\mathbf{r})\bigr\rangle_{A<\eta_{\mathrm{clr}}},
  \label{eq:clearing}
\end{equation}
where $\langle\cdot\rangle$ averages over the rays that meet the condition, selected without gradient, and $\eta_{\mathrm{clr}}<\eta_{\mathrm{cov}}$.
The coverage term encourages an opaque seabed that occludes the water behind it, while preserving scattering in front.
Surfels are removed from rays whose coverage falls below the lower threshold, so the medium alone explains them.
Silhouette rays between the thresholds receive neither term.

\textbf{Keeping the surface out of the medium.} On rays with sufficient surfel opacity, a chroma prior keeps backscatter near the water hue without constraining its brightness.
The reference hue is estimated from backscatter on rays with negligible surfel opacity, not from external labels.
An overlap term penalizes medium opacity near the blended surface depth, weighted by confidence in the recovered surface.
Only the medium receives its gradients.
The supplement defines the chroma weighting, overlap proxy, and depth window.

\textbf{Keeping the medium physical.} Smoothness between Delaunay neighbors stabilizes poorly observed cells.
Cells whose sites lie far from every surfel have no surface nearby, and we call them free water.
A further term pulls free-water cells near training cameras toward the median parameters of more distant free-water cells.
This discourages camera-adjacent medium from acting as a per-photo brightness correction that affects nearby test views.

\subsection{Training}
\label{sec:training}

We jointly optimize the scene with
\begin{equation}
  \mathcal{L} = \mathcal{L}_{\mathrm{img}} + \mathcal{L}_{\mathrm{mv}} + \mathcal{L}_{\mathrm{reg}} + \sum_{p} \lambda_p \mathcal{L}_p.
  \label{eq:objective}
\end{equation}
The photometric term combines absolute error and D-SSIM, normalized by the detached prediction to emphasize dark regions~\cite{mildenhall2022nerf}.
The remaining terms regularize multi-view geometry, surfel scale, camera parameters, and separation (\cref{sec:priors}).
Depth terms on virtual views~\cite{kweon2026oceansplat} are traced through both components, with depth blended from surfel hits alone.

Surfels start at structure-from-motion points~\cite{schonberger2016structure}, and medium sites surround the points and cameras.
Initial medium coefficients are fitted to point observations at different distances, without masks.
Densification and pruning follow 3DGS~\cite{kerbl20233d}, with attenuation-compensated gradients so water does not suppress refinement of distant surfaces.
The supplement gives the initialization, gradient compensation, loss definitions, and schedules.

For synthetic scenes with a confined medium, we first fit cells alone to discourage surfels from absorbing the clouds.
Surfels are then initialized from cells with strong variation across neighbors, followed by joint training.

\subsection{Inserted objects and secondary rays}
\label{sec:insertion}

After training, the same tracer renders inserted objects, such as meshes with their own materials.
The first object hit at distance $t_o$ terminates the current ray segment.
We composite the surfels and water along this segment using \cref{eq:recurrence}, giving
\begin{equation}
  \mathbf{L} = \mathbf{D} + \boldsymbol{\gamma}(\mathbf{d})\odot\mathbf{S} + T^G(t_o)\,\mathbf{T}^A(t_o)\odot\mathbf{L}_o,
  \label{eq:insertion}
\end{equation}
where $\mathbf{D}$ and $\mathbf{S}$ include only contributions before $t_o$.
The object's outgoing radiance $\mathbf{L}_o$ includes recursively traced reflection and refraction~\cite{whitted1980improved}.
Reflected and refracted rays start in the cell that contains their origin and cross the surfels and water like camera rays (\cref{fig:applications}).
Matte parts use scene-derived illumination, and inserted lamps add light to surfels and water under a single-scattering approximation.

\begin{figure}[t]
\centering
\includegraphics[width=\linewidth]{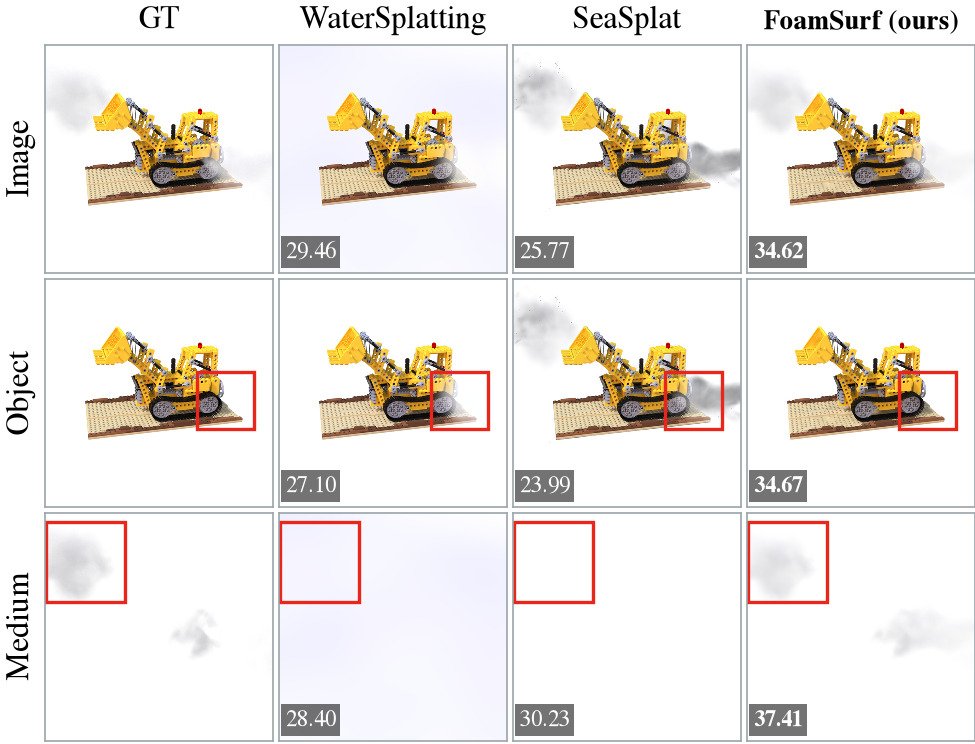}
  \caption{Lego with three surrounding clouds. Rows show the full scene, object, and medium, with PSNR against separate ground truth. All methods share initial surface points. Red boxes highlight cloud structure retained in baseline surfaces but assigned to the medium by \our{}. The white background differs from \cref{tab:synthetic}.}
  \label{fig:synthetic}
\end{figure}
\begin{figure}[t]
\centering
\captionof{table}{Foggy Garden restoration against clear reference images. Means and standard deviations over two seeds for \our{} and WaterSplatting, three otherwise. Hard-fog parameters are fitted to the published figure (\cref{sec:supp_setup}).}
  \label{tab:garden}
  \resizebox{\linewidth}{!}{
\begin{tabular}{l ccc ccc}
\toprule
 & \multicolumn{3}{c}{Easy fog} & \multicolumn{3}{c}{Hard fog} \\
\cmidrule(lr){2-4} \cmidrule(lr){5-7}
Method & PSNR$\uparrow$ & SSIM$\uparrow$ & LPIPS$\downarrow$ & PSNR$\uparrow$ & SSIM$\uparrow$ & LPIPS$\downarrow$ \\
\midrule
SeaThru-NeRF (JAX) & \cellcolor{resultOrange!80.00!resultYellow}15.20\std{0.01} & \cellcolor{resultOrange!40.00!resultYellow}.341\std{.003} & \cellcolor{resultOrange!0.00!resultYellow}.622\std{.006} & \cellcolor{resultOrange!80.00!resultYellow}13.97\std{0.55} & \cellcolor{resultOrange!40.00!resultYellow}.289\std{.025} & \cellcolor{resultOrange!0.00!resultYellow}.740\std{.033} \\
SeaThru-NeRF (Nerfstudio) & \cellcolor{resultOrange!0.00!resultYellow}12.89\std{2.47} & \cellcolor{resultOrange!0.00!resultYellow}.288\std{.040} & \cellcolor{resultOrange!40.00!resultYellow}.594\std{.023} & \cellcolor{resultOrange!0.00!resultYellow}10.02\std{0.14} & \cellcolor{resultOrange!0.00!resultYellow}.211\std{.007} & \cellcolor{resultOrange!40.00!resultYellow}.650\std{.013} \\
SeaSplat & \cellcolor{resultOrange!40.00!resultYellow}13.96\std{0.03} & \cellcolor{resultRed!20.00!resultOrange}.634\std{.002} & \cellcolor{resultRed!100.00!resultOrange}\textbf{.315}\std{.002} & \cellcolor{resultOrange!40.00!resultYellow}10.44\std{0.13} & \cellcolor{resultOrange!80.00!resultYellow}.380\std{.010} & \cellcolor{resultRed!60.00!resultOrange}.487\std{.009} \\
WaterSplatting & \cellcolor{resultRed!60.00!resultOrange}19.69\std{0.14} & \cellcolor{resultRed!100.00!resultOrange}\textbf{.696}\std{.001} & \cellcolor{resultRed!60.00!resultOrange}.318\std{.001} & \cellcolor{resultRed!60.00!resultOrange}15.46\std{0.00} & \cellcolor{resultRed!60.00!resultOrange}.495\std{.010} & \cellcolor{resultRed!20.00!resultOrange}.489\std{.001} \\
Plenodium & \cellcolor{resultRed!20.00!resultOrange}17.83\std{0.13} & \cellcolor{resultOrange!80.00!resultYellow}.601\std{.002} & \cellcolor{resultOrange!80.00!resultYellow}.351\std{.003} & \cellcolor{resultRed!20.00!resultOrange}15.29\std{0.05} & \cellcolor{resultRed!20.00!resultOrange}.423\std{.002} & \cellcolor{resultOrange!80.00!resultYellow}.522\std{.001} \\
\midrule
\our{} & \cellcolor{resultRed!100.00!resultOrange}\textbf{19.89}\std{0.29} & \cellcolor{resultRed!60.00!resultOrange}.657\std{.006} & \cellcolor{resultRed!20.00!resultOrange}.349\std{.003} & \cellcolor{resultRed!100.00!resultOrange}\textbf{16.81}\std{0.60} & \cellcolor{resultRed!100.00!resultOrange}\textbf{.584}\std{.015} & \cellcolor{resultRed!100.00!resultOrange}\textbf{.391}\std{.005} \\
\bottomrule
\end{tabular}
}
\end{figure}

\subsection{Editing and style transfer}
\label{sec:editing}
\our{} keeps surfaces and water in separate primitives that meet only in the tracer, so either can be changed after training (\cref{fig:applications}).

\textbf{Surface edits.} A deformation moves every surfel center, and its Jacobian maps the scaled tangent axes.
A thin singular value decomposition of that image gives the new axes and scales, and the cofactor fixes the normal.
Opacity, kernel exponent and color are kept, so the surfel is the first-order image of the original disk.

\textbf{Medium edits.} Cell coefficients and backscatter color are edited directly.
Scaling them clears or thickens the water, and replacing the hue gives another water type.
Matching the per-channel statistics of another scene transplants its water and keeps the reconstructed pattern.
A local edit inserts sites that inherit their nearest cell, and a field painted into them adds a plume or a haze layer.
The water of a ray can also be clipped below a water surface, at which rays are reflected and refracted (\cref{fig:draining}).

\textbf{Style transfer.} We stylize surfaces and water separately with the objective of CLIPGaussian~\cite{howil2025clipgaussian}.
The loss sees one representation rendered alone on black, against the same render before stylization.
Attenuation is never optimized, so the surfaces are seen through the stylized water as through the reconstructed one.
Each representation is optimized against its own render, so any stylized surface combines with any stylized water of the same scene (\cref{fig:applications}).

Derivations, parameters, schedules and the experimental protocol are given in the supplement.

\section{Experiments}
\label{sec:experiments}

Scene editing requires a usable surface and a usable medium, as well as accurate rendered views.
We evaluate these properties separately on real underwater scenes and on synthetic scenes with component ground truth.
Fog restoration tests the representation beyond water.

\textbf{Data and comparisons.} The four SeaThru-NeRF scenes~\cite{levy2023seathru} test reconstruction from real underwater photographs.
Lego and Ficus~\cite{mildenhall2021nerf} each contain three volumetric clouds arranged around the object, with separate ground truth for object and medium.
Foggy Garden~\cite{li2025watersplatting} provides clear reference photographs for restoration under easy and hard synthetic fog.
We compare methods with a medium~\cite{levy2023seathru,li2025watersplatting,yang2025seasplat,kweon2026oceansplat} and general scene representations without one~\cite{barron2023zip,kerbl20233d,govindarajan2025radiant}.
We also equip the strongest Gaussian baseline with PPISP~\cite{deutsch2026ppisp}, using exactly our camera model and novel-view interpolation rule.
Of the methods that give the medium a place in space, only Plenodium~\cite{wu2025plenodium} releases code, and we train it on our scenes.
The others~\cite{ramazzina2023scatternerf,chen2023dehazenerf,xing2025uw,jiang2026tri} release no code.

\textbf{Metrics and protocol.} Image quality is measured by PSNR, SSIM and VGG-based LPIPS~\cite{zhang2018unreasonable}.
The separation score treats the surface-only opacity $A$ as a soft classifier of the seabed and weights both classes equally,
\begin{equation}
  S_{\mathrm{sep}} = \tfrac{1}{2}\bigl(\langle A\rangle_{\mathrm{seabed}} + 1 - \langle A\rangle_{\mathrm{water}}\bigr),
  \label{eq:separation}
\end{equation}
so a perfect split scores one and a uniform prediction scores one half.
The masks come from Segment Anything~\cite{kirillov2023segment} and are used for evaluation only.
We report means and sample standard deviations at the final iteration.

\textbf{Novel view synthesis in water.} \our{} achieves mean PSNR comparable to the strongest baseline while representing surfaces and water with separate primitives (\cref{tab:seathru}).
The crops and error maps in \cref{fig:nvs} show the recovered seabed detail, with the lowest reconstruction error in the illustrated regions.

\our{} scores below that baseline on LPIPS (\cref{tab:seathru}).
Removing directional scattering, the separation priors or camera interpolation leaves it nearly unchanged.
Neither the medium nor the separation explains the gap.

Joint surface and medium tracing retains real-time rendering, with a frame rate below that baseline (\cref{tab:seathru}).
\our{} also stores a smaller model than that baseline.

\begin{figure}[t]
\centering
\includegraphics[width=\linewidth]{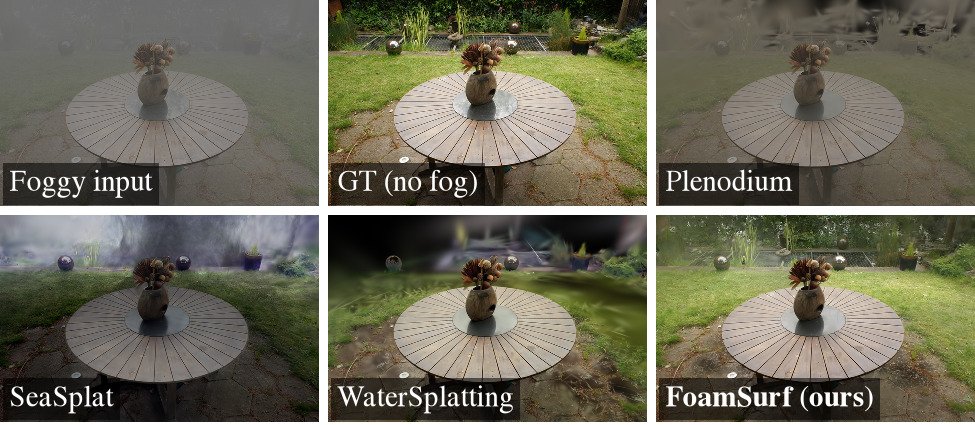}
  \caption{Hard-fog restoration with the medium removed. \our{} recovers background vegetation where baselines show residual haze, darkening or blur. See \cref{tab:garden} for quantitative results.}
  \label{fig:garden}
\end{figure}
\begin{figure}[t]
\centering
\captionof{table}{Ablation on SeaThru-NeRF, removing or replacing one component per variant. Values are means over four scenes and two seeds at the final iteration, with standard deviations in small numbers. $\Delta$PSNR is relative to the full method. Water is the surface-only opacity on water pixels, one component of the separation score of \cref{tab:seathru}.}
  \label{tab:ablation}
  \resizebox{\linewidth}{!}{
\begin{tabular}{l ccc cc}
\toprule
 & \multicolumn{3}{c}{Image quality} & \multicolumn{2}{c}{Disentanglement} \\
\cmidrule(lr){2-4} \cmidrule(lr){5-6}
Variant & PSNR$\uparrow$ & $\Delta$PSNR & SSIM$\uparrow$ & Water$\downarrow$ & Separation$\uparrow$ \\
\midrule
\our{} (full) & \cellcolor{resultRed!100.00!resultOrange}30.76\std{0.02} & \cellcolor{resultRed!100.00!resultOrange}0.00 & \cellcolor{resultRed!100.00!resultOrange}.929\std{.000} & \cellcolor{resultRed!20.00!resultOrange}.005\std{.001} & \cellcolor{resultRed!14.29!resultOrange}.901\std{.001} \\
\midrule
w/o medium (surfels only) & \cellcolor{resultOrange!0.00!resultYellow}29.03\std{0.45} & \cellcolor{resultOrange!0.00!resultYellow}$-$1.73 & \cellcolor{resultOrange!0.00!resultYellow}.915\std{.002} & \cellcolor{resultOrange!0.00!resultYellow}.977\std{.000} & \cellcolor{resultOrange!0.00!resultYellow}.506\std{.001} \\
homogeneous medium instead of cells & \cellcolor{resultRed!42.86!resultOrange}30.53\std{0.02} & \cellcolor{resultRed!33.33!resultOrange}$-$0.23 & \cellcolor{resultRed!100.00!resultOrange}.929\std{.000} & \cellcolor{resultOrange!40.00!resultYellow}.214\std{.011} & \cellcolor{resultOrange!28.57!resultYellow}.857\std{.005} \\
w/o separation priors & \cellcolor{resultRed!14.29!resultOrange}30.52\std{0.17} & \cellcolor{resultRed!33.33!resultOrange}$-$0.23 & \cellcolor{resultRed!50.00!resultOrange}.928\std{.000} & \cellcolor{resultOrange!80.00!resultYellow}.030\std{.002} & \cellcolor{resultOrange!57.14!resultYellow}.883\std{.001} \\
\midrule
w/o directional scattering & \cellcolor{resultRed!71.43!resultOrange}30.64\std{0.04} & \cellcolor{resultRed!66.67!resultOrange}$-$0.12 & \cellcolor{resultRed!100.00!resultOrange}.929\std{.000} & \cellcolor{resultRed!20.00!resultOrange}.005\std{.002} & \cellcolor{resultRed!42.86!resultOrange}.902\std{.000} \\
w/o camera model & \cellcolor{resultOrange!28.57!resultYellow}29.28\std{0.23} & \cellcolor{resultOrange!33.33!resultYellow}$-$1.48 & \cellcolor{resultOrange!100.00!resultYellow}.926\std{.001} & \cellcolor{resultRed!60.00!resultOrange}.004\std{.000} & \cellcolor{resultRed!71.43!resultOrange}.910\std{.001} \\
camera model w/o novel-view interpolation & \cellcolor{resultOrange!57.14!resultYellow}30.22\std{0.01} & \cellcolor{resultOrange!66.67!resultYellow}$-$0.53 & \cellcolor{resultRed!50.00!resultOrange}.928\std{.000} & \cellcolor{resultRed!20.00!resultOrange}.005\std{.000} & \cellcolor{resultOrange!85.71!resultYellow}.900\std{.001} \\
w/o multi-view depth terms & \cellcolor{resultOrange!85.71!resultYellow}30.30\std{0.22} & \cellcolor{resultOrange!100.00!resultYellow}$-$0.45 & \cellcolor{resultOrange!50.00!resultYellow}.924\std{.001} & \cellcolor{resultRed!100.00!resultOrange}.002\std{.001} & \cellcolor{resultRed!100.00!resultOrange}.921\std{.004} \\
\bottomrule
\end{tabular}
}
\end{figure}

\textbf{Disentanglement on real scenes.} \our{} has the highest mean separation score among the compared methods (\cref{tab:seathru}).
It also has the lowest surface opacity in open water across all scenes, with values near zero.
Plenodium matches our image quality, yet its primitives stay opaque in open water and its split scores one half.
The cells can explain scattering without surface primitives, while the separation priors discourage surfels in the water.
The surface-only renders retain the seabed structure while leaving open water nearly free of surfels (\cref{fig:decomposition}).
The surfels cover less of the seabed than those of the baselines, which assign more opacity to open water.
The uncovered pixels lie in dark crevices and on distant seabed.
There the seabed is as bright as the water, and the photographs carry almost no surface signal.
The split lets each component be rendered, edited and stylized on its own (\cref{fig:applications}).

\textbf{Synthetic scenes with ground truth.} \our{} achieves the highest object and medium reconstruction PSNR on both synthetic scenes (\cref{tab:synthetic}).
Its surfels recover the object, while its spatial medium reproduces the surrounding clouds (\cref{fig:synthetic}).
SeaSplat has higher full-image PSNR on Lego, but lower scores for both components.
Like WaterSplatting, it assigns cloud structure to Gaussian primitives instead of recovering it in the medium.
Both baselines use our surface initialization procedure, yet their medium-only renders do not show the clouds (\cref{fig:synthetic}).

\textbf{Restoration under fog.} Under hard fog, \our{} achieves the highest restoration PSNR among the compared methods (\cref{tab:garden}).
Rendering the surfels alone removes the contribution assigned to the medium and reveals the recovered scene.
The background vegetation remains visible, while the baseline renders are blurred, hazy or darkened (\cref{fig:garden}).
Under easy fog, restoration PSNR is comparable to the strongest baseline, which scores better on SSIM and LPIPS.

\textbf{Applications.}
A reconstruction that places the water in space serves as a scene, not only as a set of views (\cref{fig:applications}).
None of the compared methods can be used this way, because each evaluates its medium only along camera rays.

Insertion tests this directly, because the synthetic scenes can be re-rendered with the same object to give a reference (\cref{sec:insertion}).
\our{} reproduces the inserted object most accurately on both scenes (\cref{tab:insertion}).
The strongest baseline reconstructs one of them better without the object, and still degrades more once it is added.
Its medium follows the ray or the image, not the scene.
The water in front of a new object is therefore extrapolated and not read from the place it occupies.

The same tracer serves rays that no camera carried.
A ray that starts inside the scene crosses the surfels and cells like a camera ray, so reflection and refraction pass through the reconstructed water.
An inserted reflector shows the reconstructed seabed and the objects beside it, with the water of the path in between.
Inserted objects darken the seabed where they block the light from above.
The reconstructed surfaces occlude an inserted light, which leaves shadows in the water (\cref{fig:applications}).
Each component also changes on its own, because the priors kept them apart during reconstruction (\cref{sec:priors}).
Deforming the surfaces leaves the water as reconstructed, and editing the cells leaves the seabed untouched (\cref{fig:applications}).
Clipping the water of each ray below a surface lowers the level and exposes the seabed (\cref{fig:draining}).
Style transfer reaches either component alone (\cref{sec:editing,sec:suppl_editing}).


\textbf{Ablations.} \Cref{tab:ablation} tests the contribution of each component.
Without the medium, surfels must explain scattering, filling open water and causing the largest PSNR drop.
Homogeneous water raises the surfel opacity in open water by more than an order of magnitude (\cref{tab:ablation}, Water).
Removing the separation priors raises it several times.
The separation score averages two classes and compresses these differences.
Together, these results support spatial medium modeling and explicit separation constraints.

Removing the camera model causes the second-largest PSNR drop.
The baseline gains a similar amount from it (\cref{tab:seathru}), so this component is shared and not specific to \our{}.
Novel-view camera interpolation and directional scattering change PSNR with little effect on separation.
Multi-view depth terms improve PSNR but lower the separation score, a trade-off we retain for image quality.

\section{Conclusion}
\label{sec:conclusion}

Reconstructing a scene inside a scattering medium recovers the surfaces and the water as one.
Little can then be changed in the result.
\our{} gives each of them its own primitives, planar surfels for the surfaces and Voronoi cells for the water.
A single differentiable ray tracer renders both, and priors keep them apart during training.
The split is the most accurate among the compared methods, on real captures, on synthetic scenes and under dense fog.
The rendered views match the strongest baselines.
Each component can be edited or restyled alone, and inserted objects are seen through the recovered water.
Reflected and refracted rays cross it as camera rays do.
Beyond water, the same construction applies wherever a scene is observed through what fills it.
Captured scenes then move closer to assets that can be reused, rather than recordings that can only be replayed.
\textbf{Limitations.} The recovered water reproduces appearance and not optical properties. Its coefficients are the wideband quantities of a single-scattering model, and each cell holds one constant value. 


{
    \small
    \bibliographystyle{ieeenat_fullname}
    \bibliography{main,bib/related,bib/method}
}

\ifdefined\nosuppl\else
\clearpage
\appendix
\setcounter{page}{1}
\renewcommand{\thepage}{S\arabic{page}}
\setcounter{figure}{0}
\renewcommand{\thefigure}{S\arabic{figure}}
\renewcommand{\theHfigure}{supp.\arabic{figure}}
\setcounter{table}{0}
\renewcommand{\thetable}{S\arabic{table}}
\renewcommand{\theHtable}{supp.\arabic{table}}
\numberwithin{equation}{section}
\setcounter{topnumber}{3}
\setcounter{dbltopnumber}{3}
\raggedbottom
\maketitlesupplementary

This supplement gives the representation and rendering details, training procedures, and additional evaluations.
Reconstruction uses the differentiable hybrid tracer described in the main paper.
Inserted-object rendering is a separate, forward-only stage, whose lighting approximations are specified below.


\section{Representation details}
\label{sec:supp_representation}

\textbf{Kernel and parameterization.} The rotation quaternion is normalized to define the surfel's tangent axes and normal.
Positive tangent scales are obtained by exponentiating their learned log values, and a sigmoid maps the opacity parameter to $o_i$.
The shape exponent is one plus the softplus of its learned parameter.
At local coordinates $(u,v)$ along the scaled tangent axes, a generalized Gaussian kernel gives
\begin{equation}
  \alpha_i = o_i \exp\Bigl(-\frac{(u^2+v^2)^{\kappa_i}}{2\kappa_i}\Bigr).
  \label{eq:surfel}
\end{equation}
Here the scales and opacity are those used for rendering, including the filtering described below.
The learned shape exponent $\kappa_i>1$ flattens the top and sharpens the edge as it grows.
As $\kappa_i$ approaches one, the response becomes the Gaussian of 2DGS~\cite{huang20242d}.
For a nonparallel ray, the hit distance follows directly from intersection with the surfel plane.

\textbf{Color and filtering.} RGB spherical-harmonic coefficients encode the view-dependent color, evaluated once per surfel and camera during reconstruction.
A sigmoid scaled by $1.5$ bounds this color while allowing bright surfaces to exceed the photo range before camera processing.
The bound prevents unbounded colors in surfel-only renders.
The 3D smoothing filter of Mip-Splatting~\cite{yu2024mip} widens each surfel along its tangent axes and rescales its opacity.
Its width is derived from training-camera sampling and is not a learned representation parameter.
Spherical harmonics use degree three, with the active degree increased from zero by one every $667$ iterations.

\textbf{Medium parameterization.} Each Voronoi cell has a learned site, six raw optical coefficients, and three raw water-color parameters.
Softplus maps the optical coefficients to positive RGB attenuation and backscatter coefficients.
A sigmoid maps the water-color parameters to RGB values between zero and one.
Real underwater scenes use $5000$ cells throughout joint training, with no cell densification or pruning.
Sites move during optimization, and Delaunay adjacency is updated every $25$ iterations.
Gradients pass through ray intersections with the current cell faces, not through changes in adjacency.

\section{Hybrid ray tracing}
\label{sec:supp_tracing}

\textbf{Continuous image formation.} We write a ray as $\mathbf{x}(t)=\mathbf{o}+t\mathbf{d}$, with origin $\mathbf{o}$ and unit direction $\mathbf{d}$.
Within the tracing interval, retained surfel hits have distances $t_k$, opacities $\alpha_k$ and colors $\mathbf{c}_k$.
Their shared transmittance is $T^G(t)=\prod_{t_k<t}(1-\alpha_k)$.
The medium transmittance for the direct signal is
\begin{equation}
  \mathbf{T}^A(t)=\exp\Bigl(-\int_0^t\boldsymbol{\sigma}^A\bigl(\mathbf{x}(s)\bigr)\,ds\Bigr),
  \label{eq:transmittance}
\end{equation}
with $\mathbf{T}^B(t)$ defined analogously using $\boldsymbol{\sigma}^B$ for backscatter.
The direct signal is
\begin{equation}
  \mathbf{D}(\mathbf{r}) = \sum_k T^G(t_k)\,\alpha_k\,\mathbf{T}^A(t_k)\odot\mathbf{c}_k,
  \label{eq:direct}
\end{equation}
and the backscatter is
\begin{equation}
  \mathbf{S}(\mathbf{r}) = \int_0^{t_{\mathrm{far}}} T^G\,\mathbf{T}^B\odot\boldsymbol{\sigma}^B\odot\mathbf{B}\,dt,
  \label{eq:backscatter}
\end{equation}
Here the transmittances are evaluated at $t$, and the medium parameters at $\mathbf{x}(t)$.
Before water-light modulation and camera processing, radiance is $\mathbf{D}+\mathbf{S}$, with no background contribution.
Surfel occlusion acts on both signals through $T^G$.

\textbf{Derivation of the recurrence.} Consider segment $m$ of \cref{eq:recurrence}, with length $\ell_m$ inside cell $j_m$.
No surfel lies inside the segment, so its surfel transmittance is constant at $T^G_{m-1}$.
At distance $u$ from the segment start, backscatter transmittance is
\begin{equation}
  \mathbf{T}^B(\tau_{m-1}+u)
  =\mathbf{T}^B_{m-1}\odot e^{-\boldsymbol{\sigma}^B_{j_m}u}.
  \label{eq:supp_segment_transmittance}
\end{equation}
Integrating \cref{eq:backscatter} over this segment gives
\begin{equation}
  \Delta\mathbf{S}_m
  =T^G_{m-1}\,\mathbf{T}^B_{m-1}\odot\mathbf{B}_{j_m}\odot(\mathbf{1}-\mathbf{b}_m),
  \label{eq:supp_segment_backscatter}
\end{equation}
where $\mathbf{b}_m=e^{-\boldsymbol{\sigma}^B_{j_m}\ell_m}$.
This is the first update of \cref{eq:recurrence}.
The medium transmittances then advance to the segment endpoint.
A surfel there receives attenuation $\mathbf{T}^A_m$ and contributes the corresponding term of \cref{eq:direct}.
Only afterwards does its opacity reduce $T^G$, occluding subsequent surface and medium contributions.
At a cell boundary or the tracing bound, $\alpha_m=0$, leaving surfel transmittance unchanged.
Thus the recurrence evaluates the continuous model exactly for the piecewise-constant medium and retained hits, before early termination.

\textbf{Homogeneous-water limit.} For one opaque surfel at distance $z$ in homogeneous water, the direct signal is $\mathbf{c}\odot e^{-\boldsymbol{\sigma}^A z}$.
The surfel blocks all backscatter beyond $z$, leaving $\mathbf{B}\odot(\mathbf{1}-e^{-\boldsymbol{\sigma}^B z})$.
Their sum recovers \cref{eq:seathru}.

\textbf{Proxy polygons.} We intersect surfels through planar proxies using OptiX~\cite{parker2010optix}.
Each proxy is a regular octagon composed of six triangles in the surfel plane.
Its radius in normalized tangent coordinates is
\begin{equation}
  \varrho_i = \Bigl[\kappa_i\bigl(c_\alpha + 2\ln o_i\bigr)\Bigr]_+^{1/(2\kappa_i)},
  \label{eq:supp_proxy}
\end{equation}
with $c_\alpha=11.3449$.
At this radius the response of \cref{eq:surfel} falls to $\alpha=e^{-c_\alpha/2}\approx3.4\cdot10^{-3}$ for every opacity and shape exponent.
The vertices of the octagon lie on this level set, and its edges lie inside it.
A surfel with a lower opacity produces no hits.
Inside the octagon the response is evaluated exactly with \cref{eq:surfel}.
The opacity $o_i$ used here is the one used for rendering, after the smoothing filter and, early in training, after the depth-aware opacity of \cref{sec:supp_mv}.

\textbf{Hit buffer.} The proxy-intersection query searches forward from the ray origin.
The any-hit program keeps the $K$ nearest hits in a buffer sorted by insertion.
Once the buffer is full, a nearer hit replaces the farthest one, farther hits are rejected, and the maximum distance of the ray shrinks accordingly.
At most $K$ retained hits are composited within the finite tracing interval.
The proxy lies in the surfel plane, so the hit distance equals the exact ray and plane intersection, and the backward pass differentiates this intersection.
For the real underwater benchmark, we train with $K=64$ and render and evaluate with $K=32$.
The buffer comparison is discussed with the evaluation protocol in \cref{sec:supp_setup}.

\textbf{Termination.} One traversal carries six scalar compositing passes, one per color channel for the direct signal and one per channel for the backscatter.
For real underwater scenes, each pass stops accumulating once its own transmittance falls to $10^{-4}$, in training and evaluation alike.
This test runs after every surfel hit and every cell segment.
A ray ends when all six passes have stopped, at the far distance of $1000$ scene units, or after $1024$ cell steps.

\section{Rendering inserted objects}
\label{sec:supp_insertion}

Rendering with inserted objects (\cref{sec:insertion}) runs after training and computes no gradients.
Objects are analytic spheres or triangle meshes with a bounding volume hierarchy, intersected outside the OptiX scene of the surfels.
The hybrid tracer retains closed-form integration of the learned piecewise-constant medium along each traced segment.
Object illumination, shadows, and added lamp light use the approximations described below.
In particular, lamp transport uses spatially averaged medium parameters and numerical sampling, not the exact integration of \cref{eq:recurrence}.

\textbf{Segments.} The forward kernel accepts a far distance per ray.
For the ray up to its first object hit, it returns $\mathbf{D}$, $\mathbf{S}$ and the per-channel transmittance $T^G\,\mathbf{T}^A$ of \cref{eq:insertion}.
A secondary ray starts at its object hit, offset along its direction by a small fraction of the object size.
Traversal begins in the cell whose site is nearest to this origin.
The recursion stops after four bounces or when the weight a ray carries falls below $10^{-3}$ in every channel.
The spherical harmonics of every surfel are evaluated once per frame, for the direction towards the camera, and secondary rays reuse those colors.
A reflection therefore shows each surface with the color it has from the camera, not with the color it would have along the reflected direction.
The effect is limited to the view-dependent part of the surface color.

\textbf{Materials.} Reflective materials apply a tint to the reflected radiance.
Refractive materials use the Schlick approximation~\cite{schlick1994inexpensive} with an index of refraction relative to water.
The refracted ray leaves a sphere at its analytic second intersection and a mesh at the next intersection with the same mesh.
Total internal reflection sends the ray along the reflected direction.
A gas bubble is a sphere with a relative index below one, which casts no shadow.
Textured meshes use metal--roughness maps.
With albedo $\mathbf{a}$, metalness $m$, roughness $r$ and $\mathbf{F}_0=0.04(1-m)+m\,\mathbf{a}$, a single reflected ray is weighted by the Schlick term $\mathbf{F}$ times $(1-r)^2$.
The diffuse share $(1-m)(\mathbf{1}-\mathbf{F})\odot\mathbf{a}$ and the rough part of the specular share are lit as matte surfaces.
There is no glossy blur.

\textbf{Light on matte surfaces.} From the center of each object, $256$ rays are traced through the scene without objects.
Their radiance $\mathbf{L}_{\mathrm{env}}(\boldsymbol{\omega})$ gives the cosine-weighted irradiance of a normal $\mathbf{n}$.
Upward illumination is poorly constrained because the training photos do not observe the water surface.
We therefore add light from above calibrated on the seabed.
The mean radiance $\bar{\mathbf{L}}_{\downarrow}$ seen downward, divided by a reference albedo $\rho=0.7$, is the light on a horizontal surface.
A surface with normal $\mathbf{n}$ receives it scaled by $\bigl((1+\mathbf{n}\cdot\hat{\mathbf{u}})/2\bigr)^2$, where $\hat{\mathbf{u}}$ is the mean up direction of the training cameras.
Ambient occlusion by the other inserted objects scales this light.

\textbf{Shadows.} An object darkens the direct signal of the first surfel each ray hits, and the backscatter is left unchanged.
The factor combines the share of a cone of light around $\hat{\mathbf{u}}$ that the object blocks with its ambient occlusion.
The cone has a half-angle of $25^\circ$ and carries $60\%$ of the light.
The factor is analytic for spheres and sampled with $96$ rays for meshes.
The first surfel hits can vary irregularly between neighboring rays.
We smooth the shadow factor with a normalized Gaussian convolution restricted to pixels that see a surface.

\textbf{Directional scattering toward the surface.} Reflected rays can point toward the zenith, outside the directions observed in the training photos.
Extrapolating the water-light gain beyond those directions can produce excessive brightness.
When the direction's cosine with $\hat{\mathbf{u}}$ exceeds $0.6$, we evaluate the gain at this elevation threshold and the same azimuth.
Near the zenith, we use the azimuthal mean.
It is multiplied by the trained growth toward the zenith, softly limited to a factor of two.

\textbf{Lamps.} A lamp is an emissive sphere of intensity $I$ and color $\mathbf{e}$, optionally restricted to a cone with a smooth edge.
We normalize $I$ by the estimated scene illumination on an upward-facing surface, making its scale relative to the reconstructed lighting.
At distance $r$, the lamp delivers the irradiance
\begin{equation}
  \mathbf{E}(\mathbf{x}) = I\,\mathbf{e}\odot\frac{e^{-\bar{\boldsymbol{\sigma}}r}}{r^2}\,s(\mathbf{x})\,V(\mathbf{x}),
  \label{eq:supp_lamp}
\end{equation}
where $\bar{\boldsymbol{\sigma}}$ is the mean attenuation of the cells, $s$ the cone profile, and $V$ the visibility.
$V$ tests the first surfel along the shadow ray and the inserted objects, except emitters.
The surfel colors are treated as albedo under $\mathbf{E}_{\uparrow}$, the light of the scene on a surface facing up, computed as for matte surfaces.
At the first surfel of a ray, the lamp therefore adds $\mathbf{D}\odot\mathbf{E}_{\mathbf{n}}/\mathbf{E}_{\uparrow}$, where $\mathbf{E}_{\mathbf{n}}$ is the cosine-weighted irradiance at the surfel normal.
$\mathbf{D}$ already contains the water between the surfel and the camera.
On inserted objects, $\mathbf{E}_{\mathbf{n}}$ is added to the light on matte surfaces.

\textbf{Scattered lamp light.} Along a ray up to its first surface at $t_s$, a surfel or an object, the water scatters the lamp light toward the camera,
\begin{equation}
  \mathbf{S}_{\mathrm{lamp}} = \int_0^{t_s} \mathbf{b}\odot p_g(\mathbf{l}\cdot\mathbf{d})\,\mathbf{E}\bigl(\mathbf{x}(t)\bigr)\odot e^{-\bar{\boldsymbol{\sigma}}t}\,dt,
  \label{eq:supp_inscatter}
\end{equation}
where $\mathbf{l}$ points to the lamp and $p_g$ is the Henyey and Greenstein phase function with $g=0.9$.
We set the scattering weight to
\begin{equation}
  \mathbf{b} = \frac{\bar{\boldsymbol{\sigma}}\odot\bar{\mathbf{B}}}{\bar{\mathbf{L}}_{\mathrm{env}}},
  \label{eq:supp_scattering_weight}
\end{equation}
where $\bar{\mathbf{B}}$ is the mean water color of the cells and $\bar{\mathbf{L}}_{\mathrm{env}}$ the mean environment radiance.
With this $\mathbf{b}$, single scattering of the light of the scene gives the water color $\bar{\mathbf{B}}=\mathbf{b}\odot\bar{\mathbf{L}}_{\mathrm{env}}/\bar{\boldsymbol{\sigma}}$.
The integral uses $24$ stratified equiangular samples per lamp~\cite{kulla2012importance}.
On camera rays, each sample also tests the visibility of the lamp, which gives beams and shadows in the water.
The lamp light on the surfels and in the water is smoothed over the image by normalized Gaussian convolutions to remove the sampling noise.

\section{Directional scattering}
\label{sec:supp_waterlight}

\textbf{Learned phase gain.} Under single scattering of a distant light, the phase function scales backscatter relative to isotropic scattering.
We use the Henyey and Greenstein phase function~\cite{henyey1941diffuse},
\begin{equation}
  \boldsymbol{\gamma}(\mathbf{d}) = \frac{\mathbf{1}-\mathbf{g}\odot\mathbf{g}}{\bigl(\mathbf{1}+\mathbf{g}\odot\mathbf{g}-2\,\mathbf{g}\,(\mathbf{d}\cdot\hat{\mathbf{l}})\bigr)^{3/2}},
  \label{eq:phase}
\end{equation}
where $\hat{\mathbf{l}}$ is the normalized learned light direction and $\mathbf{g}$ is learned anisotropy per channel.
We parameterize the anisotropy as $0.95\tanh(\tilde{\mathbf{g}})$ and clamp the denominator base below by $10^{-6}$ for stability.
The parameters are shared by all cells and trained jointly with the scene.
Initialization uses the mean camera up direction and zero anisotropy, giving isotropic scattering.
The gain modulates only backscatter, never the direct signal.
Medium coefficients and water color depend on position, with directional appearance controlled by this gain.

\textbf{Evaluation-only extension.} The extension is not part of joint scene optimization.
Before each evaluation, we fit it using only the training photos and the current reconstruction.

\textbf{Samples.} We take every fourth row and column of every training photo.
We keep the rays on which the surfels cover less than $2\%$, $A(\mathbf{r})<0.02$, and whose darkest channel exceeds $0.01$.
For each kept ray we store its direction, its color, and its photo.
The elevation $\eta$ of a ray is the cosine between its direction and the mean up direction of the training cameras.

\textbf{Fit.} We fit a Henyey and Greenstein phase function of elevation to the observed colors by least squares in log space.
It has its own anisotropy per channel.
A free per-photo and per-channel log scale absorbs exposure and overall level.
The fit uses Adam with learning rate $0.02$ for $4000$ steps.
If fewer than $100$ rays are kept, the extension is disabled.

\textbf{Ratio.} Let $\eta_{\max}$ be the highest elevation among the kept rays.
We divide the fitted function at the ray elevation by its value at that elevation capped at $\eta_{\max}$.
This ratio multiplies the backscatter gain of \cref{eq:phase}.
It equals one at or below $\eta_{\max}$ and extends the fitted trend only above that elevation.
The vertical is the fixed mean camera up direction, not the learned light direction.

\section{Camera model and novel views}
\label{sec:supp_camera}

We apply the camera pipeline of PPISP~\cite{deutsch2026ppisp} after surface and medium rendering.
Exposure and color parameters vary per training photo, while vignetting and the response curve are shared.
This accounts for exposure and white balance differences without requiring corresponding changes to scene colors.
The pipeline applies exposure, vignetting, color correction, and the response curve in that order.

PPISP predicts novel-view exposure and color from rendered radiance using a controller.
We instead interpolate the learned exposure and color parameters from the four nearest training camera centers, reflecting their variation along the camera path.
Weights are proportional to inverse camera-center distance and normalized to sum to one.
Interpolation acts on the log-exposure and latent color parameters, not on rendered pixel colors.
Vignetting and response retain the shared learned parameters.
No parameters are predicted from the rendered image or fitted to test photographs.

\section{Multi-view depth terms}
\label{sec:supp_mv}

We adapt the trinocular stereo terms, the epipolar depth prior, the depth residual, and the depth-aware opacity of OceanSplat~\cite{kweon2026oceansplat}.
We implement these terms from the published equations.
All of them are evaluated with our ray tracer on the hybrid scene.

\textbf{Central render and depth.} The terms use a render of the training photo at a quarter of its resolution, with the photo resized to match.
It gives the direct signal $\mathbf{D}_c$, the backscatter $\mathbf{S}_c$, the surfel-only opacity $A$ and a depth.
Depth uses exact plane-intersection distances blended with weights $\alpha_k\prod_{k'<k}(1-\alpha_{k'})$ over the buffered surfel hits.
It excludes the medium and is converted to depth along the optical axis.
It is differentiable with respect to the hit opacities and distances.
Pixels with $A>0.5$ and positive depth form the surface mask.

\textbf{Virtual views.} Two virtual cameras share the training camera's orientation and intrinsics.
One is shifted horizontally by $b_h$ and the other vertically by $b_v$.
$b_v$ is drawn uniformly from $[-b,b]$ and $b_h=1.5\,b_v$, with $b=0.4$ in scene units.
We convert $b$ to our scene scale.
Both views are traced through the surfels and the spatial medium, with the surfel colors evaluated at the virtual camera center.
Rendered depth determines the disparity used to warp their direct signals into the central view.
Warping uses bilinear sampling and zero padding, restricted to the surface mask.

\textbf{Stereo terms.} The object term compares the warped direct signals with $\mathbf{D}_c$.
The full term passes the warped direct signal plus $\mathbf{S}_c$ through the camera model of the photo and compares it with the photo.
OceanSplat has no camera model, so this step is ours.
Both terms are absolute errors divided by $\mathbf{D}_c+\mathbf{S}_c$ under a stop-gradient plus $\epsilon=10^{-3}$.
Gradients reach the virtual renders, the central render, and the depth through the warp.

\textbf{Disparity smoothness.} An edge-aware term smooths each virtual view's disparity with weight $e^{-\gamma_s|\partial\mathbf{I}|}$, where $\gamma_s=1$.
Here $|\partial\mathbf{I}|$ is the channel mean of the photograph's image gradient.
We divide disparity by its mean over the surface mask~\cite{godard2019digging} to remove dependence on the baseline and focal length.

\textbf{Depth residual and epipolar prior.} The depth residual compares the rendered depth at the projection of each surfel center with the depth of the center.
It uses at most $65\,536$ random centers that project into the image and onto the surface mask.
Gradients reach both the rendered depth and the centers.
The epipolar prior uses the centers whose rendered opacity exceeds $0.8$ and whose projections into both virtual views fall inside the image.
OceanSplat triangulates a depth prior for each center from its two virtual projections.
For exact virtual cameras this triangulated point is the center itself, so we use its depth directly, as a target under a stop-gradient.
The prior penalizes $\log(1+|\cdot|)$ of the depth error, weighted by an edge-aware factor of the central render.

\textbf{Depth-aware opacity.} Early in training, the rendered opacity of every surfel is blended with the output of a small network.
The network has two hidden layers of $64$ units.
Its inputs are opacity, center depth normalized by scene extent, and viewing direction.
The blending weight decays linearly from $0.5$ to zero at iteration $8000$.
The blended opacity also sets the proxy radius of \cref{eq:supp_proxy}, so it changes which hits exist and not only their weights.

\textbf{Schedule and weights.} The terms are computed every second iteration from iteration $2667$ to iteration $16\,000$ of $20\,000$, with doubled weight.
The object, full, and smoothness terms have weight $0.1$, the depth residual $0.01$, and the epipolar prior decays linearly from $0.4$ to $0.2$.
The training resolution grows from a quarter to half resolution at iteration $4000$ and to full resolution at iteration $8000$.

\textbf{Implementation choices.} The normalization constants, disparity normalization, and opacity-network architecture above specify our implementation.
We also specify its opacity decay, center sampling, epipolar schedule, and detached depth target.

\section{Weights, thresholds, and schedules of the priors}
\label{sec:supp_weights}

\Cref{tab:supp_weights} lists the photometric, scale, and separation weights.
Multi-view and camera regularization are described separately in \cref{sec:supp_mv,sec:supp_camera,sec:supp_training}.
The training runs for $20\,000$ iterations.
A ramp $(s,n)$ raises the weight from zero at iteration $s$ to its full value at iteration $s+n$.
Ramps are linear except for the marked cosine ramp.
Ray-based opacity and chroma priors use up to $131\,072$ rays sampled without replacement from the training photo.
The overlap term uses $128$ rays sampled with replacement.

\begin{table}[!ht]
  \centering
  \small
  \setlength{\tabcolsep}{3pt}
  \begin{tabular}{@{}llll@{}}
    \toprule
    term & weight & ramp $(s,n)$ & threshold \\
    \midrule
    photometric & $1$ & none & \\
    surfel scale & $10^{-3}$ & none & \\
    binarization & $0.01$ & $(667,2000)$\textsuperscript{*} & \\
    smoothness & $0.1$, $0.1$, $1$ & $(0,167)$ & \\
    coverage & $0.003$ & $(667,667)$ & $\eta_{\mathrm{cov}}=0.9$ \\
    clearing & $0.01$ & $(2667,1333)$ & $\eta_{\mathrm{clr}}=0.3$ \\
    chroma & $1.0$ & $(667,667)$ & $\eta_{\mathrm{chr}}=0.02$ \\
    camera water & $0.01$ & $(667,667)$ & \\
    overlap & $0.03$ & $(667,667)$ & \\
    \bottomrule
  \end{tabular}
  \caption{Weights and schedules of the training terms. The three smoothness weights apply to the log attenuation, the log backscatter coefficient, and the water color, with Huber thresholds $0.1$, $0.1$ and $0.05$. \textsuperscript{*}Cosine ramp.}
  \label{tab:supp_weights}
\end{table}

\textbf{Backscatter chroma.} The chroma prior acts on rays with surfel-only opacity $A>\eta_{\mathrm{chr}}$, selected without gradient.
It penalizes the angle $\theta_{\mathbf{r}}$ between backscatter $\mathbf{S}(\mathbf{r})$ before the phase gain and the reference water hue $\hat{\mathbf{w}}$,
\begin{equation}
  \mathcal{L}_{\mathrm{chr}} = \bigl\langle \omega_{\mathbf{r}}\,[\sin^2\theta_{\mathbf{r}}-\sin^2\theta_0]_+ \bigr\rangle_{A>\eta_{\mathrm{chr}}}.
  \label{eq:chroma}
\end{equation}
The water hue is the normalized sum of detached backscatter on rays with $A<0.01$, tracked by a normalized exponential moving average with factor $0.99$.
The tolerated angle is $\theta_0=5^\circ$.
The weight $\omega_{\mathbf{r}}$ is the backscatter magnitude divided by its mean over the selected rays, both detached from the gradient.
This downweights dark, ill-defined hues without providing a gradient that rewards darkening.

\textbf{Medium smoothness and camera water.} Robust smoothness penalizes differences of log attenuation, log backscatter coefficients, and water colors between Delaunay neighbors.
The weights and Huber thresholds are given in \cref{tab:supp_weights}.
For the camera water term, let $\Delta$ be the median distance from a training camera to its nearest neighbor.
Free-water cells have their site farther than $\Delta$ from every surfel center.
Near cells are free-water cells within $2\Delta$ of a training camera, and reference cells are free-water cells farther than $3\Delta$ from all of them.
The sets are recomputed every $250$ iterations.
The near cells are penalized by squared differences from detached componentwise medians of the reference cells.
The penalty acts on log optical coefficients and raw water-color parameters and is zero if either cell set is empty.

\textbf{Surface–medium overlap.} Let $E$ be $1.1$ times the maximum training-camera distance from the mean camera center.
The overlap range is $L_{\mathrm{ov}}=\min(t_{\mathrm{far}},8E)$ and the window half width is $\Delta_{\mathrm{ov}}=0.02L_{\mathrm{ov}}$.
Its center is the blended surfel depth clipped to $[\Delta_{\mathrm{ov}},L_{\mathrm{ov}}-\Delta_{\mathrm{ov}}]$.
Ray confidence is the blended surfel opacity multiplied by $\exp(-\mathrm{Var}/\Delta_{\mathrm{ov}}^2)$, where $\mathrm{Var}$ is the variance of hit distances.
We form a scalar proxy density in each cell,
\begin{equation}
  k_j=\frac{1}{3}\operatorname{mean}_c\bigl(\sigma^A_{j,c}+\sigma^B_{j,c}+\sigma^B_{j,c}B_{j,c}\bigr).
  \label{eq:supp_overlap_density}
\end{equation}
Exact medium traversal over the window gives its bounded opacity proxy,
\begin{equation}
  O(\mathbf{r})=1-\exp\Bigl(-\int_{\mathrm{window}}k(\mathbf{x}(t))\,dt\Bigr).
  \label{eq:supp_overlap_opacity}
\end{equation}
The loss averages this proxy weighted by the detached ray confidence.
Reducing water color alone cannot remove the penalty because attenuation and backscatter coefficients contribute independently.
All surfel-derived quantities are detached, so gradients reach only medium coefficients, water colors, and sites.

\section{Training details}
\label{sec:supp_training}

Unless stated otherwise, the settings below apply to the real underwater scenes.
Synthetic and fog configurations are described separately in \cref{sec:supp_confined,sec:supp_setup}.

\textbf{Optimization.} We use Adam with moment parameters $(0.9,0.999)$ and $20\,000$ iterations.
The numerical offset is $10^{-8}$ for surfels and medium, and $10^{-15}$ for the camera model.
\Cref{tab:supp_learning_rates} lists the learning rates, with iteration index $n$ and decay $h(n)=2^{-n/8000}$.
For medium parameters, the initial rate is held until iteration $10\,000$, then interpolated logarithmically to the final rate.

\begin{table}[!ht]
  \centering
  \caption{Learning rates for real underwater reconstruction. Arrows denote the medium schedule described in the text.}
  \label{tab:supp_learning_rates}
  \small
  \begin{tabular}{@{}ll@{}}
    \toprule
    Parameter & Learning rate \\
    \midrule
    Surfel center & $2.5\cdot10^{-3}e^{-1.5\cdot10^{-4}n}$ \\
    Color coefficients & $\max(0.005h(n),0.001)$ \\
    Opacity & $\max(0.05h(n),0.025)$ \\
    Rotation & $\max(0.01h(n),0.001)$ \\
    Log scales & $0.0396538$ \\
    Shape exponent & $0.1$ \\
    Medium sites & $5\cdot10^{-5}\to5\cdot10^{-6}$ \\
    Optical coefficients & $10^{-2}\to10^{-3}$ \\
    Water color & $0.0025\to0.0005$ \\
    Phase parameters & $0.01$ \\
    Opacity network & $0.001$ \\
    \bottomrule
  \end{tabular}
\end{table}
The camera model starts from a base rate of $0.002$.
Its multiplier rises linearly from $0.01$ to one over $500$ iterations and follows an exponential decay of $0.01^{n/20000}$.
The depth-aware opacity network is optimized only while its contribution is active.

\textbf{Photometric objective.} The photometric term combines absolute error and D-SSIM between the camera-processed prediction and the training photo.
The absolute residual is divided per channel by the detached prediction plus $10^{-3}$.
Both inputs of D-SSIM are divided by this same quantity before computing their similarity.
This emphasizes errors in dark regions without differentiating through the normalization weights.
The D-SSIM weight is $0.2$, with the remaining weight assigned to absolute error.
The regularization term $\mathcal{L}_{\mathrm{reg}}$ contains a surfel scale regularizer and the camera regularizer of PPISP~\cite{deutsch2026ppisp}.
The multi-view term $\mathcal{L}_{\mathrm{mv}}$ is detailed in \cref{sec:supp_mv}, and the separation weights and schedules in \cref{tab:supp_weights}.

\textbf{Initialization from point tracks.} Surfels start at structure-from-motion points~\cite{schonberger2016structure} with their colors, while medium sites are scattered around the points and cameras.
All cells initially share the mean training-photo color as water color and a common pair of RGB optical coefficients.
We fit the coefficients to point tracks without segmentation masks.
A point observed at several camera distances must satisfy \cref{eq:seathru} with one surface color.
For each candidate coefficient pair, this color is solved in closed form by least squares, independently per point and color channel.
A grid search selects the coefficients minimizing the mean squared residual over the tracks.
We retain up to $3000$ points observed in at least three training photos, with a largest-to-smallest distance ratio of at least $1.3$.
Each observed color averages a $3\times3$ patch around the keypoint.
For each optical coefficient, the search uses $20$ logarithmically spaced values between $0.003$ and $0.5$.
Water color stays fixed during this fit to reduce its ambiguity with backscatter coefficients.
During joint training, the cells are free to vary and no term anchors them to this initialization.

\textbf{Attenuation-aware densification.} Surfels are densified and pruned using the screen-space gradient criterion of 3DGS~\cite{kerbl20233d}.
Water attenuates distant surfaces and their gradients, which would otherwise reduce their densification.
For each training view, let $\nu_i$ be the gradient magnitude for surfel $i$ and $\mathbf{o}$ the camera center.
We first estimate its mean RGB transmittance,
\begin{equation}
  \bar T_i=\operatorname{mean}_c\exp\bigl(-\bar\sigma^A_c\,\lVert\boldsymbol\mu_i-\mathbf{o}\rVert\bigr),
  \label{eq:supp_densify_transmittance}
\end{equation}
where $\bar{\boldsymbol\sigma}^A$ is the componentwise median attenuation over cells.
The compensated magnitude is
\begin{equation}
  \tilde\nu_i=\nu_i\min\bigl(1/\bar T_i,20\bigr).
  \label{eq:densify}
\end{equation}
Compensation is applied before averaging over views in which the surfel is visible.
The averaged gradients guide densification, with accumulated statistics reset after each adaptation event.
Adaptation is performed every $167$ iterations between iterations $333$ and $10\,000$.
Candidates have mean compensated gradient at least $2\cdot10^{-4}$.
Surfels with maximum tangent scale at most $0.02E$ are cloned, while larger candidates are split into two children.
The surfel count is capped at $700\,000$.
Opacity pruning starts after iteration $667$, using threshold $0.5$ before iteration $10\,000$ and $0.1$ afterwards.
At least $1000$ surfels are retained.
Opacity resets cap opacity at $0.5$ every $500$ iterations before iteration $10\,000$.
The smoothing filter is recomputed every $100$ iterations and after densification or pruning.

\section{Initialization of a confined medium}
\label{sec:supp_confined}

On each synthetic scene, the medium consists of three clouds arranged around the object.
We first fit the photos with a cell-only radiance field and adapt its cells.
For each cell we measure how much its values vary across its Delaunay neighbors.
A valley in the histogram of this variation separates the surface cells from the rest.
We initialize $100{,}000$ surfels from the identified surface cells.
Each cell's scalar density initializes both attenuation and backscatter coefficients in every color channel.
Joint training then continues from the input photos alone.
Joint optimization runs for $20{,}000$ iterations with the separation priors, medium smoothness, opacity binarization, and surfel densification.
The synthetic cameras are ideal, so we omit the camera model.
We also omit directional scattering and its extension, multi-view depth terms, the point-track coefficient fit, and the camera-water prior.
The weight of the coverage prior is $0.03$, the value with which these runs were trained, against $0.003$ on the real scenes.
The ground-truth renders of the object and of the clouds are used for evaluation only.

\section{Editing and style transfer}
\label{sec:supp_editing}

All edits act on a trained model and need no further optimization, except for style transfer.
Every edited model is stored as an ordinary run, so the renderer loads it like any reconstruction.

\textbf{Reference frame.} Deformations and spatial water edits are defined in a frame fitted to the visible seabed of one camera.
We render the surfels for every fourth pixel and keep the surface points of pixels with positive depth and opacity above $0.7$.
The center $c$ is the mean of the points that remain after discarding the $10\%$ farthest from their median.
The normal $n$ is the eigenvector of smallest eigenvalue of their covariance, oriented toward the camera.
The horizontal axes are the viewing direction projected on that plane and its cross product with $n$.
The horizontal scale $L$ is the $0.9$ quantile of the horizontal distance to $c$, and the vertical scale $H$ is the difference of the $0.995$ and $0.005$ quantiles of height.

\textbf{Deformations.} For a deformation $\phi$ given in these coordinates, each surfel center moves to $\phi(\boldsymbol{\mu}_i)$.
The Jacobian $J_i$ is evaluated by central differences with a step of $10^{-3}L$.
Writing $A_i=[\,s_{i,u}\mathbf{t}_{i,u},\,s_{i,v}\mathbf{t}_{i,v}\,]$, the thin decomposition $J_iA_i=U_i\Sigma_iV_i^\top$ gives the new tangent axes and scales.
The new normal is $\mathbf{u}_{i,1}\times\mathbf{u}_{i,2}$, oriented like $\mathrm{cof}(J_i)\mathbf{n}_i$, and the quaternion follows from the resulting rotation.
Opacity, kernel exponent and spherical-harmonic coefficients are unchanged, so the view-dependent bands are not rotated.
The three-dimensional filter is recomputed from the training cameras for the new positions.
The deformations of \cref{fig:suppl_edit_surface} raise a Gaussian hill of amplitude $0.22L$ under the seabed, scale relief by $2.5$ about the fitted plane, and bend the plane onto a cylinder of radius $0.6L$.

\textbf{Water edits.} Coefficients are edited after the softplus and colors after the sigmoid, then mapped back.
Scaling multiplies all six coefficients of every cell by $0.25$ or by $2.5$.
The algal variant keeps the luminance of each cell, replaces its hue, and multiplies attenuation per channel by $(1.0,0.95,1.5)$, so blue light is lost fastest.
The transplant matches the per-channel mean and standard deviation of the log coefficients and of the color logits to those of another scene, which leaves the spatial pattern of the reconstructed water in place.
The haze and plume edits first refine the medium, by splitting each cell into eight or by drawing new sites inside an ellipsoid, each new site taking the parameters of its nearest neighbor.
A smooth noise field then modulates the added coefficients, and the color is blended toward a sediment color.

\textbf{Lowering the water level.} The renderer receives a wet interval for every ray and intersects it with each integration segment.
Surfel hits and the Voronoi diagram are unchanged, and attenuation and backscatter vanish outside the interval.
Two cases are checked against the unclipped renderer: the full interval reproduces the original render, and the empty interval reproduces a medium with zero coefficients.
The interface carries one reflection and one refraction with the unpolarized Fresnel term and $\eta=1.333$.
Both rays are traced through the surfels and the clipped water, and the sky is procedural.
The animated waves, the foam and the spray of \cref{fig:draining} are visual effects and are not part of the reconstruction.

\textbf{Style transfer.} The loss is the objective of CLIPGaussian~\cite{howil2025clipgaussian}.
It combines a directional CLIP term, a term over $64$ random perspective crops of $128$ pixels, a VGG content term on \texttt{conv4\_2} and \texttt{conv5\_2}, a background term that holds the silhouette of the surfaces, and a pull toward the reconstructed parameters.
Their weights are $5$, $90$, $0.8$, $1000$ and $10^{-4}$, and the background term is used for the surfaces only.
Both runs use Adam for $5{,}000$ iterations and draw one training or test view of the scene per iteration.
For the surfaces we optimize the direct color coefficients at learning rate $0.04$ with the higher bands frozen, and the geometry at much smaller rates.
The geometry is held near the reconstruction by a penalty on the scales, by clipping the geometry gradient, and by clamping the log scales to their reconstructed range.
Two further measures suppress per-surfel hue noise, a cap on the color and an average of the color gradient over the eight nearest surfels.
For the water each cell is first split into $27$, which leaves the render unchanged, and we then optimize the backscatter colors at learning rate $0.045$, the backscatter coefficients and the sites.
The Delaunay topology is rebuilt every $25$ steps, and the attenuation is never optimized.
A run takes between $18$ and $32$ minutes on an RTX 3090, depending on the photo resolution.

\textbf{Why bright styles need frozen backscatter.} The loss sees the water alone, without the surfaces behind it.
With trainable backscatter coefficients and a bright style, the water in front of the seabed becomes opaque and hides it.
Freezing those coefficients keeps the opacity of the reconstruction, and the style then changes colors and the layout of the cells.
We use this setting for the fire prompt of \cref{fig:suppl_style}.

\section{Experimental details}
\label{sec:supp_setup}

\textbf{Data.} The four SeaThru-NeRF scenes~\cite{levy2023seathru} are real underwater captures, and every eighth photo is held out for testing.
The two synthetic scenes use the Lego and Ficus objects from the NeRF Blender set~\cite{mildenhall2021nerf}.
Each object has three volumetric clouds arranged around it.
Each scene has ground truth for the object alone and for the clouds alone, on a black background.
Each synthetic scene provides $200$ training views and $50$ test views at $800\times800$ pixels.
Foggy Garden adds the synthetic fog of WaterSplatting~\cite{li2025watersplatting} to Garden from Mip-NeRF 360~\cite{barron2022mip}.
We evaluate easy and hard fog levels against the clear photographs.
The fog model is $I=O\,e^{-\beta^{D}z}+B^{\infty}(1-e^{-\beta^{B}z})$, applied to each color channel with coefficients shared across channels.
Here $O$ is the clear image in $[0,1]$ and $z$ is the distance along the pixel ray in Nerfstudio scene units, which relate to COLMAP units by the scale $0.228479$.
Easy fog uses the published $\beta^{D}=0.6$, $\beta^{B}=0.6$ and $B^{\infty}=0.5$.
The published hard values do not reproduce the corresponding figure of that paper, so hard fog uses $\beta^{D}=1.1$, $\beta^{B}=0.7$ and $B^{\infty}=0.47$, fitted to that figure.

\textbf{Baselines.} SeaThru-NeRF runs in its original and in its Nerfstudio~\cite{tancik2023nerfstudio} implementation.
WaterSplatting runs as released and with our per-photo camera model and its rule for novel views, so this component is the same as in \our{}.
For OceanSplat~\cite{kweon2026oceansplat}, we report the authors' published values, not rerun measurements, because it releases no code.
On the synthetic scenes, WaterSplatting and SeaSplat start both from their own initialization and from ours.
Their own initialization draws random points in a box around the object, as both methods do for Blender scenes.
Our initialization supplies surface points and colors produced by the procedure in \cref{sec:supp_confined} for each scene and seed.

\textbf{Metrics.} LPIPS uses VGG features.
The separation score treats the opacity of the surface alone, with the medium ignored, as a soft classifier of the seabed.
The score averages mean surface opacity on seabed pixels and mean transparency on water pixels, giving equal weight to both classes.
A perfect split scores one, while uniformly opaque or uniformly transparent predictions score one half.
Segmentation masks are used only for evaluation.
On the synthetic scenes, the surface alone is compared with the object and the medium alone with the clouds.
Restoration compares the surface-only render against the clear photograph.
For \our{}, this render uses a black background and is evaluated before the camera model.

\textbf{Protocol.} Training on the GPU is not bitwise deterministic, so every method we run is trained with several seeds.
We report the final iteration, never the best test iteration.
Rerun methods use an RTX 4090 for training and rendering, except RadFoam training, which requires an A100 for memory capacity.
RadFoam rendering is measured on the same RTX 4090 as the other methods.
Published values are marked separately and excluded from rerun rankings.

\textbf{Hits per ray.} The tracer collects the surfel hits of a ray in a buffer of fixed size.
On the real underwater benchmark, \our{} is trained with $64$ hits per ray and evaluated with $32$.
In the buffer-size comparison, reducing the buffer changed PSNR by at most $0.004$\,dB and left LPIPS unchanged.
The final configuration's rendering speed is reported in \cref{tab:seathru}.
All values of \our{} in the main table come from one benchmark with $32$ hits per ray.

\textbf{Foggy Garden.} We reconstruct the Foggy Garden benchmark from clear photographs using the image formation model of WaterSplatting~\cite{li2025watersplatting}.
Depth is measured along camera rays in Nerfstudio scene units~\cite{tancik2023nerfstudio}.
These conventions were selected to match the published fog illustration.
The easy level uses the stated parameters $\beta^D=0.6$, $\beta^B=0.6$ and $B^\infty=0.5$.
The stated hard parameters do not reproduce the hard image of the same figure.
We therefore fit them to that image and obtain $\beta^D=1.1$, $\beta^B=0.7$ and $B^\infty=0.47$.
Every eighth view is held out for testing, at a resolution of $1297\times840$ pixels.

\textbf{\our{} under fog.} We adjust the underwater configuration for fog.
The multi-view depth terms are omitted because their epipolar term does not test visibility, which is important for the occlusions in Garden.
The clearing prior of \cref{eq:clearing} is left out, because the fogged background is part of the scene and not open water.
A term weighted by $0.03$ anchors attenuation to its initial fit.
Directional scattering and its evaluation extension are disabled, and the surfel RGB sigmoid has unit scale.
The coverage and opacity-binarization weights are $0.003$ and $0.001$, respectively.
The opacity-pruning threshold rises from $0.005$ to $0.1$ along a cosine ramp over $2000$ iterations, starting at iteration $667$.
Pruning runs every $167$ iterations and removes at most $10\%$ of surfels per event. Opacity resets are disabled.
The hit buffer holds $64$ intersections in both training and evaluation.
Ray termination uses threshold $10^{-4}$ during training and $0.01$ during evaluation.

\clearpage
\onecolumn
\textbf{LPIPS of the ablation variants.} Measured on the saved models of \cref{tab:ablation} with the same VGG network as the main table, averaged over the
four scenes and two seeds: homogeneous medium $.197$, full method $.202$, camera model without novel-view
interpolation $.202$, without directional scattering $.203$, without the separation priors $.204$, without the multi-view
depth terms $.211$. The spread of the variants that keep a medium is smaller than the gap to the strongest
baseline, so the difference in LPIPS does not come from the medium or from the separation priors.

\section{Per-scene results on SeaThru-NeRF}
\label{sec:suppl_per_scene}
\Cref{tab:suppl_per_scene,tab:suppl_separation} break down \cref{tab:seathru} by scene.
Values are the mean and the standard deviation over three seeds at the final iteration.
The opacity components distinguish surface coverage from residual surface primitives in open water.
They complement the aggregate separation score without changing the evaluation masks or protocol.

\begin{center}
  \begin{minipage}{\textwidth}
  \centering
  \captionof{table}{Per-scene image quality on the SeaThru-NeRF dataset. Mean and standard deviation over three seeds at the final iteration. LPIPS uses VGG features.}
  \label{tab:suppl_per_scene}
  \scriptsize
  \setlength{\tabcolsep}{2.5pt}
  \resizebox{\textwidth}{!}{
\begin{tabular}{l ccc ccc ccc ccc}
\toprule
 & \multicolumn{3}{c}{IUI3 Red Sea} & \multicolumn{3}{c}{Cura\c{c}ao} & \multicolumn{3}{c}{J.\ Gardens Red Sea} & \multicolumn{3}{c}{Panama} \\
\cmidrule(lr){2-4} \cmidrule(lr){5-7} \cmidrule(lr){8-10} \cmidrule(lr){11-13}
Method & PSNR$\uparrow$ & SSIM$\uparrow$ & LPIPS$\downarrow$ & PSNR$\uparrow$ & SSIM$\uparrow$ & LPIPS$\downarrow$ & PSNR$\uparrow$ & SSIM$\uparrow$ & LPIPS$\downarrow$ & PSNR$\uparrow$ & SSIM$\uparrow$ & LPIPS$\downarrow$ \\
\midrule
SeaThru-NeRF (JAX) & \cellcolor{resultOrange!44.44!resultYellow}26.87\std{0.02} & \cellcolor{resultOrange!0.00!resultYellow}.845\std{.002} & \cellcolor{resultOrange!22.22!resultYellow}.334\std{.003} & \cellcolor{resultOrange!66.67!resultYellow}30.89\std{0.08} & \cellcolor{resultOrange!75.00!resultYellow}.898\std{.003} & \cellcolor{resultOrange!25.00!resultYellow}.324\std{.005} & \cellcolor{resultOrange!44.44!resultYellow}22.16\std{0.05} & \cellcolor{resultOrange!28.57!resultYellow}.789\std{.006} & \cellcolor{resultOrange!25.00!resultYellow}.351\std{.002} & \cellcolor{resultOrange!22.22!resultYellow}28.53\std{0.07} & \cellcolor{resultOrange!0.00!resultYellow}.850\std{.003} & \cellcolor{resultOrange!0.00!resultYellow}.353\std{.004} \\
SeaThru-NeRF (Nerfstudio) & \cellcolor{resultOrange!66.67!resultYellow}26.92\std{0.23} & \cellcolor{resultOrange!100.00!resultYellow}.888\std{.002} & \cellcolor{resultRed!55.56!resultOrange}.233\std{.003} & \cellcolor{resultRed!11.11!resultOrange}31.19\std{0.95} & \cellcolor{resultOrange!100.00!resultYellow}.940\std{.002} & \cellcolor{resultOrange!100.00!resultYellow}.192\std{.002} & \cellcolor{resultOrange!88.89!resultYellow}23.45\std{0.15} & \cellcolor{resultRed!14.29!resultOrange}.882\std{.001} & \cellcolor{resultRed!75.00!resultOrange}.178\std{.003} & \cellcolor{resultOrange!88.89!resultYellow}31.68\std{0.08} & \cellcolor{resultRed!25.00!resultOrange}.947\std{.001} & \cellcolor{resultRed!55.56!resultOrange}.148\std{.002} \\
Zip-NeRF & \cellcolor{resultRed!55.56!resultOrange}30.91\std{0.09} & \cellcolor{resultRed!25.00!resultOrange}.901\std{.001} & \cellcolor{resultRed!100.00!resultOrange}\textbf{.211}\std{.000} & \cellcolor{resultOrange!88.89!resultYellow}30.96\std{0.12} & \cellcolor{resultRed!25.00!resultOrange}.945\std{.001} & \cellcolor{resultRed!50.00!resultOrange}.173\std{.002} & \cellcolor{resultRed!11.11!resultOrange}24.06\std{0.20} & \cellcolor{resultRed!71.43!resultOrange}.895\std{.002} & \cellcolor{resultRed!100.00!resultOrange}\textbf{.172}\std{.008} & \cellcolor{resultRed!55.56!resultOrange}32.55\std{0.12} & \cellcolor{resultRed!100.00!resultOrange}\textbf{.953}\std{.000} & \cellcolor{resultRed!100.00!resultOrange}\textbf{.136}\std{.001} \\
3DGS & \cellcolor{resultOrange!22.22!resultYellow}25.44\std{0.16} & \cellcolor{resultOrange!50.00!resultYellow}.867\std{.002} & \cellcolor{resultRed!33.33!resultOrange}.241\std{.003} & \cellcolor{resultOrange!22.22!resultYellow}27.95\std{0.67} & \cellcolor{resultOrange!25.00!resultYellow}.867\std{.012} & \cellcolor{resultOrange!50.00!resultYellow}.226\std{.002} & \cellcolor{resultOrange!22.22!resultYellow}21.15\std{0.56} & \cellcolor{resultOrange!57.14!resultYellow}.856\std{.001} & \cellcolor{resultOrange!50.00!resultYellow}.206\std{.002} & \cellcolor{resultOrange!66.67!resultYellow}29.08\std{0.32} & \cellcolor{resultOrange!50.00!resultYellow}.897\std{.000} & \cellcolor{resultOrange!44.44!resultYellow}.195\std{.001} \\
RadFoam & \cellcolor{resultOrange!0.00!resultYellow}24.94\std{2.66} & \cellcolor{resultOrange!25.00!resultYellow}.850\std{.048} & \cellcolor{resultOrange!0.00!resultYellow}.339\std{.052} & \cellcolor{resultOrange!0.00!resultYellow}20.63\std{1.47} & \cellcolor{resultOrange!0.00!resultYellow}.728\std{.064} & \cellcolor{resultOrange!0.00!resultYellow}.447\std{.063} & \cellcolor{resultOrange!0.00!resultYellow}20.77\std{0.58} & \cellcolor{resultOrange!0.00!resultYellow}.781\std{.007} & \cellcolor{resultOrange!0.00!resultYellow}.367\std{.014} & \cellcolor{resultOrange!0.00!resultYellow}27.75\std{0.09} & \cellcolor{resultOrange!25.00!resultYellow}.882\std{.001} & \cellcolor{resultOrange!22.22!resultYellow}.285\std{.003} \\
SeaSplat & \cellcolor{resultOrange!88.89!resultYellow}27.60\std{0.19} & \cellcolor{resultOrange!75.00!resultYellow}.869\std{.001} & \cellcolor{resultRed!77.78!resultOrange}.212\std{.012} & \cellcolor{resultOrange!44.44!resultYellow}29.90\std{0.19} & \cellcolor{resultOrange!50.00!resultYellow}.894\std{.003} & \cellcolor{resultOrange!75.00!resultYellow}.207\std{.001} & \cellcolor{resultOrange!66.67!resultYellow}22.71\std{0.27} & \cellcolor{resultOrange!85.71!resultYellow}.871\std{.001} & \cellcolor{resultOrange!100.00!resultYellow}.185\std{.002} & \cellcolor{resultOrange!44.44!resultYellow}28.68\std{0.34} & \cellcolor{resultOrange!75.00!resultYellow}.898\std{.002} & \cellcolor{resultOrange!88.89!resultYellow}.162\std{.004} \\
WaterSplatting & \cellcolor{resultRed!11.11!resultOrange}29.98\std{0.07} & \cellcolor{resultRed!50.00!resultOrange}.913\std{.001} & \cellcolor{resultOrange!66.67!resultYellow}.250\std{.001} & \cellcolor{resultRed!33.33!resultOrange}31.75\std{0.36} & \cellcolor{resultRed!50.00!resultOrange}.955\std{.000} & \cellcolor{resultRed!50.00!resultOrange}.173\std{.001} & \cellcolor{resultRed!77.78!resultOrange}24.90\std{0.04} & \cellcolor{resultRed!100.00!resultOrange}\textbf{.899}\std{.001} & \cellcolor{resultRed!25.00!resultOrange}.182\std{.002} & \cellcolor{resultRed!11.11!resultOrange}31.87\std{0.12} & \cellcolor{resultOrange!100.00!resultYellow}.946\std{.001} & \cellcolor{resultRed!11.11!resultOrange}.156\std{.001} \\
WaterSplatting + PPISP & \cellcolor{resultRed!77.78!resultOrange}31.10\std{0.18} & \cellcolor{resultRed!75.00!resultOrange}.914\std{.002} & \cellcolor{resultRed!11.11!resultOrange}.245\std{.001} & \cellcolor{resultRed!100.00!resultOrange}\textbf{34.86}\std{0.03} & \cellcolor{resultRed!100.00!resultOrange}\textbf{.961}\std{.000} & \cellcolor{resultRed!100.00!resultOrange}\textbf{.163}\std{.000} & \cellcolor{resultRed!55.56!resultOrange}24.71\std{0.14} & \cellcolor{resultRed!100.00!resultOrange}\textbf{.899}\std{.002} & \cellcolor{resultRed!75.00!resultOrange}.178\std{.001} & \cellcolor{resultRed!33.33!resultOrange}32.16\std{0.07} & \cellcolor{resultRed!50.00!resultOrange}.948\std{.000} & \cellcolor{resultRed!33.33!resultOrange}.153\std{.001} \\
Plenodium & \cellcolor{resultRed!33.33!resultOrange}30.85\std{0.10} & \cellcolor{resultRed!50.00!resultOrange}.913\std{.000} & \cellcolor{resultOrange!88.89!resultYellow}.246\std{.000} & \cellcolor{resultRed!55.56!resultOrange}33.93\std{0.28} & \cellcolor{resultRed!75.00!resultOrange}.960\std{.000} & \cellcolor{resultRed!75.00!resultOrange}.167\std{.001} & \cellcolor{resultRed!100.00!resultOrange}\textbf{25.07}\std{0.03} & \cellcolor{resultRed!100.00!resultOrange}\textbf{.899}\std{.001} & \cellcolor{resultRed!50.00!resultOrange}.179\std{.001} & \cellcolor{resultRed!100.00!resultOrange}\textbf{32.89}\std{0.02} & \cellcolor{resultRed!75.00!resultOrange}.952\std{.000} & \cellcolor{resultRed!77.78!resultOrange}.147\std{.001} \\
\rowcolor{resultNA}
OceanSplat$^\dagger$ & 30.17 & .913 & n/a & 34.56 & .961 & n/a & 25.35 & .908 & n/a & 32.74 & .957 & n/a \\
\midrule
\our{} & \cellcolor{resultRed!100.00!resultOrange}\textbf{31.78}\std{0.12} & \cellcolor{resultRed!100.00!resultOrange}\textbf{.915}\std{.002} & \cellcolor{resultOrange!44.44!resultYellow}.254\std{.001} & \cellcolor{resultRed!77.78!resultOrange}34.33\std{0.08} & \cellcolor{resultRed!75.00!resultOrange}.960\std{.000} & \cellcolor{resultRed!25.00!resultOrange}.181\std{.001} & \cellcolor{resultRed!33.33!resultOrange}24.33\std{0.06} & \cellcolor{resultRed!42.86!resultOrange}.893\std{.001} & \cellcolor{resultOrange!75.00!resultYellow}.194\std{.002} & \cellcolor{resultRed!77.78!resultOrange}32.57\std{0.12} & \cellcolor{resultRed!50.00!resultOrange}.948\std{.001} & \cellcolor{resultOrange!66.67!resultYellow}.179\std{.001} \\
\bottomrule
\end{tabular}
}
  \end{minipage}
\end{center}

\begin{center}
  \begin{minipage}{\textwidth}
  \centering
  \captionof{table}{Per-scene separation on the SeaThru-NeRF dataset. Seabed and Water give mean surface-only opacity on the respective evaluation masks. Separation averages seabed opacity and water transparency, as in \cref{tab:seathru}. Values are means and standard deviations over three seeds at the final iteration.}
  \label{tab:suppl_separation}
  \scriptsize
  \setlength{\tabcolsep}{2.5pt}
  \resizebox{\textwidth}{!}{
\begin{tabular}{l ccc ccc ccc ccc}
\toprule
 & \multicolumn{3}{c}{IUI3 Red Sea} & \multicolumn{3}{c}{Cura\c{c}ao} & \multicolumn{3}{c}{J.\ Gardens Red Sea} & \multicolumn{3}{c}{Panama} \\
\cmidrule(lr){2-4} \cmidrule(lr){5-7} \cmidrule(lr){8-10} \cmidrule(lr){11-13}
Method & Seabed$\uparrow$ & Water$\downarrow$ & Separation$\uparrow$ & Seabed$\uparrow$ & Water$\downarrow$ & Separation$\uparrow$ & Seabed$\uparrow$ & Water$\downarrow$ & Separation$\uparrow$ & Seabed$\uparrow$ & Water$\downarrow$ & Separation$\uparrow$ \\
\midrule
SeaThru-NeRF (JAX) & \cellcolor{resultOrange!100.00!resultYellow}.936\std{.044} & \cellcolor{resultRed!33.33!resultOrange}.222\std{.160} & \cellcolor{resultRed!33.33!resultOrange}.857\std{.064} & \cellcolor{resultOrange!66.67!resultYellow}.778\std{.086} & \cellcolor{resultRed!66.67!resultOrange}.156\std{.026} & \cellcolor{resultRed!66.67!resultOrange}.811\std{.030} & \cellcolor{resultOrange!100.00!resultYellow}.922\std{.110} & \cellcolor{resultOrange!66.67!resultYellow}.724\std{.472} & \cellcolor{resultOrange!66.67!resultYellow}.599\std{.181} & \cellcolor{resultOrange!100.00!resultYellow}.901\std{.047} & \cellcolor{resultRed!33.33!resultOrange}.342\std{.101} & \cellcolor{resultRed!33.33!resultOrange}.779\std{.027} \\
SeaThru-NeRF (Nerfstudio) & \cellcolor{resultOrange!66.67!resultYellow}.914\std{.008} & \cellcolor{resultRed!66.67!resultOrange}.160\std{.003} & \cellcolor{resultRed!66.67!resultOrange}.877\std{.005} & \cellcolor{resultOrange!100.00!resultYellow}.958\std{.035} & \cellcolor{resultRed!33.33!resultOrange}.419\std{.289} & \cellcolor{resultRed!33.33!resultOrange}.769\std{.131} & \cellcolor{resultOrange!66.67!resultYellow}.881\std{.012} & \cellcolor{resultRed!66.67!resultOrange}.110\std{.007} & \cellcolor{resultRed!66.67!resultOrange}.885\std{.010} & \cellcolor{resultOrange!66.67!resultYellow}.890\std{.035} & \cellcolor{resultRed!66.67!resultOrange}.144\std{.043} & \cellcolor{resultRed!100.00!resultOrange}\textbf{.873}\std{.039} \\
SeaSplat & \cellcolor{resultOrange!0.00!resultYellow}.823\std{.009} & \cellcolor{resultOrange!100.00!resultYellow}.336\std{.075} & \cellcolor{resultOrange!100.00!resultYellow}.743\std{.033} & \cellcolor{resultOrange!0.00!resultYellow}.508\std{.002} & \cellcolor{resultOrange!100.00!resultYellow}.544\std{.026} & \cellcolor{resultOrange!0.00!resultYellow}.482\std{.014} & \cellcolor{resultOrange!0.00!resultYellow}.630\std{.031} & \cellcolor{resultRed!33.33!resultOrange}.368\std{.092} & \cellcolor{resultOrange!100.00!resultYellow}.631\std{.031} & \cellcolor{resultOrange!0.00!resultYellow}.487\std{.026} & \cellcolor{resultOrange!33.33!resultYellow}.842\std{.089} & \cellcolor{resultOrange!0.00!resultYellow}.322\std{.046} \\
WaterSplatting & \cellcolor{resultRed!66.67!resultOrange}.998\std{.002} & \cellcolor{resultOrange!33.33!resultYellow}.997\std{.003} & \cellcolor{resultOrange!0.00!resultYellow}.500\std{.002} & \cellcolor{resultRed!33.33!resultOrange}.994\std{.001} & \cellcolor{resultOrange!66.67!resultYellow}.690\std{.041} & \cellcolor{resultOrange!100.00!resultYellow}.652\std{.021} & \cellcolor{resultRed!33.33!resultOrange}.990\std{.003} & \cellcolor{resultOrange!33.33!resultYellow}.809\std{.114} & \cellcolor{resultOrange!33.33!resultYellow}.591\std{.057} & \cellcolor{resultRed!66.67!resultOrange}.992\std{.004} & \cellcolor{resultOrange!100.00!resultYellow}.489\std{.024} & \cellcolor{resultOrange!100.00!resultYellow}.751\std{.010} \\
WaterSplatting + PPISP & \cellcolor{resultRed!33.33!resultOrange}.993\std{.004} & \cellcolor{resultOrange!66.67!resultYellow}.865\std{.144} & \cellcolor{resultOrange!66.67!resultYellow}.564\std{.070} & \cellcolor{resultRed!66.67!resultOrange}.996\std{.001} & \cellcolor{resultOrange!33.33!resultYellow}.928\std{.013} & \cellcolor{resultOrange!66.67!resultYellow}.534\std{.006} & \cellcolor{resultRed!66.67!resultOrange}.991\std{.003} & \cellcolor{resultOrange!100.00!resultYellow}.710\std{.021} & \cellcolor{resultRed!33.33!resultOrange}.640\std{.009} & \cellcolor{resultRed!33.33!resultOrange}.990\std{.004} & \cellcolor{resultOrange!66.67!resultYellow}.535\std{.015} & \cellcolor{resultOrange!66.67!resultYellow}.727\std{.006} \\
Plenodium & \cellcolor{resultRed!100.00!resultOrange}\textbf{1.000}\std{.000} & \cellcolor{resultOrange!0.00!resultYellow}.998\std{.001} & \cellcolor{resultOrange!33.33!resultYellow}.501\std{.001} & \cellcolor{resultRed!100.00!resultOrange}\textbf{.998}\std{.001} & \cellcolor{resultOrange!0.00!resultYellow}.977\std{.012} & \cellcolor{resultOrange!33.33!resultYellow}.510\std{.006} & \cellcolor{resultRed!100.00!resultOrange}\textbf{.998}\std{.000} & \cellcolor{resultOrange!0.00!resultYellow}.997\std{.001} & \cellcolor{resultOrange!0.00!resultYellow}.500\std{.000} & \cellcolor{resultRed!100.00!resultOrange}\textbf{.996}\std{.001} & \cellcolor{resultOrange!0.00!resultYellow}.975\std{.020} & \cellcolor{resultOrange!33.33!resultYellow}.510\std{.010} \\
\midrule
\our{} & \cellcolor{resultOrange!33.33!resultYellow}.902\std{.006} & \cellcolor{resultRed!100.00!resultOrange}\textbf{.001}\std{.000} & \cellcolor{resultRed!100.00!resultOrange}\textbf{.950}\std{.003} & \cellcolor{resultOrange!33.33!resultYellow}.741\std{.005} & \cellcolor{resultRed!100.00!resultOrange}\textbf{.008}\std{.000} & \cellcolor{resultRed!100.00!resultOrange}\textbf{.866}\std{.002} & \cellcolor{resultOrange!33.33!resultYellow}.861\std{.003} & \cellcolor{resultRed!100.00!resultOrange}\textbf{.007}\std{.004} & \cellcolor{resultRed!100.00!resultOrange}\textbf{.927}\std{.001} & \cellcolor{resultOrange!33.33!resultYellow}.724\std{.004} & \cellcolor{resultRed!100.00!resultOrange}\textbf{.001}\std{.001} & \cellcolor{resultRed!66.67!resultOrange}.861\std{.002} \\
\bottomrule
\end{tabular}
}
  \end{minipage}
\end{center}

\clearpage
\section{Qualitative comparison with every baseline}
\label{sec:suppl_qualitative}
\Cref{fig:suppl_reconstruction} extends \cref{fig:nvs} to every method of \cref{tab:seathru} and a third view.

\begin{center}
  \begin{minipage}{\textwidth}
  \centering
  \includegraphics[width=\textwidth]{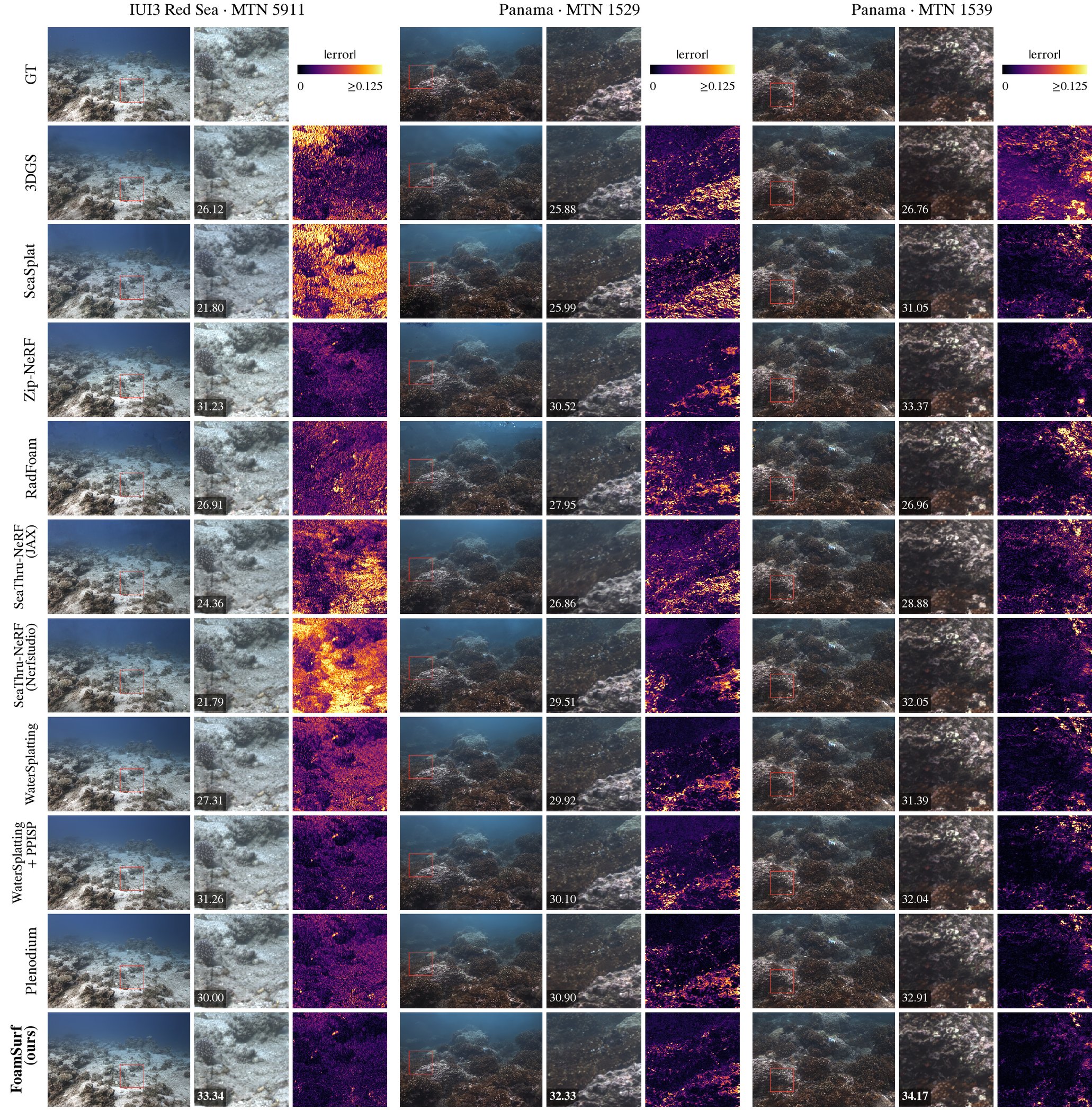}
  \captionof{figure}{Novel-view reconstruction on three SeaThru-NeRF views. Each block shows a full view, the red-box crop with its PSNR, and absolute error against the reference photograph. Error maps share one color scale. All methods use seed 42.}
  \label{fig:suppl_reconstruction}
  \end{minipage}
\end{center}

\clearpage
\section{Additional surface–medium separation}
\label{sec:suppl_decomposition}
\Cref{fig:suppl_decomposition} complements \cref{fig:decomposition} with a view of Japanese Gardens.

\begin{center}
  \begin{minipage}{\textwidth}
  \centering
  \includegraphics[width=\textwidth]{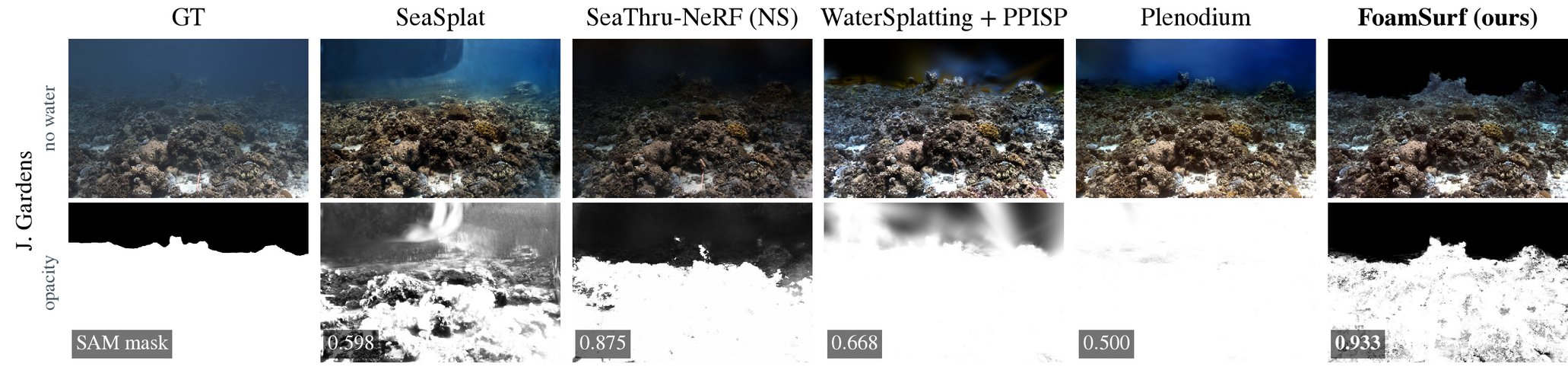}
  \captionof{figure}{Surface–medium separation on Japanese Gardens. Water-free renders appear above surface-only opacity maps, with per-view separation scores. White denotes seabed in the evaluation mask and opaque surfaces in predictions. Methods and visualization conventions match \cref{fig:decomposition}.}
  \label{fig:suppl_decomposition}
  \end{minipage}
\end{center}

\textbf{Away from the test cameras.}
The separation figures so far use held-out test views.
Those cameras sit among the training cameras, where every method is fitted.
A continuous path leaves that neighbourhood and asks what each representation holds away from the photographs.
\Cref{fig:suppl_orbit_iui3,fig:suppl_orbit_curacao} render one path per scene with every method, with the medium and without it.
No reference photographs exist along the path, so the comparison is qualitative.

The rendered rows agree.
The rows without the medium do not.
The baselines keep part of the water in the surface, as haze over the seabed or as a bright band where the open water was.
\our{} leaves the open water empty and the seabed with its own colour.
The clearing prior of \cref{sec:supp_weights} removes surfels from rays whose opacity marks them as open water.
No surface primitive then remains to carry the haze once the medium is gone.
This is the behaviour that \cref{tab:suppl_prior_ablation} attributes to that prior.

\begin{center}
  \begin{minipage}{\textwidth}
  \centering
  \includegraphics[width=0.86\textwidth]{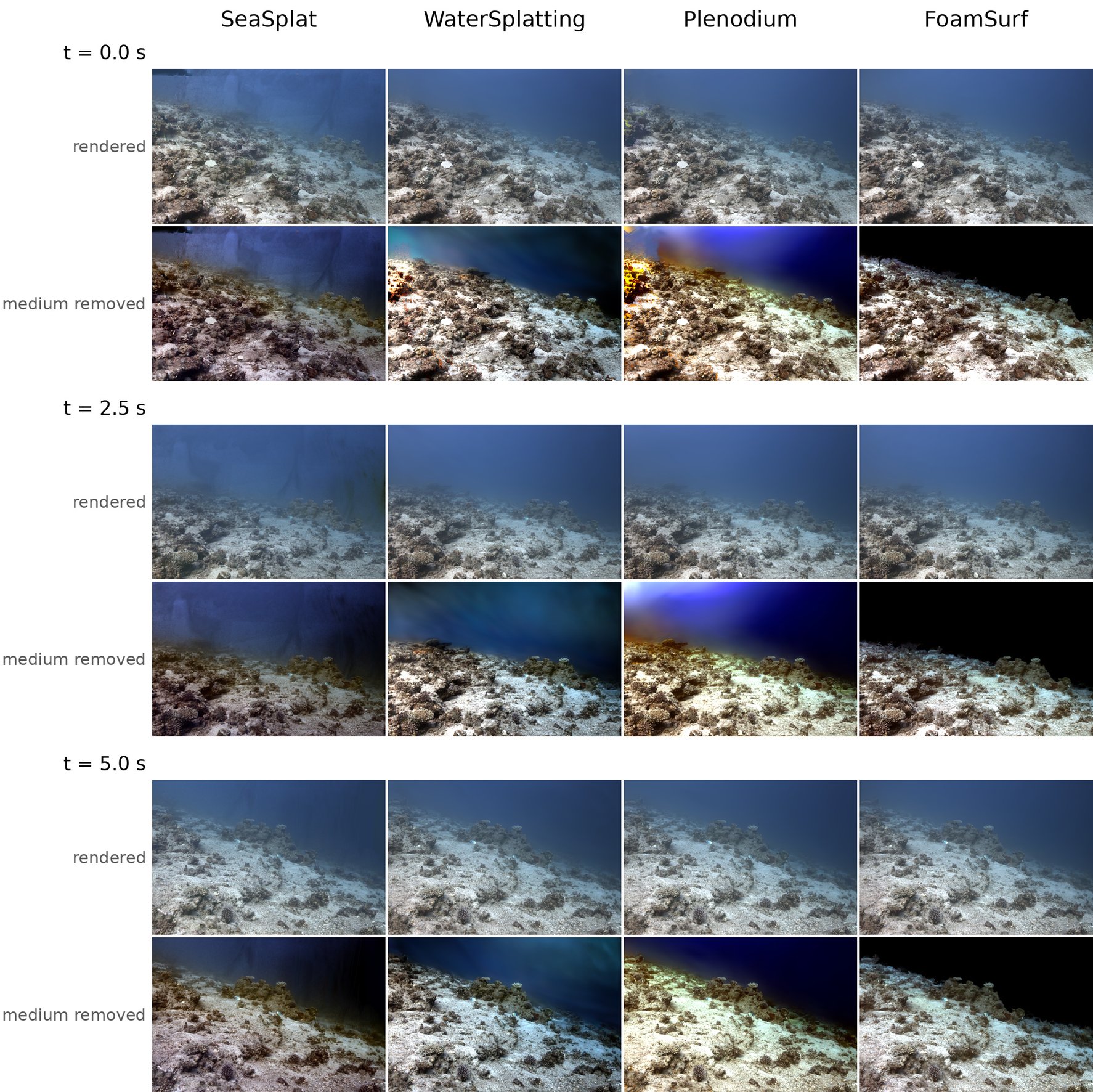}
  \captionof{figure}{A camera path on IUI3 Red Sea. Each block is one moment of the path. The upper row is the rendered frame and the lower row is the same frame with the medium removed. Columns follow the order of \cref{tab:seathru}.}
  \label{fig:suppl_orbit_iui3}
  \end{minipage}
\end{center}

\clearpage
\begin{center}
  \begin{minipage}{\textwidth}
  \centering
  \includegraphics[width=0.86\textwidth]{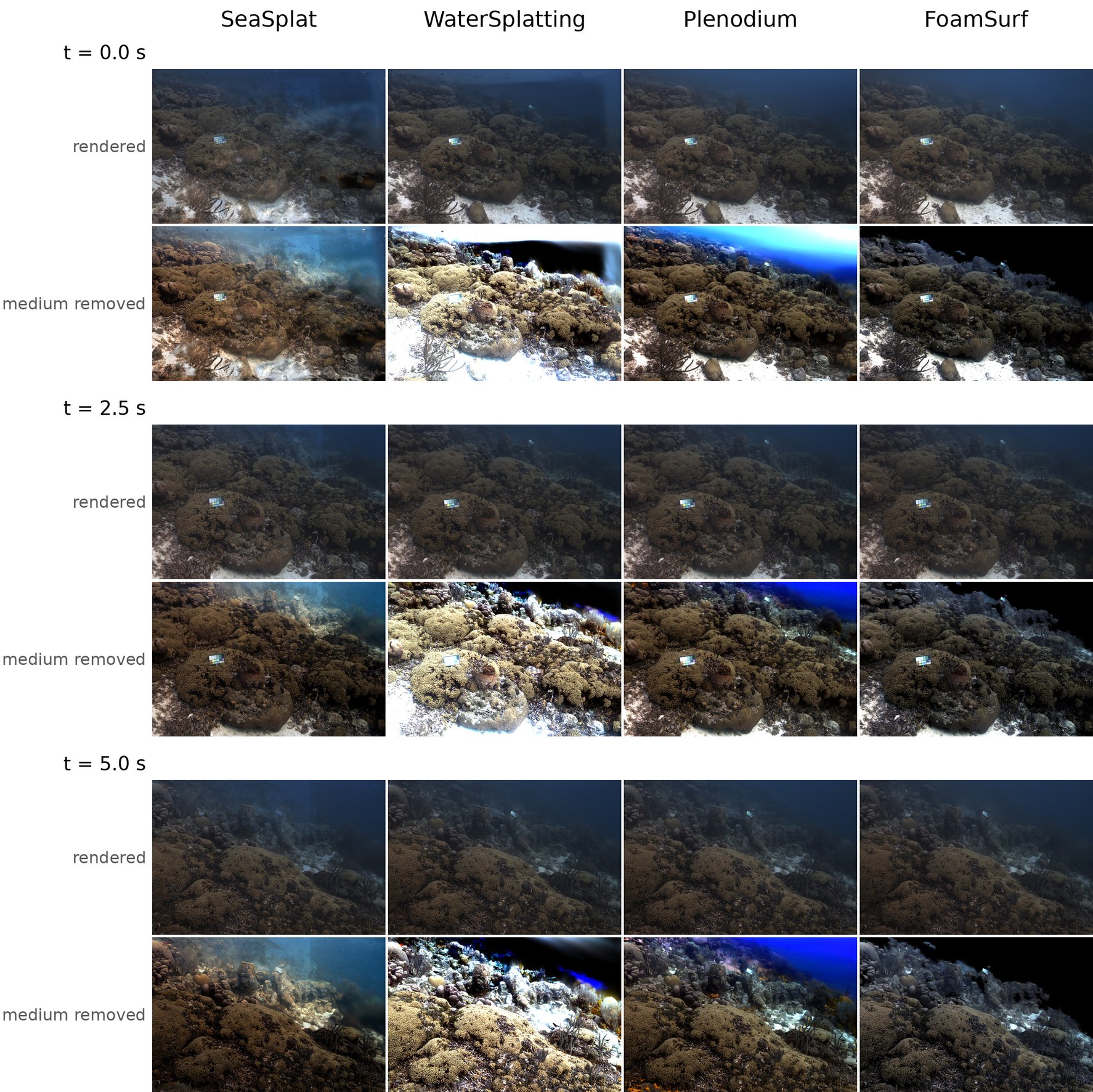}
  \captionof{figure}{A camera path on Cura\c{c}ao. Rows, columns and conventions follow \cref{fig:suppl_orbit_iui3}. The open water above the reef stays empty only under \our{}.}
  \label{fig:suppl_orbit_curacao}
  \end{minipage}
\end{center}

\clearpage

\section{Object insertion}
\label{sec:suppl_insertion}

\textbf{Protocol.} A sphere of constant color is added at a fixed world position to the two synthetic scenes, chosen without looking at any method's output and checked to stand clear of the object.
It emits a fixed radiance and is visible to camera rays only, so it neither lights the scene nor casts a shadow, and the comparison tests occlusion, attenuation, and veiling rather than a lighting model.
Blender renders the truth with the same sphere, the same $50$ test cameras, and the settings of the dataset.
Every method renders the same cameras from its trained model, without retraining and without fitting anything to the images that contain the sphere.
The region of interest is the full silhouette of the sphere, including the pixels where reconstructed geometry stands in front of it.
Outside that silhouette, dilated by three pixels, every prediction is unchanged from the render without the sphere, and the reported base quality agrees with the values shipped with the models to within $10^{-5}$\,dB.

\textbf{The baselines.} Neither baseline can place an object in a scene, so we added one to each native rasterizer for inference only.
Both cut the sorted sequence of primitives at the depth of the sphere, give the sphere the color that their own medium leaves after attenuating it to that depth, and accumulate their backscatter only up to it.
The additions are in the forward kernels and the backward kernels are untouched.
They are our adaptations and not features of those methods.

\textbf{What the two scenes agree and disagree on.} On both scenes the strongest baseline loses more from the insertion than \our{} does: the drop from the scene to the object is $2.21$ against $0.83$\,dB on Lego and $4.80$ against $2.64$\,dB on Ficus.
Where that loss falls differs.
On Ficus \our{} leads on $47$ of $50$ views.
On Lego the median view is a tie and the advantage sits in the tail, $2.5$\,dB over the ten views where the baseline is weakest.
WaterSplatting fails on both scenes, and on Lego it fails most where the sphere stands in open space, which its medium renders as an almost empty image.
The hardest views are not the same ones on the two scenes, and the symptom depends on how each baseline fitted its medium.
What the two scenes share is the property behind the cost: neither baseline represents the medium as a function of position along the ray, so the water in front of an object at a new depth is extrapolated from a model fitted to the surfaces that were there.
We therefore report the cost of the insertion and that property, and not a single failure mode.

\begin{center}
  \begin{minipage}{\textwidth}
  \centering
  \includegraphics[width=0.84\textwidth]{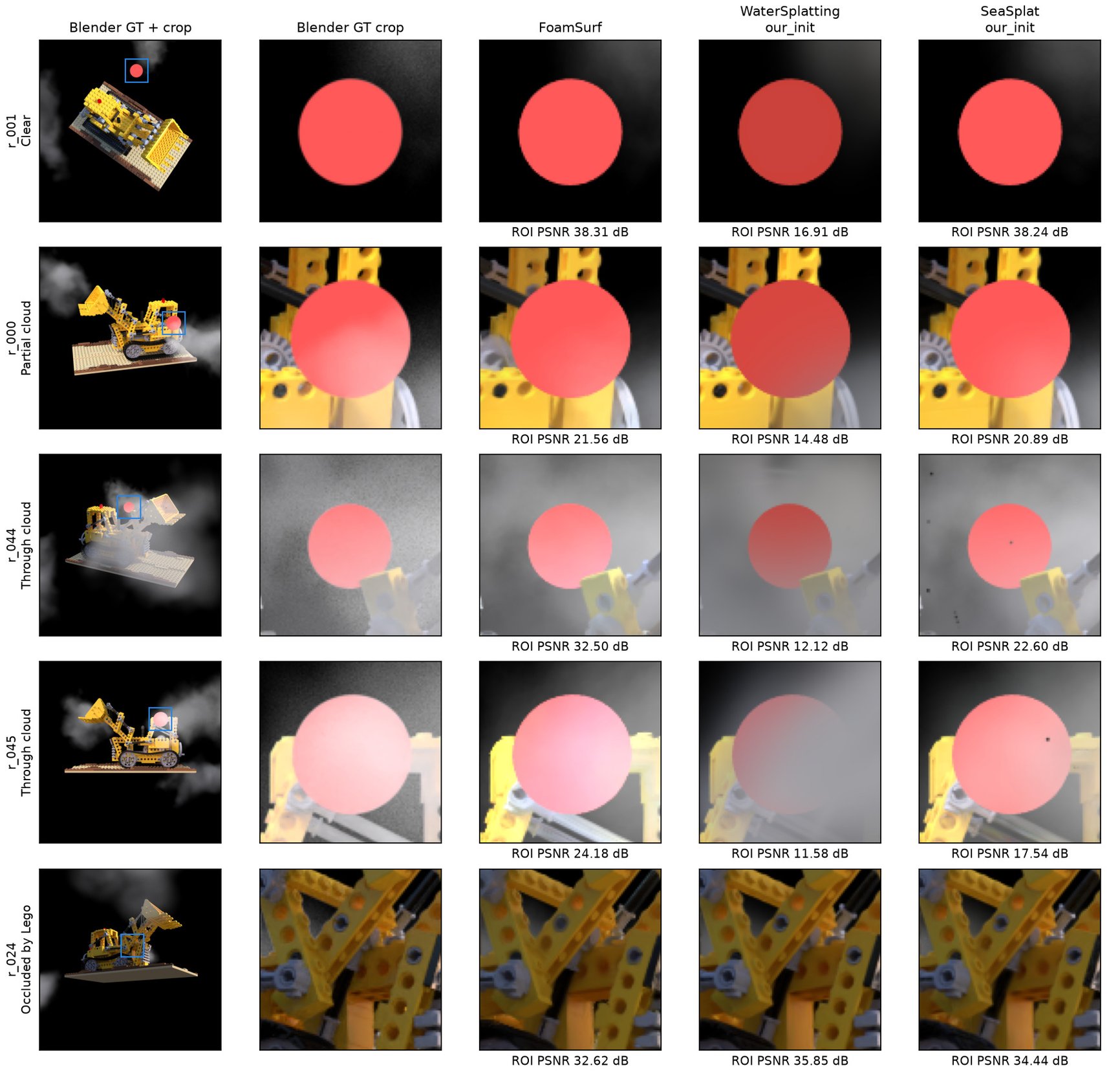}
  \captionof{figure}{Object insertion on Lego. Each row is one test camera, named on the left with what stands between the camera and the sphere. The first column is the Blender rendering with the crop marked, the second its crop, and the rest the same crop from each reconstruction with the PSNR inside the sphere's silhouette. Both baselines start from our initial points.}
  \label{fig:suppl_insertion_lego}
  \end{minipage}
\end{center}

\begin{center}
  \begin{minipage}{\textwidth}
  \centering
  \includegraphics[width=0.84\textwidth]{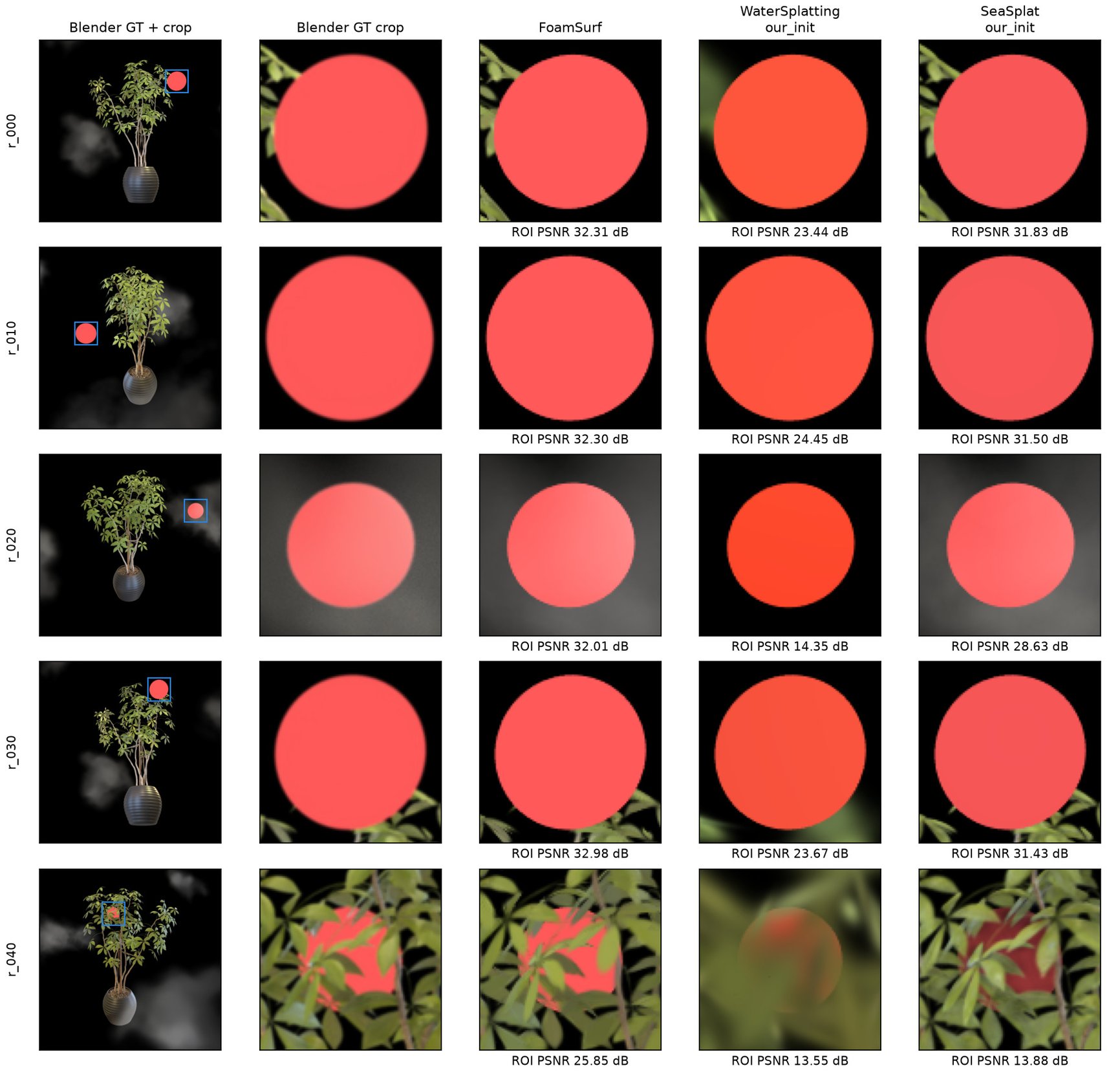}
  \captionof{figure}{Object insertion on Ficus, with the conventions of \cref{fig:suppl_insertion_lego}.}
  \label{fig:suppl_insertion_ficus}
  \end{minipage}
\end{center}

\begin{center}
  \begin{minipage}{\textwidth}
  \centering
  \captionof{table}{Object insertion, all variants and metrics. Scene: the reconstruction without the sphere. Object: the mean over cameras inside its silhouette. Object (pooled): one value from the squared error of every silhouette pixel of every camera, which weights the hard views more. MAE: the mean absolute error in the same region. Own init.: the baseline started from its own initial points instead of ours.}
  \label{tab:suppl_insertion}
  \small
  \resizebox{\textwidth}{!}{
\begin{tabular}{l cccc cccc}
\toprule
 & \multicolumn{4}{c}{Lego} & \multicolumn{4}{c}{Ficus} \\
\cmidrule(lr){2-5} \cmidrule(lr){6-9}
 & Scene & Object & Object (pooled)$\uparrow$ & MAE$\downarrow$ & Scene & Object & Object (pooled)$\uparrow$ & MAE$\downarrow$ \\
Method & PSNR$\uparrow$ & PSNR$\uparrow$ &  &  & PSNR$\uparrow$ & PSNR$\uparrow$ &  &  \\
\midrule
WaterSplatting, own init. & \cellcolor{resultOrange!0.00!resultYellow}11.80 & \cellcolor{resultOrange!0.00!resultYellow}4.09 & \cellcolor{resultOrange!0.00!resultYellow}3.72 & \cellcolor{resultOrange!0.00!resultYellow}.5663 & \cellcolor{resultOrange!0.00!resultYellow}15.62 & \cellcolor{resultOrange!0.00!resultYellow}4.00 & \cellcolor{resultOrange!0.00!resultYellow}3.92 & \cellcolor{resultOrange!0.00!resultYellow}.5601 \\
WaterSplatting & \cellcolor{resultOrange!50.00!resultYellow}32.85 & \cellcolor{resultOrange!50.00!resultYellow}17.25 & \cellcolor{resultOrange!66.67!resultYellow}15.89 & \cellcolor{resultOrange!50.00!resultYellow}.1406 & \cellcolor{resultOrange!50.00!resultYellow}19.27 & \cellcolor{resultOrange!50.00!resultYellow}20.82 & \cellcolor{resultOrange!50.00!resultYellow}18.84 & \cellcolor{resultOrange!50.00!resultYellow}.0764 \\
SeaSplat, own init. & \cellcolor{resultRed!50.00!resultOrange}35.16 & \cellcolor{resultRed!50.00!resultOrange}33.11 & \cellcolor{resultRed!33.33!resultOrange}26.16 & \cellcolor{resultOrange!100.00!resultYellow}.0180 & \cellcolor{resultOrange!100.00!resultYellow}30.60 & \cellcolor{resultOrange!100.00!resultYellow}26.23 & \cellcolor{resultOrange!100.00!resultYellow}19.94 & \cellcolor{resultOrange!100.00!resultYellow}.0519 \\
SeaSplat & \cellcolor{resultRed!100.00!resultOrange}\textbf{35.30} & \cellcolor{resultOrange!100.00!resultYellow}33.09 & \cellcolor{resultRed!33.33!resultOrange}26.16 & \cellcolor{resultRed!50.00!resultOrange}.0179 & \cellcolor{resultRed!50.00!resultOrange}31.58 & \cellcolor{resultRed!50.00!resultOrange}26.78 & \cellcolor{resultRed!50.00!resultOrange}23.28 & \cellcolor{resultRed!50.00!resultOrange}.0446 \\
\midrule
\our{} & \cellcolor{resultOrange!100.00!resultYellow}34.22 & \cellcolor{resultRed!100.00!resultOrange}\textbf{33.39} & \cellcolor{resultRed!100.00!resultOrange}\textbf{27.69} & \cellcolor{resultRed!100.00!resultOrange}\textbf{.0147} & \cellcolor{resultRed!100.00!resultOrange}\textbf{32.86} & \cellcolor{resultRed!100.00!resultOrange}\textbf{30.22} & \cellcolor{resultRed!100.00!resultOrange}\textbf{28.92} & \cellcolor{resultRed!100.00!resultOrange}\textbf{.0124} \\
\bottomrule
\end{tabular}
}
  \end{minipage}
\end{center}

\textbf{Insertion into a real reconstruction.}
The protocol above restricts the sphere to camera rays, which isolates occlusion and veiling and leaves lighting out.
\Cref{fig:suppl_sequence_insertion_reef} removes that restriction on a reconstructed reef, where no reference exists and the evaluation is qualitative.
Nothing in the scene is stylized and nothing is retrained, so the panels contain the reconstruction and the inserted bodies alone.

A body is placed on the reef and then moves above it along a path, and a stream of bubbles is released behind it.
Four effects follow from the insertion taking part in the same transport as the reconstruction.
The body is veiled by the water in front of it, and more so as it moves away.
That follows from the cells between it and the camera attenuating it at the depth it occupies.
No separate light is added for the bubbles.
They are lit by the recovered medium, so they carry the color of the water around them.
A refracting body sits on the reef in every panel and shows the scene behind it through the same water.
As the inserted body passes above the reef it darkens the surface under it.
Shadow rays are traced through the medium and the surfels together.

That last effect is the one the synthetic protocol was designed to exclude.
It is reported here as a qualitative result, with no reference to score it against.

\begin{center}
  \begin{minipage}{\textwidth}
  \centering
  \includegraphics[width=\textwidth]{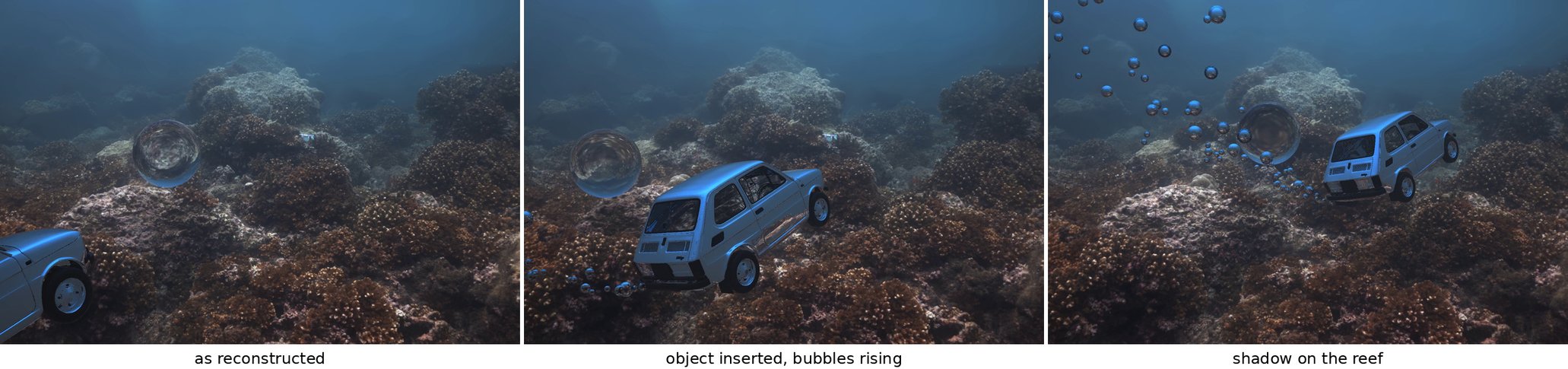}
  \captionof{figure}{Insertion into a reconstructed reef, with lighting left in. Look at the bubbles, which the recovered medium alone lights. Look also at the darkening of the reef under the body in the last panel.}
  \label{fig:suppl_sequence_insertion_reef}
  \end{minipage}
\end{center}

\section{Restoration under fog for every baseline}
\label{sec:suppl_fog}
\Cref{fig:suppl_garden} extends \cref{fig:garden} to both fog levels, both implementations of SeaThru-NeRF, and error maps.

\begin{center}
  \begin{minipage}{\textwidth}
  \includegraphics[width=\textwidth]{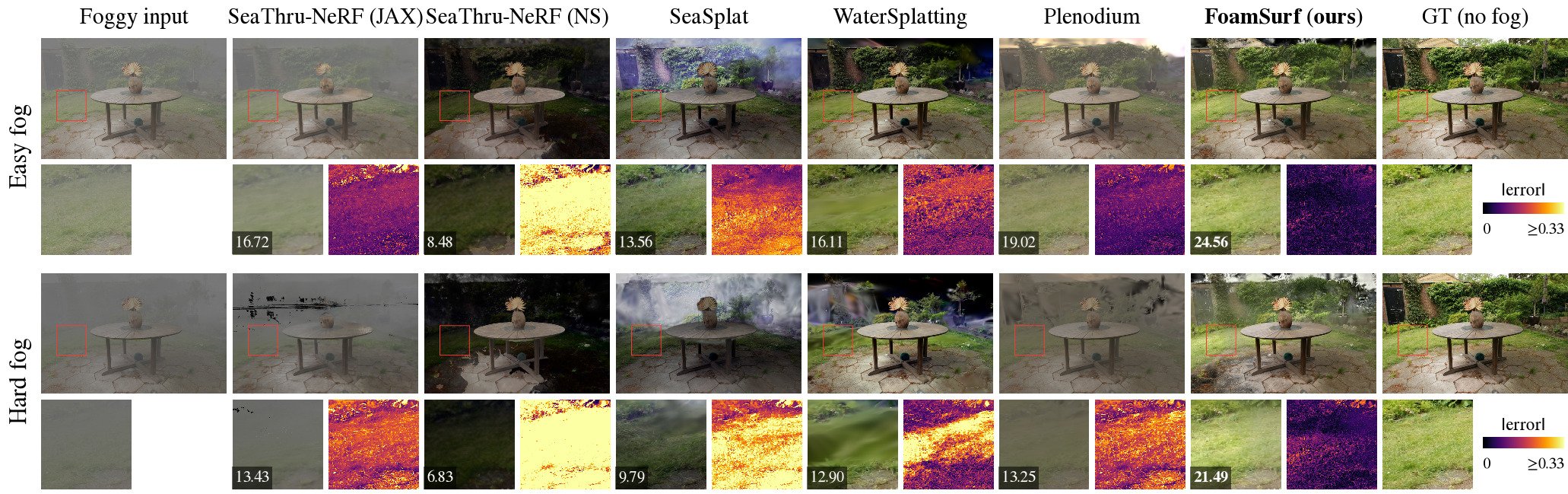}
  \captionof{figure}{Restoration of a Garden test view under easy fog (top) and hard fog (bottom). Each method renders the scene without its medium. Below each view are the red-box crop, its PSNR against the clear photograph, and its absolute error. Error maps share one color scale. The crop was selected to maximize the margin of \our{} over the best baseline, so it shows a larger gap than the full-image averages in \cref{tab:garden}. All methods use seed 42.}
  \label{fig:suppl_garden}
  \end{minipage}
\end{center}

\clearpage
\section{Synthetic scenes for every baseline}
\label{sec:suppl_synthetic}
\Cref{tab:suppl_synthetic} extends \cref{tab:synthetic} with structural similarity, LPIPS, the surface opacity in the cloud and both initializations of each baseline.
\Cref{fig:suppl_synthetic} extends \cref{fig:synthetic} to both synthetic scenes and shows the opacity of each medium.
Plenodium is absent here because its medium is present on every ray and has no setting that empties it.
On these scenes the background is empty space rather than water, so the medium covers the object and the reconstruction fails.
We therefore evaluate it on the real underwater captures, where its formulation applies.

\begin{center}
  \begin{minipage}{\textwidth}
  \centering
  \captionof{table}{Synthetic reconstruction on a black background. Values are means and standard deviations over three seeds, except WaterSplatting with its own initialization on Ficus (two seeds, one run failed). PSNR, SSIM and LPIPS evaluate full images. Object and Cloud evaluate surface-only and medium-only renders against their respective references. Opac. measures mean surface opacity on cloud-only pixels. Own init. and our init. distinguish each baseline's initialization from our procedure.}
  \label{tab:suppl_synthetic}
  \small
  \resizebox{\textwidth}{!}{
\begin{tabular}{l cccccc cccccc}
\toprule
 & \multicolumn{6}{c}{Lego} & \multicolumn{6}{c}{Ficus} \\
\cmidrule(lr){2-7} \cmidrule(lr){8-13}
Method & PSNR$\uparrow$ & SSIM$\uparrow$ & LPIPS$\downarrow$ & Object$\uparrow$ & Cloud$\uparrow$ & Opac.$\downarrow$ & PSNR$\uparrow$ & SSIM$\uparrow$ & LPIPS$\downarrow$ & Object$\uparrow$ & Cloud$\uparrow$ & Opac.$\downarrow$ \\
\midrule
WaterSplatting, own init. & \cellcolor{resultOrange!0.00!resultYellow}11.80\std{0.01} & \cellcolor{resultOrange!0.00!resultYellow}.640\std{.000} & \cellcolor{resultOrange!0.00!resultYellow}.303\std{.000} & \cellcolor{resultOrange!0.00!resultYellow}4.85\std{0.16} & \cellcolor{resultOrange!100.00!resultYellow}19.51\std{0.01} & \cellcolor{resultOrange!0.00!resultYellow}.915\std{.011} & \cellcolor{resultOrange!0.00!resultYellow}15.58\std{0.04} & \cellcolor{resultOrange!0.00!resultYellow}.654\std{.000} & \cellcolor{resultOrange!0.00!resultYellow}.255\std{.003} & \cellcolor{resultOrange!0.00!resultYellow}4.79\std{0.13} & \cellcolor{resultOrange!0.00!resultYellow}18.62\std{0.60} & \cellcolor{resultOrange!0.00!resultYellow}.977\std{.003} \\
WaterSplatting, our init. & \cellcolor{resultOrange!50.00!resultYellow}32.90\std{0.06} & \cellcolor{resultRed!33.33!resultOrange}.971\std{.001} & \cellcolor{resultOrange!66.67!resultYellow}.112\std{.000} & \cellcolor{resultOrange!50.00!resultYellow}15.44\std{0.21} & \cellcolor{resultOrange!0.00!resultYellow}19.47\std{0.00} & \cellcolor{resultOrange!50.00!resultYellow}.603\std{.003} & \cellcolor{resultOrange!50.00!resultYellow}24.10\std{4.18} & \cellcolor{resultOrange!50.00!resultYellow}.822\std{.101} & \cellcolor{resultOrange!50.00!resultYellow}.149\std{.066} & \cellcolor{resultOrange!50.00!resultYellow}18.07\std{3.56} & \cellcolor{resultOrange!100.00!resultYellow}21.71\std{0.00} & \cellcolor{resultOrange!100.00!resultYellow}.207\std{.143} \\
SeaSplat, own init. & \cellcolor{resultRed!50.00!resultOrange}35.27\std{0.10} & \cellcolor{resultRed!100.00!resultOrange}\textbf{.975}\std{.000} & \cellcolor{resultRed!100.00!resultOrange}\textbf{.094}\std{.000} & \cellcolor{resultRed!50.00!resultOrange}19.03\std{0.06} & \cellcolor{resultOrange!0.00!resultYellow}19.47\std{0.00} & \cellcolor{resultOrange!100.00!resultYellow}.340\std{.004} & \cellcolor{resultOrange!100.00!resultYellow}31.06\std{0.54} & \cellcolor{resultOrange!100.00!resultYellow}.972\std{.003} & \cellcolor{resultOrange!100.00!resultYellow}.094\std{.002} & \cellcolor{resultOrange!100.00!resultYellow}21.21\std{0.37} & \cellcolor{resultOrange!100.00!resultYellow}21.71\std{0.00} & \cellcolor{resultOrange!50.00!resultYellow}.229\std{.020} \\
SeaSplat, our init. & \cellcolor{resultRed!100.00!resultOrange}\textbf{35.44}\std{0.12} & \cellcolor{resultRed!100.00!resultOrange}\textbf{.975}\std{.000} & \cellcolor{resultRed!100.00!resultOrange}\textbf{.094}\std{.000} & \cellcolor{resultOrange!100.00!resultYellow}19.02\std{0.05} & \cellcolor{resultOrange!0.00!resultYellow}19.47\std{0.00} & \cellcolor{resultRed!50.00!resultOrange}.329\std{.008} & \cellcolor{resultRed!50.00!resultOrange}31.69\std{0.10} & \cellcolor{resultRed!50.00!resultOrange}.975\std{.000} & \cellcolor{resultRed!50.00!resultOrange}.091\std{.001} & \cellcolor{resultRed!50.00!resultOrange}21.31\std{0.30} & \cellcolor{resultOrange!100.00!resultYellow}21.71\std{0.00} & \cellcolor{resultRed!50.00!resultOrange}.201\std{.006} \\
\midrule
\our{} & \cellcolor{resultOrange!100.00!resultYellow}34.13\std{0.08} & \cellcolor{resultOrange!66.67!resultYellow}.970\std{.000} & \cellcolor{resultRed!33.33!resultOrange}.098\std{.000} & \cellcolor{resultRed!100.00!resultOrange}\textbf{34.77}\std{0.09} & \cellcolor{resultRed!100.00!resultOrange}\textbf{38.04}\std{0.03} & \cellcolor{resultRed!100.00!resultOrange}\textbf{.000}\std{.000} & \cellcolor{resultRed!100.00!resultOrange}\textbf{32.88}\std{0.02} & \cellcolor{resultRed!100.00!resultOrange}\textbf{.978}\std{.000} & \cellcolor{resultRed!100.00!resultOrange}\textbf{.089}\std{.000} & \cellcolor{resultRed!100.00!resultOrange}\textbf{32.89}\std{0.08} & \cellcolor{resultRed!100.00!resultOrange}\textbf{45.12}\std{4.02} & \cellcolor{resultRed!100.00!resultOrange}\textbf{.000}\std{.000} \\
\bottomrule
\end{tabular}
}
  \end{minipage}
\end{center}

\begin{center}
  \begin{minipage}{\textwidth}
  \centering
  \includegraphics[width=\textwidth]{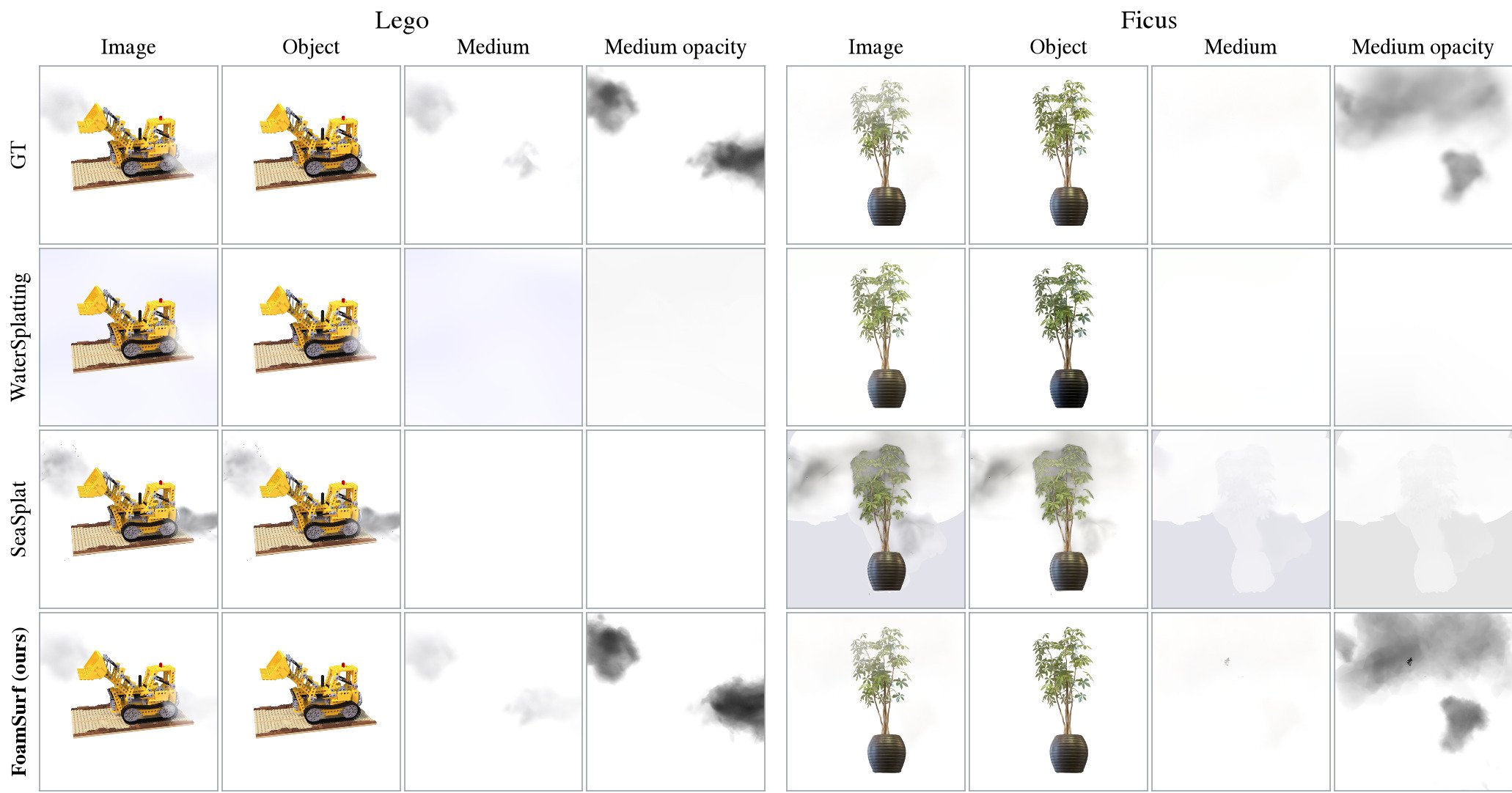}
  \captionof{figure}{One test view of each synthetic scene on a white background. Columns show the full image, surface alone, medium alone, and medium opacity (black is opaque). Baselines use our initialization procedure. Reference opacity describes the clouds alone. For WaterSplatting and SeaSplat, opacity is derived from attenuation up to each pixel's surface depth. Their medium renders do not recover the cloud shapes, whereas \our{} represents the clouds in its spatial medium. Quantitative evaluation uses a black background (\cref{tab:suppl_synthetic}), since a white background can favor a uniformly light medium.}
  \label{fig:suppl_synthetic}
  \end{minipage}
\end{center}

\section{Editing and style transfer}
\label{sec:suppl_editing}

The results below are qualitative and come from the models and renders of \cref{sec:supp_editing}.
Every panel is a test view of a reconstructed scene, rendered after the edit without further training.
\Cref{fig:suppl_edit_surface_layers,fig:suppl_edit_water_layers} render the edited component alone, which shows that each edit stays inside the representation it was applied to.

\begin{center}
  \begin{minipage}{\textwidth}
  \centering
  \includegraphics[width=\textwidth]{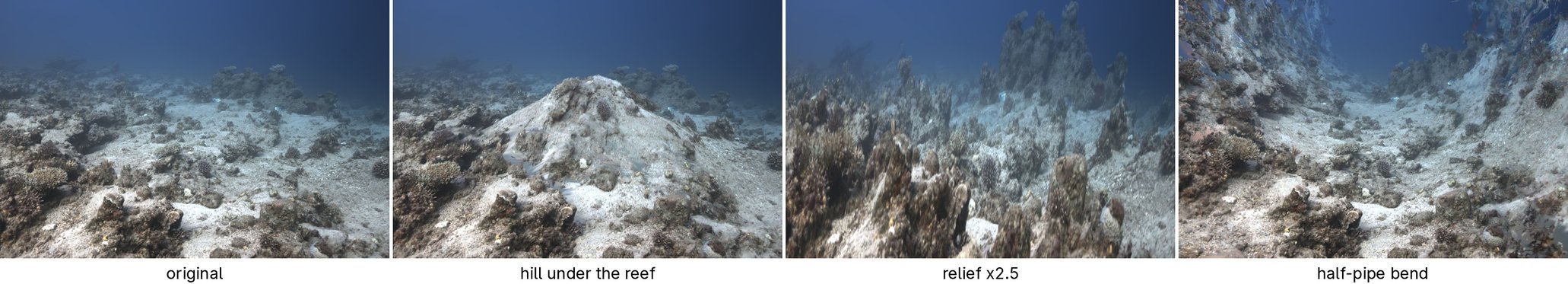}
  \captionof{figure}{Surface edits on one real scene. A deformation moves the surfels, while the water stays as reconstructed. The water in front of the seabed therefore keeps its attenuation and backscatter, and the raised hill is veiled by it in the same way as the original seabed.}
  \label{fig:suppl_edit_surface}
  \end{minipage}
\end{center}

\begin{center}
  \begin{minipage}{\textwidth}
  \centering
  \includegraphics[width=\textwidth]{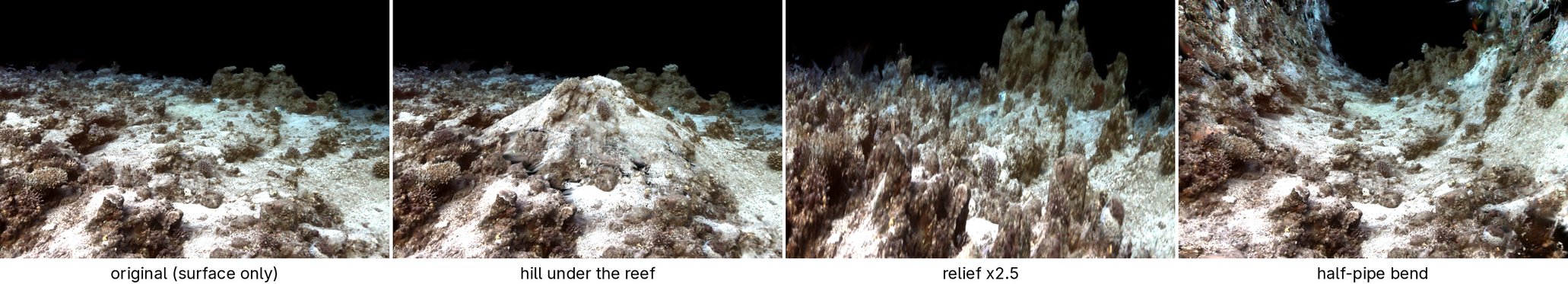}
  \captionof{figure}{The surfels of \cref{fig:suppl_edit_surface} rendered alone. The deformation changes the surfaces and nothing else. Where a deformation uncovers places that no camera saw, such as the far side of a rock, no surfels exist and the view passes through.}
  \label{fig:suppl_edit_surface_layers}
  \end{minipage}
\end{center}

\begin{center}
  \begin{minipage}{\textwidth}
  \centering
  \includegraphics[width=\textwidth]{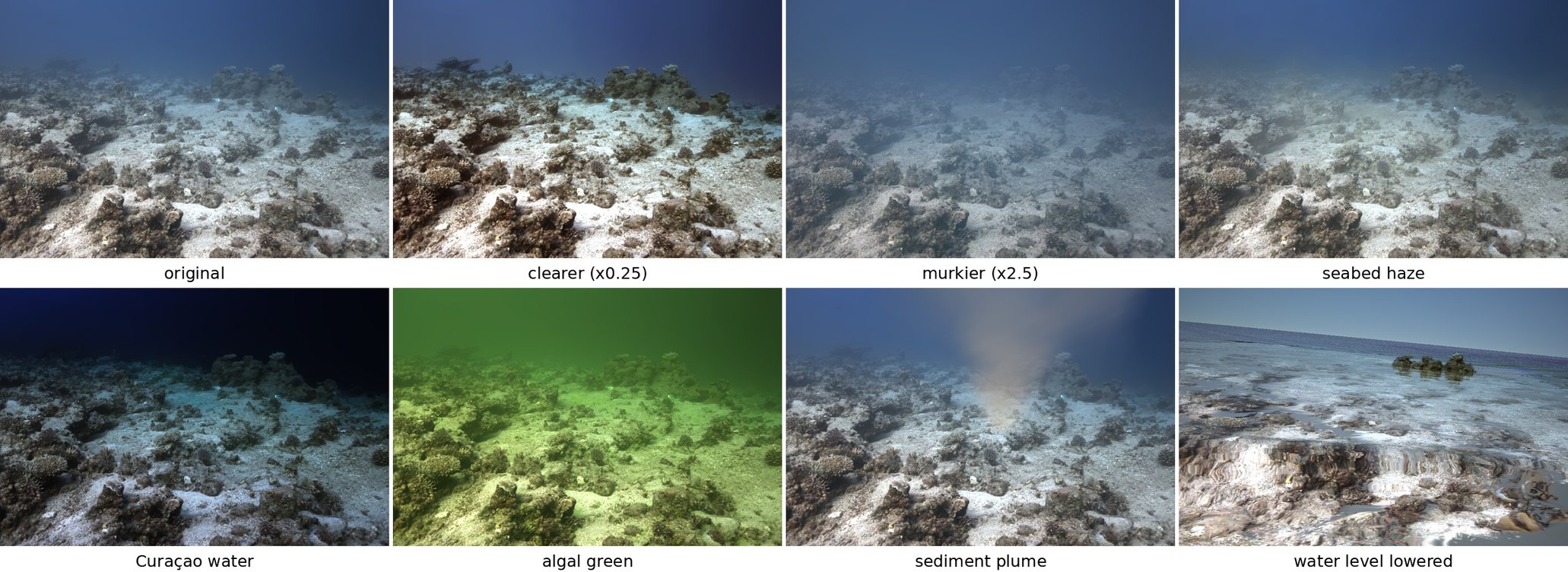}
  \captionof{figure}{Medium edits on the same scene and the same view. The surfels are the reconstructed ones in every panel. Scaling the cell coefficients clears or thickens the water, the water of another scene is transplanted by matching per-channel statistics, and a plume and a haze layer are painted into locally refined cells. In the last panel the water of each ray is clipped below a wavy surface, which exposes the seabed.}
  \label{fig:suppl_edit_water}
  \end{minipage}
\end{center}

\begin{center}
  \begin{minipage}{\textwidth}
  \centering
  \includegraphics[width=\textwidth]{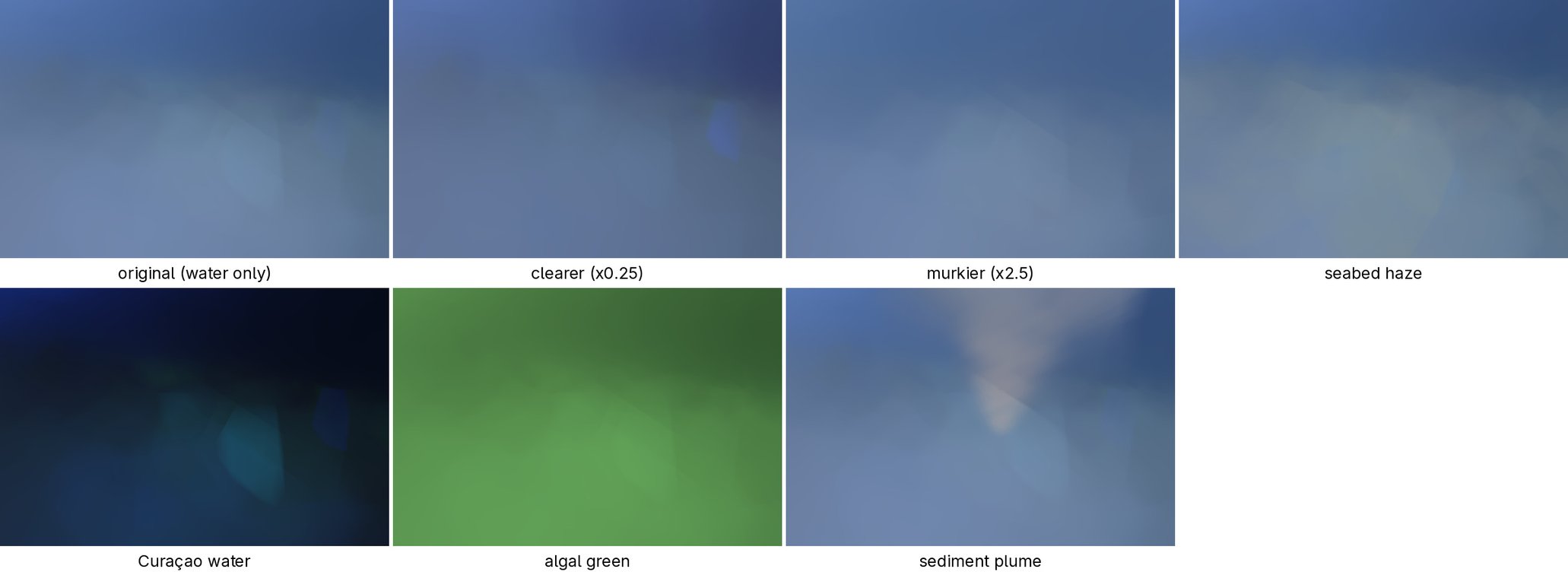}
  \captionof{figure}{The medium of \cref{fig:suppl_edit_water} rendered alone. Each edit is confined to the cells, and the spatial pattern of the reconstructed water survives the transplant.}
  \label{fig:suppl_edit_water_layers}
  \end{minipage}
\end{center}

\textbf{An edit driven by an event.}
The medium edits above are set by hand and hold still.
\Cref{fig:suppl_sequence_sediment} drives one from the scene instead.
Two edits are combined there.
The seabed is deformed into a mound, which moves the surfels and leaves the water as reconstructed.
A body is then released above the mound and falls onto it under the same insertion machinery as \cref{sec:suppl_insertion}.
Where it lands, the attenuation and backscatter coefficients of the cells around the impact are raised and then relaxed over the following frames.
That reads as sediment thrown up and settling.
The cloud is at its densest two seconds after the impact and has spread by the fifth second.

Three properties of the representation are visible in that sequence.
The cloud occupies space instead of forming a layer over the image.
It therefore veils the far side of the mound while the near side stays clear.
It is carried by the cells themselves, so no particle system and no second medium are involved.
The water of the rest of the scene is untouched.
A refracting body rests on the left of every panel, and the scene behind it is traced through the edited water.
The cloud therefore appears inside that body as it rises.
All three follow from the medium being a field in space that the tracer reads along every ray, primary or secondary.

\begin{center}
  \begin{minipage}{\textwidth}
  \centering
  \includegraphics[width=\textwidth]{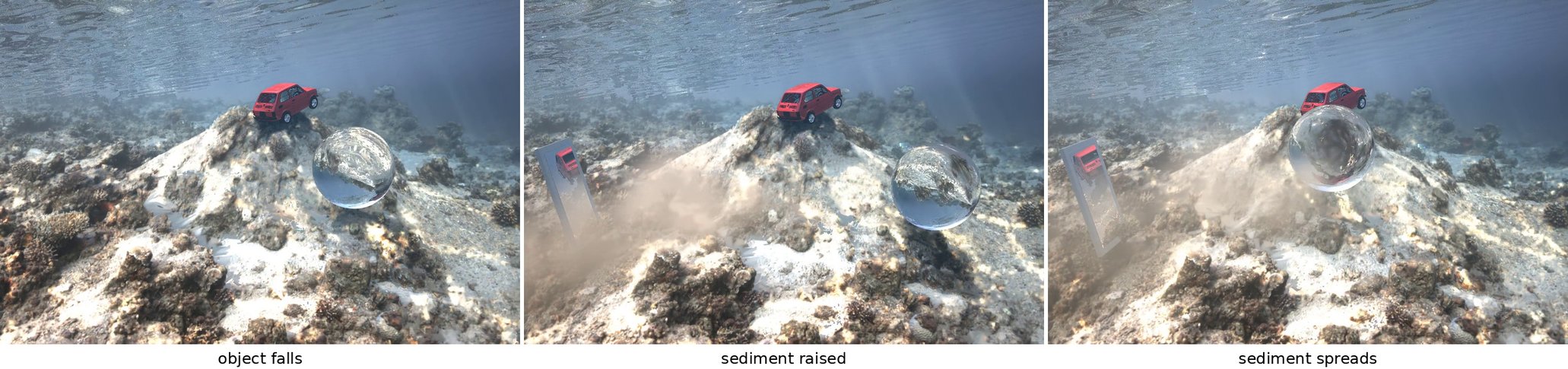}
  \captionof{figure}{Sediment raised by an impact, one, three and five seconds after it. The surfels carry the deformed seabed and the cells carry the cloud. Look at the far side of the mound, which the cloud veils. Look also at the refracting body on the left, which shows the same cloud through it.}
  \label{fig:suppl_sequence_sediment}
  \end{minipage}
\end{center}

\begin{center}
  \begin{minipage}{\textwidth}
  \centering
  \includegraphics[width=\textwidth]{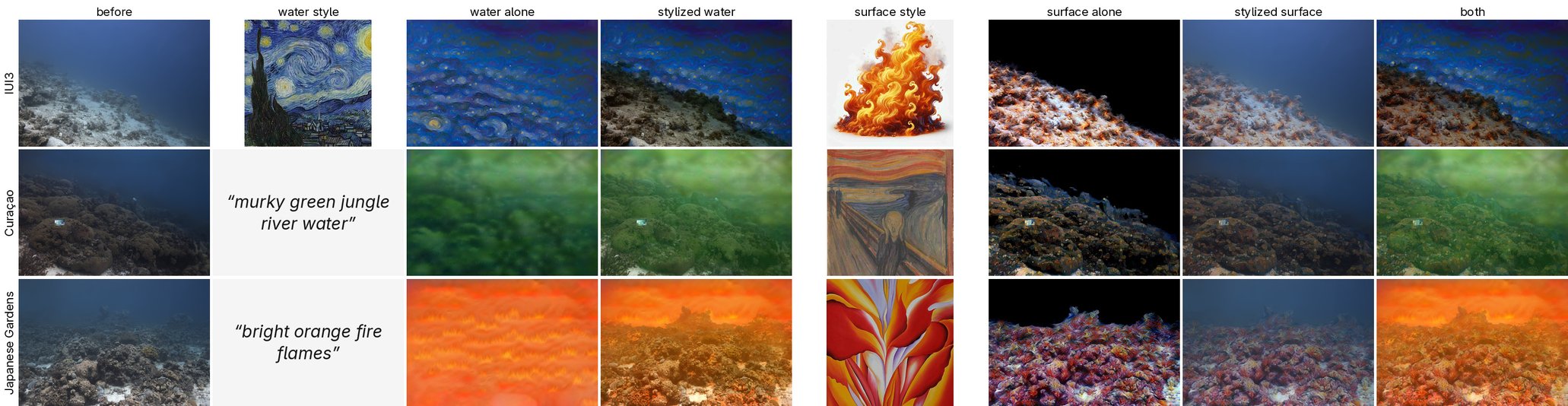}
  \captionof{figure}{Style transfer applied to each component separately, on three real scenes. The style of the water is an image in the first row and a text prompt in the others. Each component is optimized against its own render, so the stylized water and the stylized surfaces can be viewed alone or combined. The attenuation of the water is never optimized, so the seabed stays visible through it. For the fire prompt the backscatter coefficients are frozen as well (\cref{sec:supp_editing}).}
  \label{fig:suppl_style}
  \end{minipage}
\end{center}

\section{Composition of the teaser}
\label{sec:suppl_teaser}

\Cref{fig:teaser} of the main paper combines a reconstruction, several edits, and a few effects that belong to none of them.
\Cref{fig:suppl_sequence_teaser} separates that combination into the steps that produced it, so that each panel can be attributed.

The scene was rendered in Blender and reconstructed from the resulting photographs, every one of which looks through water.
The first panel is that reconstruction, rendered from a novel view and otherwise untouched.
In the second the medium is removed inside a boundary.
What appears there is seabed that no camera ever saw directly, only through the water.
It appears because the surfaces and the water occupy separate primitives, so emptying the cells leaves the surfels in place.
The third panel adds shafts of light.
The fourth inserts a statue, and the fifth adds two refracting spheres beside it.
Both are traced along rays that cross the water the same way camera rays do.
They therefore darken toward the far columns and pick up the color of the water between them and the viewer.

Three elements are not reconstructed and not edits.
The waves at the surface, the sky above them, and the shafts of light are visual effects composited into the rendering.
They are stated as such in the caption of \cref{fig:teaser}, and the panels here make the boundary explicit.
Everything else in every panel comes from the reconstruction or from an edit of it, and no panel involves further training.

\begin{center}
  \begin{minipage}{\textwidth}
  \centering
  \includegraphics[width=\textwidth]{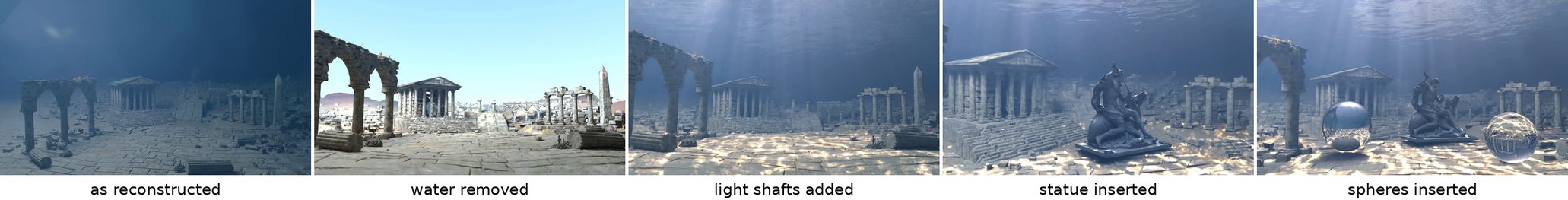}
  \captionof{figure}{The teaser of the main paper, separated into the steps that built it. One edit is added per panel and the reconstruction underneath is the same throughout. Light shafts, waves and sky are visual effects.}
  \label{fig:suppl_sequence_teaser}
  \end{minipage}
\end{center}

\clearpage
\section{The separation priors one at a time}
\label{sec:suppl_prior_ablation}
The ablation of \cref{tab:ablation} switches every separation prior off together.
It shows what the group contributes and cannot say which member contributes it.
Here each prior is removed on its own, with every other setting at the default of \cref{sec:supp_weights}.
Runs cover the four SeaThru-NeRF scenes with two seeds.
Each is compared against the full method trained on that scene with that seed.
This follows the protocol of \cref{tab:ablation}, which also uses two seeds.

\Cref{tab:suppl_prior_ablation} reports the result.
Six of the seven priors move PSNR by less than the spread the full method shows across seeds.
Medium smoothness is the exception, and its cost concentrates on two of the four scenes.
The water clearing prior is the one whose removal is visible in the opacity components.
Without it the surfels take opacity both on the seabed and in the open water.
The separation score averages seabed opacity against water transparency, so two increases of similar size cancel inside it.
That score is computed from surfel-only opacity.
It therefore cannot see what the medium does behind the surface, and two of the priors act there.

The near-zero column of \cref{tab:suppl_prior_ablation} is the result, not a missing effect.
A rendered photograph is a composite of the surface and the medium.
Moving responsibility from one to the other leaves the composite where it was and changes the decomposition underneath.
\Cref{fig:suppl_prior_same_image} shows one view under the full method and under all seven variants.
The eight photographs are not distinguishable.
\Cref{fig:suppl_prior_alpha,fig:suppl_prior_medium,fig:suppl_prior_surfels} show what separates them.
Surfel opacity fills the open water without the clearing prior and opens holes in the seabed without coverage or binarization.
The medium loses the hue of the water without the chroma prior and loses spatial coherence without medium smoothness.
The surfels absorb brightness that belongs to the medium without the overlap prior and without the camera water prior.
We measured the difference from the full method over four scenes, two seeds and three views.
On the opacity map it reaches $6.4\%$ to $9.3\%$ of the signal.
On the photograph it stays between $0.7\%$ and $1.1\%$.

\begin{center}
  \begin{minipage}{\textwidth}
  \centering
  \captionof{table}{Each separation prior removed on its own. Values are changes against the full method, paired on the scene and the seed, then averaged over four scenes and two seeds. Seabed and Water give surface-only opacity on the evaluation masks of \cref{tab:seathru}. Rows run from the largest loss in PSNR to the smallest.}
  \label{tab:suppl_prior_ablation}
  \small
\begin{tabular}{l rrrr}
\toprule
Prior removed & $\Delta$PSNR & $\Delta$Seabed & $\Delta$Water & $\Delta$Separation \\
\midrule
Medium smoothness & -0.45 & -0.009 & -0.001 & -0.004 \\
Backscatter chroma & -0.07 & -0.011 & +0.001 & -0.006 \\
Opacity binarization & -0.06 & -0.019 & 0.000 & -0.010 \\
Seabed coverage & -0.02 & -0.016 & 0.000 & -0.008 \\
Water clearing & -0.01 & +0.052 & +0.047 & +0.002 \\
Homogeneous water at the cameras & +0.01 & -0.001 & 0.000 & -0.001 \\
Surface–medium overlap & +0.03 & +0.001 & 0.000 & 0.000 \\
\bottomrule
\end{tabular}

  \end{minipage}
\end{center}

\begin{center}
  \begin{minipage}{\textwidth}
  \centering
  \includegraphics[width=\textwidth]{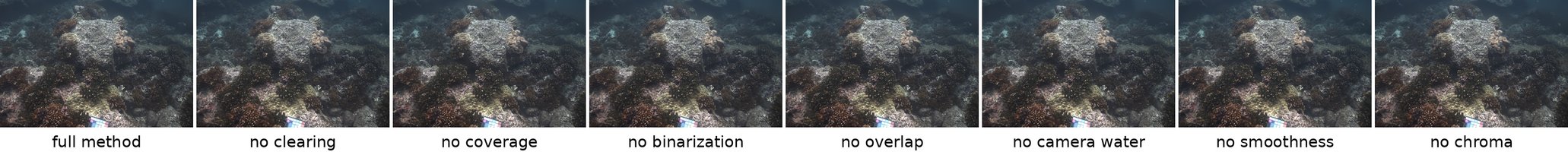}
  \captionof{figure}{One prior removed, the same photograph. A single test view of Panama, rendered by the full method and by each variant. Nothing is marked because the eight renders agree. \Cref{fig:suppl_prior_alpha,fig:suppl_prior_medium,fig:suppl_prior_surfels} show the decompositions behind them.}
  \label{fig:suppl_prior_same_image}
  \end{minipage}
\end{center}

\begin{center}
  \begin{minipage}{\textwidth}
  \centering
  \includegraphics[width=\textwidth]{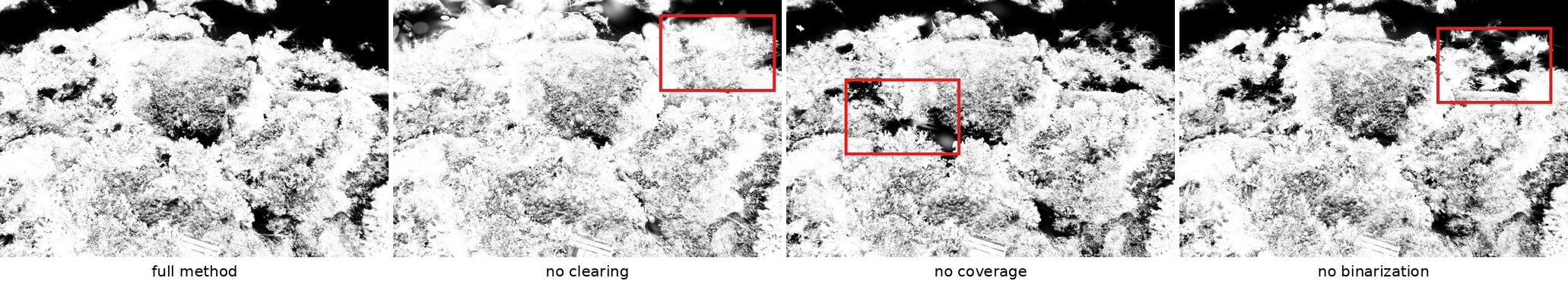}
  \captionof{figure}{Surface-only opacity on the view of \cref{fig:suppl_prior_same_image}. White is opaque surface. The band of open water above the reef is empty under the full method and takes opacity once the clearing prior is removed. Coverage and binarization leave holes in the seabed instead. Each box marks the window that departs most from the full method, measured on the difference between the two renders. The differences extend beyond the box.}
  \label{fig:suppl_prior_alpha}
  \end{minipage}
\end{center}

\begin{center}
  \begin{minipage}{\textwidth}
  \centering
  \includegraphics[width=\textwidth]{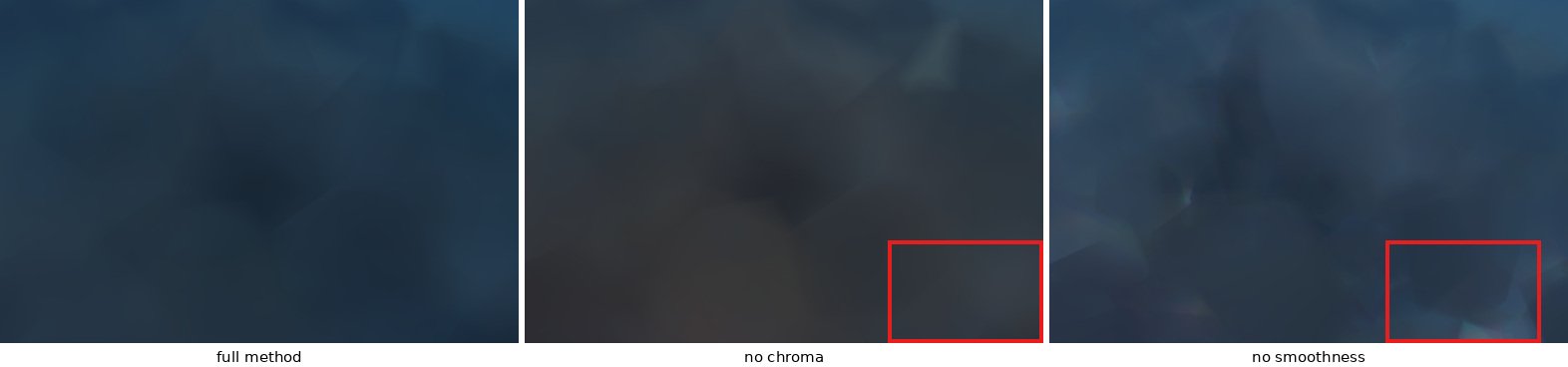}
  \captionof{figure}{The medium rendered alone, same view and box convention as \cref{fig:suppl_prior_alpha}. Without the chroma prior the haze takes a brown cast that the water does not have. Without medium smoothness the field breaks into patches. Both changes cover the whole frame, not only the marked window.}
  \label{fig:suppl_prior_medium}
  \end{minipage}
\end{center}

\begin{center}
  \begin{minipage}{\textwidth}
  \centering
  \includegraphics[width=\textwidth]{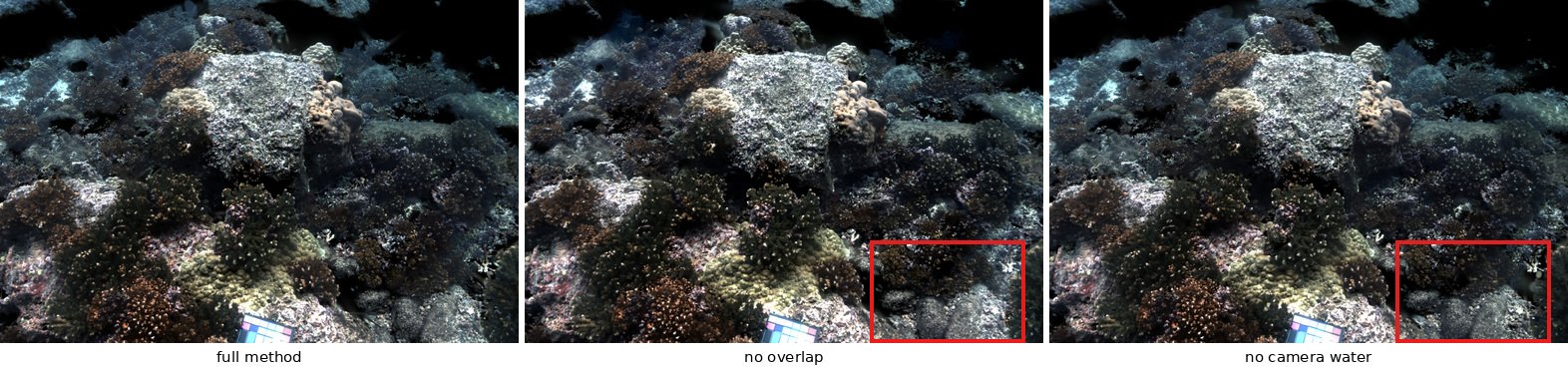}
  \captionof{figure}{The surfels rendered alone, with the medium removed. The view and the box follow \cref{fig:suppl_prior_alpha}. The full method keeps the surface dark where the water carried the brightness. Removing either the overlap prior or the camera water prior leaves that brightness in the surface albedo.}
  \label{fig:suppl_prior_surfels}
  \end{minipage}
\end{center}

\fi

\end{document}